\documentclass[preprint,12pt]{elsarticle}
\usepackage{rawfonts}
\usepackage{times}
\usepackage{helvet}
\usepackage{courier}
\usepackage{amssymb}
\usepackage{amsmath}
\usepackage{array}
\usepackage{float}
\usepackage{graphics,graphicx,epsfig,epstopdf}
\usepackage{psfrag}
\usepackage{subfigure}
\usepackage{tikz}
\usepackage{rotating}
\usepackage{url}
\usepackage{color}
\usepackage{theorem}
\usepackage{algorithm}
\usepackage{algpseudocode}
\usepackage{verbatim}
\usepackage{enumitem}
\usepackage{multirow}
\usepackage{slashbox}
\newtheorem{theorem}{Theorem}

\newtheorem{lemma}[theorem]{Lemma}
\newtheorem{definition}{Definition}
\newtheorem{example}{Example}

\newtheorem{proposition}{Proposition}
\newtheorem{question}{Question}
\newtheorem{remark}{Remark}

\DeclareMathOperator*{\argmin}{arg\,min}
\DeclareMathOperator*{\argmax}{arg\,max}
\journal{Artificial Intelligence}

\begin{document}

\begin{frontmatter}



\title{Adversarial patrolling with spatially uncertain alarm signals}

\author[unimi]{Nicola Basilico}
\author[polimi]{Giuseppe De Nittis}
\author[polimi]{Nicola Gatti}
\address[unimi]{Department of Computer Science, University of Milan, Milano, Italy}
\address[polimi]{Dipartimento di Elettronica, Informazione e Bioingegneria, Politecnico di Milano, Milano, Italy}


\begin{abstract}
When securing complex infrastructures or large environments, constant surveillance of every area is not affordable. To cope with this issue, a common countermeasure is the usage of cheap but wide--ranged sensors, able to detect suspicious events that occur in large areas, supporting patrollers to improve the effectiveness of their strategies. However, such sensors are commonly affected by uncertainty. In the present paper, we focus on spatially uncertain alarm signals. That is, the alarm system is able to \emph{detect} an attack but it is uncertain on the \emph{exact position} where the attack is taking place. This is common when the area to be secured is wide such as in border patrolling and fair site surveillance. We propose, to the best of our knowledge, the first Patrolling Security Game model where a \emph{Defender} is supported by a spatially uncertain \emph{alarm system} which non--deterministically generates \emph{signals} once a target is under attack. We show that finding the optimal strategy in arbitrary graphs is $\mathcal{APX}$--hard even in zero--sum games and we provide two (exponential time) exact algorithms and two (polynomial time) approximation algorithms. Furthermore, we analyse what happens in environments with special topologies, showing that in linear and cycle graphs the optimal patrolling strategy can be found in polynomial time, \textit{de facto} allowing our algorithms to be used in real--life scenarios, while in trees the problem is $\mathcal{NP}$--hard. Finally, we show that without false positives and missed detections, the best patrolling strategy reduces to stay in a place, wait for a signal, and respond to it at best. This strategy is optimal even with non--negligible missed detection rates, which, unfortunately, affect every commercial alarm system. We evaluate our methods in simulation, assessing both quantitative and qualitative aspects.
\end{abstract}

\begin{keyword}
Security Games \sep Adversarial Patrolling \sep Algorithmic Game Theory



\end{keyword}

\end{frontmatter}
%
%
%

\section{Introduction}
\label{sec:introduction}

Security Games model the task of protecting physical environments as a non--cooperative game between a {\em Defender} and an {\em Attacker}~\cite{jain2012overview}. These games usually take place under a \emph{Stackelberg} (a.k.a. \emph{leader--follower}) paradigm~\cite{von2004leadership}, where the Defender (\emph{leader}) commits to a strategy and the Attacker (\emph{follower}) first observes such commitment, then best responds to it. As discussed in the seminal work~\cite{commitTo}, finding a leader--follower equilibrium is computationally tractable in games with one follower and complete information, while it becomes hard in Bayesian games with different types of Attacker. The availability of such computationally tractable aspects of Security Games led to the development of algorithms capable of scaling up to large problems, making them deployable in the security enforcing systems of several real--world applications. The first notable examples are the deployment of police checkpoints at the Los Angels International Airport~\cite{armor} and the scheduling of federal air marshals over the U.S. domestic airline flights~\cite{iris}. More recent case studies include the positioning of U.S. Coast Guard patrols to secure crowded places, bridges, and ferries~\cite{protect} and the arrangement of city guards to stop fare evasion in Los Angeles Metro~\cite{trusts}. Finally, a similar approach is being tested and evaluated in Uganda, Africa, for the protection of wildlife~\cite{paws}. Thus, Security Games emerged as an interesting game theoretical tool and then showed their on--the-field effectiveness in a number of real security scenarios. 

We focus on a specific class of security games, called \emph{Patrolling Security Games}. These games are modelled as infinite--horizon extensive--form games in which the Defender controls one or more {\em patrollers} moving within an environment represented as a discrete graph. The Attacker, besides having knowledge of the strategy to which the Defender committed to, can observe the movements of the patrollers at any time and use such information in deciding the most convenient time and target location to attack~\cite{BasilicoGA12}. When multiple patrollers are available, coordinating them at best is in general a hard task which, besides computational aspects, must also keep into account communication issues~\cite{BasilicoGattiAAAI2010}. However, the patrolling problem is tractable, even with multiple patrollers, in border security (e.g., linear and cycle graphs), when patrollers have homogeneous moving and sensing capabilities and all the vertices composing the border share the same features~\cite{gallo1}. Scaling this model involved the study of how to compute patrolling strategies in scenarios where the Attacker is allowed to perform multiple attacks~\cite{multiattack}. Similarly, coordination strategies among multiple Defenders are investigated in~\cite{gallo2}. In~\cite{an2014adp}, the authors study the case in which there is a temporal discount on the targets. Extensions are discussed in~\cite{BO1}, where coordination strategies between defenders are explored, in~\cite{BO2}, where a resource can cover multiple targets, and in~\cite{AgmonEvents} where attacks can detected at different stages with different associated utilities. Finally, some theoretical results about properties of specific patrolling settings are provided in~\cite{alpern}. In the present paper, we provide a new model of patrolling security games in which the Defender is supported by an \emph{alarm system} deployed in the environment.

\subsection{Motivation scenarios}
Often, in large environments, a constant surveillance of every area is not affordable while focused inspections triggered by alarms are more convenient. Real--world applications include UAVs surveillance of large infrastructures~\cite{BasilicoCarpinChungDARS2014}, wildfires detection with CCD cameras~\cite{wildfire2013}, agricultural fields monitoring~\cite{wsnagri2011}, and surveillance based on wireless sensor networks~\cite{wsnsurvey2008}, and border patrolling~\cite{borderPatrol}. Alarm systems are in practice affected by \emph{detection} uncertainty, e.g., missed detections and false positives, and \emph{localization} (a.k.a. spatial) uncertainty, e.g., the alarm system is uncertain about the exact target under attack. We summarily describe two practical security problems that can be ascribed to this category. We report them as examples, presenting features and requirements that our model can properly deal with. In the rest of the paper we will necessarily take a general stance, but we encourage the reader to keep in mind these two cases as reference applications for a real deployment of our model.

\subsubsection{Fight to illegal poaching} Poaching is a widespread environmental crime that causes the endangerment of wildlife in several regions of the world. Its devastating impact makes the development of surveillance techniques to contrast this kind of activities one of the most important matters in national and international debates. Poaching typically takes place over vast and savage areas, making it costly and ineffective to solely rely on persistent patrol by ranger squads. To overcome this issue, recent developments have focused on providing rangers with environmental monitoring systems to better plan their inspections, concentrating them in areas with large likelihood of spotting a crime. Such systems include the use of UAVs flying over the area, alarmed fences, and on--the--field sensors trying to recognize anomalous activities.\footnote{See, for example, \url{http://wildlandsecurity.org/}.} In all these cases, technologies are meant to work as an alarm system: once the illegal activity is recognized, a signal is sent to the rangers base station from where a response is undertaken. In the great majority of cases, a signal corresponds to a spatially uncertain localization of the illegal activity. For example, a camera--equipped UAV can spot the presence of a pickup in a forbidden area but cannot derive the actual location to which poachers are moving. In the same way, alarmed fences and sensors can only transmit the location of violated entrances or forbidden passages. In all these cases a signal implies a restricted, yet not precise, localization of the poaching activity. The use of security games in this particular domain is not new (see, for example,~\cite{paws}). However, our model allows the computation of alarm response strategies for a given alarm system deployed on the field. This can be done by adopting a discretization of the environment, where each target corresponds to a sector, values are related to the expected population of animals in that sector, and deadlines represent the expected completion time of illegal hunts (these parameters can be derived from data, as discussed in~\cite{paws}).

\subsubsection{Safety of fair sites}\label{sec:expo}
Fairs are large public events attended by thousands of visitors, where the problem of guaranteeing safety for the hosting facilities can be very hard. For example, Expo 2015, the recent Universal Exposition hosted in Milan, Italy, estimates an average of about 100,000 visits per day. This poses the need for carefully addressing safety risks, which can also derive from planned act of vandalism or terrorist attacks. Besides security guards patrols, fair sites are often endowed with locally installed monitoring systems. Expo 2015 employs around 200 baffle gates and 400 metal detectors at the entrance of the site. The internal area is constantly monitored by 4,000 surveillance cameras and by 700 guards. Likely, when one or more of these devices/personnel identify a security breach, a signal is sent to the control room together with a circumscribed request of intervention. This approach is required because, especially in this kind of environments, detecting a security breach and neutralizing it are very different tasks. The latter one, in particular, usually requires a greater effort involving special equipment and personnel whose employment on a demand basis is much more convenient. Moreover, the detecting location of a threat is in many cases different from the location where it could be neutralized, making the request of intervention spatially uncertain. For instance, consider a security guard or a surveillance camera detecting the visitors' reactions to a shooting rampage performed by some attacker. In examples like these, we can restrict the area where the security breach happened but no precise information about the location can be gathered since the attacker will probably have moved. Our model could be applied to provide a policy with which schedule interventions upon a security breach is detected in some particular section of the site. In such case, targets could correspond to buildings or other installations where visitors can go. Values and deadlines can be chosen according to the importance of targets, their expected crowding, and the required response priority.

\subsection{Alarms and security games}
While the problem of managing a sensor network to optimally guard security--critical infrastructure has been investigated in restricted domains, e.g.~\cite{krause}, the problem of integrating alarm signals together with adversarial patrolling is almost completely unexplored. The only work that can be classified under this scope is~\cite{munoz2013introducing}. The paper proposes a skeleton model of an alarm system where sensors have no spatial uncertainty in detecting attacks on single targets. The authors analyse how sensory information can improve the effectiveness of patrolling strategies in adversarial settings. They show that, when sensors are not affected by false negatives and false positives, the best strategy prescribes that the patroller just responds to an alarm signal rushing to the target under attack without patrolling the environment. As a consequence, in such cases the model treatment becomes trivial. On the other hand, when sensors are affected only by false negatives, the treatment can be carried out by means of an easy variation of the algorithm for the case without sensors~\cite{BasilicoGA12}. In the last case, where false positives are admitted, the problem becomes computationally intractable. To the best of our knowledge, no previous result dealing with spatial uncertain alarm signals in adversarial patrolling is present in the literature.

Effectively exploiting an alarm system and determining a good deployment for it (e.g., selecting the location where install sensor) are complementary but radically different problems. The results we provide in this work lie in the scope of the first one while the treatment of the second one is left for future works. In other words, we assume that a deployed alarm system is given and we deal with the problem of strategically exploiting it at best. Any approach to search for the optimal deployment should, in principle, know how to evaluate possible deployments. In such sense, our problem needs to be addressed before one might deal with the deployment one.

\subsection{Contributions}
In this paper, we propose the first Security Game model that integrates a spatially uncertain alarm system in game--theoretic settings for patrolling.\footnote{A very preliminary short version of the present paper is~\cite{patrolSensor14}.} Each alarm signal carries the information about the set of targets that can be under attack and it is described by the probability of being generated when each target is attacked. Moreover, the Defender can control only one  patroller. The game can be decomposed in a finite number of finite--horizon subgames, each called Signal Response Game from $v$ (SRG--$v$) and capturing the situation in which the Defender is in a vertex $v$ and the Attacker attacked a target, and an infinite--horizon game, called Patrolling Game (PG), in which the Defender moves in absence of any alarm signal. We show that, when the graph has arbitrary topology, finding the equilibrium in each SRG--$v$ is $\mathcal{APX}$--hard even in the zero--sum case. We provide two exact algorithms. The first one, based on dynamic programming, performs a breadth--first search, while the second one, based on branch--and--bound approach, performs a depth--first search. We use the same two approaches to design two approximation  algorithms. Furthermore, we provide a number of additional results for the SRG--$v$.  We study special topologies, showing that there is a polynomial time algorithm solving a SRG--$v$ on linear and cycle graphs, while it is $\mathcal{NP}$--hard with trees. Then, we study the PG, showing that when no false positives and no missed detections are present, the optimal Defender strategy is to stay in a fixed location, wait for a signal, and respond to it at best. This strategy keeps being optimal even when non--negligible missed detection rates are allowed. We experimentally evaluate the scalability of our exact algorithms and we compare them w.r.t. the approximation ones in terms of solution quality and compute times, investigating in hard instances the gap between our hardness results and the theoretical guarantees of our approximation algorithms. We show that our approximation algorithms provide very high quality solutions even in hard instances. Finally, we provide an example of resolution for a realistic instance, based on Expo 2015, and we show that our exact algorithms can be applied for such kind of settings. Moreover, in our realistic instance we assess how the optimal patrolling strategy coincides with a static placement even when allowing a false positive rate of less or equal to $30\%$. 

\subsection{Paper structure}
In Section~\ref{sec:problem_formulation}, we introduce our game model. In Section~\ref{sec:srg_on_arbitrary_graphs}, we study the problem of finding the strategy of the Defender for responding to an alarm signal in an arbitrary graph while in Section~\ref{sec:srg_on_special_topologies}, we provide results for specific topologies. In Section~\ref{sec:patrolling_game}, we study the patrolling problem. In Section~\ref{sec:experimental_evaluation}, we experimentally evaluate our algorithms. In Section~\ref{sec:related_works}, we briefly discuss the main Security Games research directions that have been explored in the last decades. Finally, Section~\ref{sec:conclusions_and_future_research} concludes the paper. \ref{appendix:notation} includes a notation table, while \ref{appendix:boxplots} reports some additional experimental results.
\section{Problem statement}
\label{sec:problem_formulation}
In this section we formalize the problem we study. More precisely, in Section~\ref{subsec:model} we describe the patrolling setting and the game model, while in Section~\ref{subsec:questions} we state the computational questions we shall address in this work.

\subsection{Game model}
\label{subsec:model}
Initially, in Section~\ref{subsubsec:model_setting}, we introduce a basic patrolling security game model integrating the main features from models currently studied in literature. Next, in Section~\ref{subsubsec:model_model}, we extend our game model by introducing alarm signals. In Section~\ref{subsubsec:game_tree}, we depict the game tree of our patrolling security game with alarm signals and we decompose it in notable subgames to facilitate its study.

\subsubsection{Basic patrolling security game}
\label{subsubsec:model_setting}
As is customary in the artificial intelligence literature~\cite{BasilicoGA12,an2014adp}, we deal with discrete, both in terms of space and time, patrolling settings, representing an approximation of a continuous environment. Specifically, we model the environment to be patrolled as an undirected connected graph $G=(V,E)$, where vertices represent different areas connected by various corridors/roads, formalized through the edges. Time is  discretized in turns. We define $T\subseteq V$ the subset of sensible vertices, called \emph{targets}, that must be protected from possible attacks. Each target $t\in T$ is characterized by a value $\pi(t)\in (0,1]$ and a penetration time $d(t)\in \mathbb{N}^+$ which measures the number of turns needed to complete an attack to~$t$.

\begin{example}
We report in Figure~\ref{fig:graph_clessidra_single_signal} an example of patrolling setting. Here, $V=\{v_0,v_1,v_2,v_3,v_4\}$, $T=\{t_1, t_2, t_3, t_4\}$ where $t_i=v_i$ for $i\in\{1,2,3,4\}$. All the targets $t$ present the same value $\pi(t)$ and the same penetration time $d(t)$.

\begin{figure}[!htbp]
\centering
\begin{tabular}{c}
\begin{tikzpicture}
  [scale=.8,auto=left,every node/.style={circle,fill=white,draw,minimum size = 1cm,font=\sffamily\large\bfseries}]
  \node (n1) at (0,0) {$v_0$};
  \node (n2) at (-2,2) {$t_1$};
  \node (n3) at (2,2) {$t_2$};
  \node (n4) at (2,-2) {$t_3$};
  \node (n5) at (-2,-2) {$t_4$};

  \path[every node/.style={font=\sffamily\small}]
    (n1) edge node [loop] {} (n2)
	(n1) edge node [loop] {} (n3)
	(n1) edge node [loop] {} (n4)
	(n1) edge node [loop] {} (n5)
	(n2) edge node [loop] {} (n3)
	(n4) edge node [loop] {} (n5);
	
\end{tikzpicture}
\end{tabular}
\begin{tabular}{ c | c | c }
    $t$ & $\pi(t)$ & $d(t)$ \\ \hline
    $t_1$ & 0.5 & 4   \\ \hline
    $t_2$ & 0.5 & 4 \\ \hline
    $t_3$ & 0.5 & 4 \\ \hline
    $t_4$ & 0.5 & 4  
\end{tabular}
\caption{Example of patrolling setting.}
\label{fig:graph_clessidra_single_signal}
\end{figure}
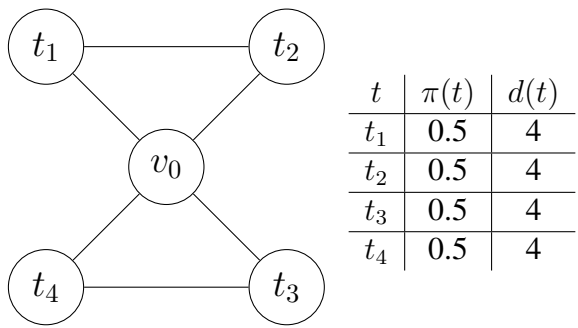
\end{example}

At each turn, an \emph{Attacker} $\mathcal{A}$ and a \emph{Defender} $\mathcal{D}$ play simultaneously having the following available actions:
\begin{itemize}
\item if $\mathcal{A}$ has not attacked in the previous turns, it can observe the position of $\mathcal{D}$ in the graph $G$\footnote{Partial observability of $\mathcal{A}$ over the position of $\mathcal{D}$ can be introduced as discussed in~\cite{DBLP:conf/iat/BasilicoGRCA09}.} and decide whether to attack a target or to wait for a turn. The attack is instantaneous, meaning that there is no delay between the decision to attack and the actual presence of a threat in the selected target\footnote{This is a worst--case assumption according to which $\mathcal{A}$ is as strong as possible. It can be relaxed by associating execution costs to the Attacker's actions as shown in~\cite{BasilicoGatti2009capturing}.};
\item $\mathcal{D}$ has no information about the actions undertaken by $\mathcal{A}$ in previous turns and selects the next vertex to patrol among those adjacent to the current one; each movement is a non--preemptive traversal of a single edge $(v,v') \in E$ and takes one turn to be completed (along the paper, we shall use $\omega^*_{v,v'}$ to denote the temporal cost expressed in turns of the shortest path between any $v$ and $v' \in V$). 
\end{itemize}

The game may conclude in correspondence of any of the two following events. The first one is when $\mathcal{D}$ patrols a target $t$ that is under attack by $\mathcal{A}$ from less than $d(t)$ turns. In such case the attack is prevented and $\mathcal{A}$ is captured. The second one is when target $t$ is attacked and $\mathcal{D}$ does not patrol $t$ during the $d(t)$ turns that follow the beginning of the attack. In such case the attack is successful and $\mathcal{A}$ escapes without being captured.
When $\mathcal{A}$ is captured, $\mathcal{D}$ receives a utility of $1$ and $\mathcal{A}$ receives a utility of $0$. When an attack to $t$ has success, $\mathcal{D}$ receives $1-\pi(t)$ and $\mathcal{A}$ receives $\pi(t)$. The game may not conclude if $\mathcal{A}$ decides to never attack (namely to wait for every turn). In such case, $\mathcal{D}$ receives $1$ and $\mathcal{A}$ receives $0$. Notice that the game is constant sum and therefore it is equivalent to a zero--sum game through an affine transformation.

The above game model is in extensive form (being played sequentially), with imperfect information ($\mathcal{D}$ not observing the actions undertaken by $\mathcal{A}$), and with infinite horizon ($\mathcal{A}$ being in the position to wait forever). The fact that $\mathcal{A}$ can observe the actions undertaken by $\mathcal{D}$ before acting makes the \emph{leader--follower} equilibrium  the natural solution concept for our problem, where $\mathcal{D}$ is the \emph{leader} and $\mathcal{A}$ is the \emph{follower}. Since we focus on zero--sum games, the leader's strategy at the leader--follower equilibrium is its maxmim strategy and it can be found by employing linear mathematical programming, which requires polynomial time in the number of actions available to the players~\cite{leyton2008mas}.

\subsubsection{Introducing alarm signals}
\label{subsubsec:model_model}
We extend the game model described in the previous section by introducing a \emph{spatial uncertain alarm system} that can be exploited by $\mathcal{D}$. The basic idea is that an alarm system uses a number of sensors spread over the environment to gather information about possible attacks and raises an alarm signal at any time an attack occurs. The alarm signal provides some information about the location (target) where the attack is ongoing, but it is affected by uncertainty. In other words, the alarm system detects an attack but it is uncertain about the target under attack. Formally, the alarm system is defined as a pair $(S, p)$, where $S = \{s_1, \cdots, s_m\}$ is a set of $m \ge 1$ {\em signals} and $p: S \times T \rightarrow [0, 1]$ is a function that specifies the probability of having the system generating a signal $s$ given that target $t$ has been attacked. With a slight abuse of notation, for a signal $s$ we define $T(s)=\{t \in T \mid p(s \mid t) > 0\}$  and, similarly, for a target $t$ we have $S(t)=\{s \in S \mid p(s \mid t) > 0\}$. In this work, we assume that:

\begin{itemize}
\item the alarm system is not affected by false positives (signals generated when no attack has occurred). Formally, $p(s \mid \bigtriangleup) = 0$, where $\bigtriangleup$ indicates that no targets are under attack;
\item the alarm system is not affected by false negatives (signals not generated even though an attack has occurred). Formally, $p(\bot \mid t) = 0$, where $\bot$ indicates that no signals have been generated; in Section~\ref{sec:patrolling_game} we will show that the optimal strategies we compute under this assumption can preserve optimality even in presence of non--negligible false negatives rates.
\end{itemize}

\begin{example} 
We report two examples of alarm systems for the patrolling setting depicted in Figure~\ref{fig:graph_clessidra_single_signal}. The first example is reported in Figure~\ref{fig:graph_multiple_signals}(a). It is a low--accuracy alarm system that generates the same  signal anytime each target is under attack and therefore the alarm system does not provide any information about the target under attack. The second example is reported in Figure~\ref{fig:graph_multiple_signals}(b). It provides more accurate information about the localization of the attack than the previous example. Here, each target $t_i$, once attacked, generates an alarm signal $s_i$ with high probability and  a different signal with low probability. That is, if alarm signal $s_i$ has been observed, it is more likely that the attack is in target $t_i$ (given a uniform strategy of $\mathcal{A}$ over the targets).

\begin{figure}[!htbp]
\centering

\subfigure[Alarm system with a single signal for all the targets.]{
\begin{tabular}{c}
\begin{tikzpicture}
  [scale=.8,auto=left,every node/.style={circle,fill=white,draw,minimum size = 1cm,font=\sffamily\large\bfseries}]
  \node (n1) at (0,0) {$v_0$};
  \node (n2) at (-2,2) {$t_1$};
  \node (n3) at (2,2) {$t_2$};
  \node (n4) at (2,-2) {$t_3$};
  \node (n5) at (-2,-2) {$t_4$};

  \path[every node/.style={font=\sffamily\small}]
    (n1) edge node [loop] {} (n2)
	(n1) edge node [loop] {} (n3)
	(n1) edge node [loop] {} (n4)
	(n1) edge node [loop] {} (n5)
	(n2) edge node [loop] {} (n3)
	(n4) edge node [loop] {} (n5);
	
\end{tikzpicture}
\end{tabular}
\begin{tabular}{ c | c | c | c }
    $t$ & $\pi(t)$ & $d(t)$ & $p(s_1 \mid t)$ \\ \hline
    $t_1$ & 0.5 & 4  & 1   \\ \hline
    $t_2$ & 0.5 & 4 & 1  \\ \hline
    $t_3$ & 0.5 & 4 & 1 \\ \hline
    $t_4$ & 0.5 & 4 & 1 
\end{tabular}
}

\subfigure[Alarm system with multiple signals.]{
\begin{tabular}{ c | c | c | c | c | c | c | c }
    $t$ & $\pi(t)$ & $d(t)$ & $p(s_1 \mid t)$ & $p(s_2 \mid t)$& $p(s_3 \mid t)$& $p(s_4 \mid t)$& $p(s_5 \mid t)$\\ \hline
    $t_1$ & 0.5 & 4 & 0.1 & 0.6 & 0.1 & 0.1  & 0.1 \\ \hline
    $t_2$ & 0.5 & 4 & 0.1 & 0.1 & 0.6 & 0.1  & 0.1 \\ \hline
    $t_3$ & 0.5 & 4 & 0.1 & 0.1 & 0.1 & 0.6  & 0.1 \\ \hline
    $t_4$ & 0.5 & 4 & 0.1 & 0.1 & 0.1 & 0.1  & 0.6 
\end{tabular}
}
\caption{Examples of alarm systems.}
\label{fig:graph_multiple_signals}
\end{figure}
\end{example}


Given the presence of an alarm system defined as above, the game mechanism changes in the following way. At each turn, before deciding its next move, $\mathcal{D}$ observes whether or not a signal has been generated by the alarm system and then makes its decision considering such information. This introduces in our game a node of chance implementing the non--deterministic selection of signals, which characterizes the alarm system.

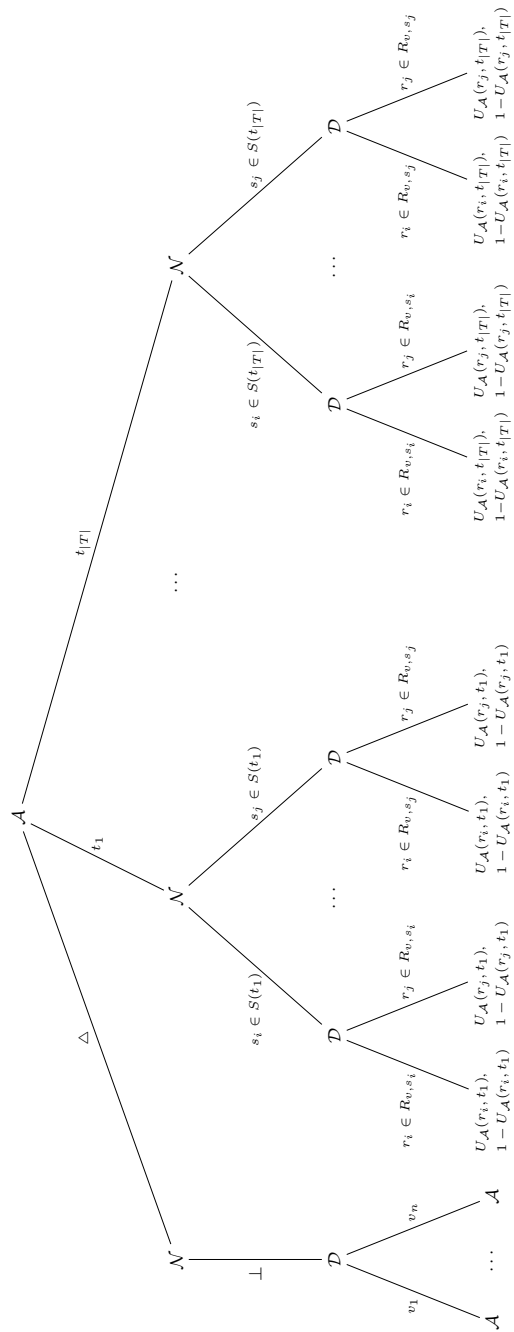
\begin{figure}[!htbp]
\scriptsize
\vspace{-0.7cm}
\begin{center}
\begin{sideways}
\tikzstyle{level 1}=[level distance=2.5cm, sibling distance=2.5cm]
\tikzstyle{level 2}=[level distance=2.5cm, sibling distance=2.2cm]
\tikzstyle{level 3}=[level distance=2.5cm, sibling distance=1cm]
\tikzstyle{level 4}=[level distance=2.5cm, sibling distance=1cm]
\scalebox{0.84}{ 
\begin{tikzpicture}
[noattack/.style={sibling distance=2cm}]
\node {$\mathcal{A}$}
child[noattack]{
	node(x){$\mathcal{N}$}
	child{
		node{$\mathcal{D}$}
		child{
		node{$\mathcal{A}$}
		edge from parent 
		node[above,left]{\tiny{$v_1$}}
		}
		child{node{$\cdots$} edge from parent[draw=none] }
		child{
		node{$\mathcal{A}$}
		edge from parent 
		node[above,right]{\tiny{$v_n$}}
		}
		edge from parent 
		node[above,left]{$\bot$}
	}
	edge from parent 
	node[above]{\tiny{$\bigtriangleup$}}
	}
child{edge from parent[draw=none] }
child{edge from parent[draw=none] }
child{
node(x){$\mathcal{N}$}
	child{
		node(a1){$\mathcal{D}$}
		child{
			node[text width=1.7cm]{\tiny{$U_\mathcal{A}(r_i,t_1)$, $1 - U_\mathcal{A}(r_i,t_1)$}}
			edge from parent 
			node[above,left]{\tiny{$r_i \in R_{v,s_i}$}}
		}
		child{edge from parent[draw=none] }
		child{
			node[text width=1.7cm]{\tiny{$U_\mathcal{A}(r_j,t_1)$, $1 - U_\mathcal{A}(r_j,t_1)$}}
			edge from parent 
			node[above,right]{\tiny{$r_j \in R_{v,s_i}$}}
		}
		edge from parent 
		node[above,left]{\tiny{$s_i \in S(t_1)$}}
	}
	child{node{$\cdots$} edge from parent[draw=none] }
	child{
		node(b1){$\mathcal{D}$}
		child{
			node[text width=1.7cm]{\tiny{$U_\mathcal{A}(r_i,t_1)$, $1 - U_\mathcal{A}(r_i,t_1)$}}
			edge from parent 
			node[above,left]{\tiny{$r_i \in R_{v,s_j}$}}
		}
		child{edge from parent[draw=none] }
		child{
			node[text width=1.7cm]{\tiny{$U_\mathcal{A}(r_j,t_1)$, $1 - U_\mathcal{A}(r_j,t_1)$}}
			edge from parent 
			node[above,right]{\tiny{$r_j \in R_{v,s_j}$}}
		}
		edge from parent 
		node[above,right]{\tiny{$s_j \in S(t_1)$}}
	}
	edge from parent 
	node[above,right]{\tiny{$t_1$}}
}
child{edge from parent[draw=none] }
child{node{$\cdots$} edge from parent[draw=none] }
child{edge from parent[draw=none] }
child{
node(x){$\mathcal{N}$}
	child{
		node(a2){$\mathcal{D}$}
		child{
			node[text width=1.75cm]{\tiny{$U_\mathcal{A}(r_i,t_{|T|})$, $1 - U_\mathcal{A}(r_i,t_{|T|})$}}
			edge from parent 
			node[above,left]{\tiny{$r_i \in R_{v,s_i}$}}
		}
		child{edge from parent[draw=none]}
		child{
			node[text width=1.8cm]{\tiny{$U_\mathcal{A}(r_j,t_{|T|})$, $1 - U_\mathcal{A}(r_j,t_{|T|})$}}
			edge from parent 
			node[above,right]{\tiny{$r_j \in R_{v,s_i}$}}
		}
		edge from parent 
		node[above,left]{\tiny{$s_i \in S(t_{|T|})$}}
	}
	child{node{$\cdots$} edge from parent[draw=none] }
	child{
		node(b2){$\mathcal{D}$}
		child{
			node[text width=1.75cm]{\tiny{$U_\mathcal{A}(r_i,t_{|T|})$, $1 - U_\mathcal{A}(r_i,t_{|T|})$}}
			edge from parent 
			node[above,left]{\tiny{$r_i \in R_{v,s_j}$}}
		}
		child{edge from parent[draw=none] }
		child{
			node[text width=1.8cm]{\tiny{$U_\mathcal{A}(r_j,t_{|T|})$, $1 - U_\mathcal{A}(r_j,t_{|T|})$}}
			edge from parent 
			node[above,right]{\tiny{$r_j \in R_{v,s_j}$}}
		}
		edge from parent 
		node[above,right]{\tiny{$s_j \in S(t_{|T|})$}}
	}
	edge from parent 
	node[above]{\tiny{$t_{|T|}$}}
}
;

\end{tikzpicture}
}
\end{sideways}
\end{center}
\caption[Game tree, $v$ is assumed to the be current vertex for $\mathcal{D}$]{Game tree, $v$ is assumed to the be current vertex for $\mathcal{D}$. $r$ is a collapsed sequence of vertices, called \textit{route}, we shall introduce in the next section.}
\label{fig:tree}
\end{figure}

\subsubsection{The game tree and its decomposition}
\label{subsubsec:game_tree}
Here we depict the game tree of our game model, decomposing it in some recurrent subgames. A portion of the game is depicted in Figure~\ref{fig:tree}. Such tree can be read along the following steps.

\begin{itemize}
\item {\em Root of the tree.} $\mathcal{A}$ decides whether to wait for a turn (this action is denoted by the symbol $\bigtriangleup$ since no target is under attack) or to attack a target $t_i \in T$ (this action is denoted by the label $t_i$ of the target to attack).
\item {\em Second level of the tree.} $\mathcal{N}$ denotes the alarm system, represented by a nature--type player. Its behavior is {\em a priori} specified by the conditional probability mass function $p$ which determines the generated signal given the attack performed by $\mathcal{A}$. In particular, it is useful to distinguish between two cases:
\begin{itemize}
\item[(\emph{a})] if no attack is present, then no signal will be generated under the assumption that $p(\bot \mid \bigtriangleup)=1$;
\item[(\emph{b})] if an attack on target $t_i$ is taking place, a signal $s$ will be drawn from $S(t_i)$ with probability $p(s \mid t_i)$ (recall that we assumed $p(\bot \mid t_i)=0$).
\end{itemize}
\item {\em Third level of the tree.} $\mathcal{D}$ observes the signal raised by the alarm system and decides the next vertex to patrol among those adjacent to the current one (the current vertex is initially chosen by $\mathcal{D}$).
\item {\em Fourth level of the tree and on.} It is useful to distinguish between two cases:
\begin{itemize}
\item[(\emph{a})] if no attack is present, then the tree of the subgame starting from here is the same of the tree of the whole game, except for the position of $\mathcal{D}$ that may be different from the initial one;
\item[(\emph{b})] if an attack is taking place on target $t_i$, then only $\mathcal{D}$ can act.
\end{itemize}
\end{itemize}

Such game tree can be decomposed in a number of finite recurrent subgames such that the best strategies of the agents in each subgame are independent from those in other subgames. This decomposition allows us to apply a \emph{divide et impera} approach, simplifying the resolution of the problem of finding an equilibrium. More precisely, we denote with $\Gamma$ one of these subgames. We define $\Gamma$ as a game subtree that can be extracted from the tree of Figure~\ref{fig:tree} as follows. Given $\mathcal{D}$'s current vertex $v \in V$, select a decision node for $\mathcal{A}$ and call it $i$. Then, extract the subtree rooted in $i$ discarding the branch corresponding to action $\Delta$ (no attack)\footnote{Rigorously speaking, our definition of subgame is not compliant with the definition provided in game theory~\cite{maschler2013game}, which requires that all the actions of a node belong to the same subgame (and therefore we could not separate action $\Delta$ from actions $t_i$). However,  we can slightly change the structure of our game making our definition of subgame compliant with the one from game theory. More precisely, it is sufficient to split each node of $\mathcal{A}$ into two nodes: in the first $\mathcal{A}$ decides to attack a target or to wait for one turn, and in the second, conditioned to the fact that $\mathcal{A}$ decided to attack, $\mathcal{A}$ decides which target to attack. This way, the subtree whose root is the second node of $\mathcal{A}$ is a subgame compliant with game theory. It can be easily observed that this change to the game tree structure does not affect the behaviour of the agents.}. Intuitively, such subgame models the players interaction when the Defender is in some given vertex $v$ and the Attacker will perform an attack.
As a consequence, each $\Gamma$ obtained in such way is finite (once an attack on $t$ started, the maximum length of the game is $d(t)$). Moreover, the set of {\em different} $\Gamma$s we can extract is finite since we have one subgame for each possible current vertex for $\mathcal{D}$, as a consequence we can extract at most $|V|$ different subgames. Notice that, due to the infinite horizon, each subgame can recur an infinite number of times along the game tree. However, being such repetitions independent and the game zero--sum, we only need to solve one subgame to obtain the optimal strategy to be applied in each of its repetitions. In other words, when assuming that an attack will be performed, the agents' strategies can be split in a number of independent strategies solely depending on the current position of the Defender. The reason why we discarded the branch corresponding to action $\Delta$ in each subgame is that we seek to deal with such complementary case exploiting a simple backward induction approach as explained in the following.

First, we call {\em Signal Response Game} from $v$ the subgame $\Gamma$ defined as above and characterized by a vertex $v$ representing the current vertex of $\mathcal{D}$ (for brevity, we shall use SRG--$v$). In an SRG--$v$, the goal of $\mathcal{D}$ is to find the best strategy starting from vertex~$v$ to respond to any alarm signal. All the SRG--$v$s are independent one each other and thus the best strategy in each subgame does not depend on the strategies of the other subgames. The intuition is that the best strategies in an SRG--$v$ does not depend on the vertices visited by $\mathcal{D}$ before the attack. Given an SRG--$v$, we denote by $\sigma^{\mathcal{D}}_{v,s}$ the strategy of $\mathcal{D}$ once observed signal $s$, by $\sigma^{\mathcal{D}}_{v}$ the strategy profile $\sigma^{\mathcal{D}}_{v}= (\sigma^{\mathcal{D}}_{v,s_1}\ldots, \sigma^{\mathcal{D}}_{v,s_m})$ of $\mathcal{D}$, and by $\sigma^{\mathcal{A}}_{v}$ the strategy of $\mathcal{A}$. Let us notice that in an SRG--$v$, given a signal $s$, $\mathcal{D}$ is the only agent that plays and therefore each sequence of moves between vertices of $\mathcal{D}$ in the tree can be collapsed in a single action. Thus, SRG--$v$ is essentially a two--level game in which $\mathcal{A}$ decides the target to attack and $\mathcal{D}$ decides  the sequence of moves to visit the targets.

Then, according to classical backward induction arguments~\cite{maschler2013game}, once we have found the best strategies of each SRG--$v$, we can substitute the subgames with the agents' equilibrium utilities and then we can find the best strategy of $\mathcal{D}$ for patrolling the vertices whenever no alarm signal has been raised and the best strategy of attack for $\mathcal{A}$. We call this last problem \emph{Patrolling Game} (for conciseness, we shall use PG). We denote by $\sigma^{\mathcal{D}}$ and $\sigma^{\mathcal{A}}$ the strategies of $\mathcal{D}$ and $\mathcal{A}$ respectively in the PG.

\subsection{The computational questions we pose}
\label{subsec:questions}
In the present paper, we focus on some questions whose answers play a fundamental role in the design of the best algorithms to find an equilibrium of our game. More precisely, we investigate the computational complexity of the following four problems. The first problem concerns the PG.

\begin{question}
Which is the best patrolling strategy for $\mathcal{D}$ maximizing its expected utility?
\end{question}

The other three questions concern SRG--$v$. For the sake of simplicity, we focus on the case in which there is only one signal $s$, we shall show that it is possible to scale linearly in the number of signals. 

\begin{question}
Given a starting vertex $v$ and a signal $s$, is there any strategy of $\mathcal{D}$ that allows $\mathcal{D}$ to visit all the targets in $T(s)$, each within its deadline?
\end{question}

\begin{question}
Given a starting vertex $v$ and a signal $s$, is there any pure strategy of $\mathcal{D}$ giving $\mathcal{D}$ an expected utility of at least $k$?
\end{question}

\begin{question}
Given a starting vertex $v$ and a signal $s$, is there any mixed strategy of $\mathcal{D}$ giving $\mathcal{D}$ an expected utility of at least $k$?
\end{question}

In the following, we shall take a bottom--up approach answering the above questions starting from the last three and then dealing with the first one at the whole--game level.
\section{Signal response game on arbitrary graphs}
\label{sec:srg_on_arbitrary_graphs}
In this section we show how to deal with an SRG--$v$ on arbitrary graphs. Specifically, in Section~\ref{subsec:complexity} we prove the hardness of the problem, analyzing its computational complexity. Then, in Section~\ref{subsec:alg_dyn_prog} and in Section~\ref{subsec:alg_b_and_b} we propose two algorithms, the first based on \textit{dynamic programming} (breadth--first search) while the second adopts a \textit{branch and bound} (depth--first search) paradigm. Furthermore, we provide a variation for each algorithm, approximating the optimal solution.

\subsection{Complexity results}
\label{subsec:complexity}
In this section we analyse SRG--$v$ from a computational point of view. We initially observe that  each SRG--$v$ is characterized by $|T|$ actions available to $\mathcal{A}$, each corresponding to a target $t$, and by $O(|V|^{\max_t\{d(t)\}})$ decision nodes of $\mathcal{D}$. The portion of game tree played by $\mathcal{D}$ can be safely reduced by observing that $\mathcal{D}$ will move between any two targets along the minimum path. This allows us to discard from the tree all the decision nodes where $\mathcal{D}$ occupies a non--target vertex. However, this reduction keeps the size of the game tree exponential in the parameters of the game, specifically $O(|T|^{|T|})$.\footnote{A more accurate bound is $O(|T|^{\min\{|T|,\max_t\{d(t)\}\}})$.} The exponential size of the game tree does not constitute a proof that finding the equilibrium strategies of an SRG--$v$ requires exponential time in the parameters of the game because it does not exclude the existence of some compact representation of $\mathcal{D}$'s strategies, e.g., Markovian strategies. Indeed such representation should be polynomially upper bounded in the parameters of the game and therefore they would allow the design of a polynomial--time algorithm to find an equilibrium. We show below that it is unlikely that such a representation exists in arbitrary graphs, while it exists for special topologies as we shall discuss later.

We denote by $g_v$ the expected utility of $\mathcal{A}$ from SRG--$v$ and therefore the expected utility of $\mathcal{D}$ is $1-g_v$. Then, we define the following problem.


\begin{definition}[$k$--SRG--$v$] The decision problem $k$--SRG--$v$ is defined as:

\noindent INSTANCE: an instance of SRG--$v$;

\noindent QUESTION: is there any $\sigma^{\mathcal{D}}$ such that $g_{v} \le k$ (when $\mathcal{A}$ plays its best response)?
\end{definition}

\begin{theorem}\label{thm:SRG-npc}
k--SRG--$v$ is strongly $\mathcal{NP}$--hard even when $|S|=1$.
\end{theorem}

\noindent
\textit{Proof.} Let us consider the following reduction from HAMILTONIAN--PATH~\cite{garey1990computers}.
Given an instance of HAMILTONIAN--PATH $G_H = (V_H, E_H)$, we build an instance for k--SRG--$v$ as:
\begin{itemize}
\item$V = V_H \cup \{v\}$; 
\item$E = E_H \cup \{(v, h), \forall h \in V_H\}$;
\item$T = V_H$; 
\item$d ( t ) = |V_H|$; 
\item$\pi(t) \in (0, 1]$, for all $t \in T$ (any value);
\item$S = \{s_1\}$; 
\item $T(s_1)=T$;
\item$p(s_1 \mid t) = 1$, for all $t \in T$; 
\item$k=0$.
\end{itemize}

If $g_{s} \le 0$, then there must exist a path starting from $v$ and visiting all the targets in $T$ by $d=|V_H|$. Given the penetration times assigned in the above construction and recalling that edge costs are unitary, the path must visit each target exactly once. Therefore, since $T=V_H$, the game's value is less than or equal to zero if and only if $G_H$ admits a Hamiltonian path. This concludes the proof.\hfill$\Box$

Notice that the problem of assessing the membership of $k$--SRG--$v$ to $\mathcal{NP}$ is left open and it strictly depends on the size of the support of the strategy of $\sigma^{D}_v$. That is, if any strategy $\sigma^{D}_v$ has a polynomially upper bounded support, then $k$--SRG--$v$ is in $\mathcal{NP}$. We conjecture it is not and therefore there can be optimal strategies in which an exponential number of actions are played with strictly positive probability. Furthermore, the above result shows that with arbitrary graphs:
\begin{itemize}
\item answering to Question~1 is $\mathcal{FNP}$--hard,
\item answering to Questions~2, 3, 4 is $\mathcal{NP}$--hard.
\end{itemize}
As a consequence a polynomial--time algorithm solving those problems is unlikely to exist. In particular, the above proof shows that we cannot produce a compact representation (a.k.a. information lossless abstractions) of the space of strategies of $\mathcal{D}$ that is smaller than $O(2^{|T(s)|})$, unless there is an algorithm better than the best--known algorithm for HAMILTONIAN--PATH. This is due to the fact that the best pure maxmin strategy can be found in linear time in the number of the pure strategies and the above proof shows that it cannot be done in a time less than $O(2^{|T(s)|})$. More generally, no polynomially bounded representation of the space of the strategies can exist, unless $\mathcal{P} = \mathcal{NP}$. 

Although we can deal only with exponentially large representations of $\mathcal{D}$'s strategies, we focus on the problem of finding the most efficient representation. Initially, we provide the following definitions.
\begin{definition}[Route]
Given a starting vertex $v$ and a signal $s$, a route (over the targets) $r$ is a generic sequence of targets of $T(s)$ such that:  
\begin{itemize}
\item  $r$ starts from $v$,
\item each target $t\in T(s)$ occurs at most once in $r$ (but some targets may not occur),
\item $r(i)$ is the $i$--th visited target in $r$ (in addition, $r(0)=v$).
\end{itemize}
\end{definition}
Among all the possible routes we restrict our attention on a special class of routes that we call \emph{covering} and are defined as follows.
\begin{definition}[Covering Route]
Given a starting vertex $v$ and a signal $s$, a route $r$ is covering when, denoted by $A_{r}(r(i))  = \sum_{h=0}^{i-1}\omega^*_{r(h),r(h+1)}$ the time needed by $\mathcal{D}$ to visit target $t=r(i)\in T(s)$ starting from $r(0)=v$ and moving along the shortest path between each pair of consecutive targets in $r$, for every target $t$ occurring in $r$ it holds $A_r(r(i)) \leq d(r(i))$ (i.e., all the targets are visited within their penetration times).
\end{definition}

With a slight abuse of notation, we denote by $T(r)$ the set of targets covered by $r$ and we denote by $c(r)$ the total temporal cost of $r$, i.e., $c(r) = A_{r}(r(|T(r)|))$. Notice that in the worst case the number of covering routes  is $O(|T(s)|^{|T(s)|})$, but computing all of them may be unnecessary since some covering routes will never be played by $\mathcal{D}$ due to strategy domination and therefore they can be safely discarded~\cite{osborne2004introduction}. More precisely, we introduce the following two forms of dominance.

\begin{definition}[Intra--Set Dominance]\label{def:intradom}
Given a starting vertex $v$, a signal $s$ and two different covering routes $r, r'$ such that $T(r)=T(r')$, if $c(r) \le c(r')$ then $r$ dominates $r'$.
\end{definition}

\begin{definition}[Inter--Set Dominance]\label{def:interdom}
Given a starting vertex $v$, a signal $s$ and two different covering routes $r, r'$, if $T(r) \supset T(r')$ then $r$ dominates $r'$.
\end{definition}
 
Furthermore, it is convenient to introduce the concept of \emph{covering set}, which is strictly related to the concept of covering route. It is defined as follows.
\begin{definition}[Covering Set]
Given a starting vertex $v$ and a signal $s$, a covering set $Q$ is a subset of targets $T(s)$ such that there exists a covering route $r$ with $T(r)=Q$.
\end{definition}

Let us focus on Definition~\ref{def:intradom}. It suggests that we can safely use only one route per covering set. Covering sets suffice for describing all the outcomes of the game, since the agents' payoffs depend only on the fact that $\mathcal{A}$ attacks a target $t$ that is covered by $\mathcal{D}$ or not, and in the worst case are $O(2^{|T(s)|})$, with a remarkable reduction of the search space w.r.t. $O(|T(s)|^{|T(s)|})$. However, any algorithm restricting on covering sets should be able to determine whether or not a set of targets is a covering one. Unfortunately, this problem is hard too.

\begin{definition} [COV--SET] The decision problem COV--SET is defined as:

\noindent INSTANCE: an instance of SRG--$v$ with a target set $T$;

\noindent QUESTION: is $T$ a covering set for some covering route $r$?
\end{definition}

By trivially adapting the same reduction for Theorem~\ref{thm:SRG-npc} we can state the following theorem.

\begin{theorem}\label{thm:covsetnpc}
COV--SET is $\mathcal{NP}$--complete.
\end{theorem}

Therefore, computing a covering route for a given set of targets (or deciding that no covering route exists) is not doable in polynomial time unless $\mathcal{P} = \mathcal{NP}$. This shows that, while covering sets suffice for defining the payoffs of the game and therefore the size of payoffs matrix can be bounded by the number of covering sets, in practice we also need covering routes to certificate that a given subset of targets is covering. Thus, we need to work with covering routes, but we just need the routes corresponding to the covering sets, limiting the number of covering routes that are useful for the game to the number of covering sets. In addition, Theorem~\ref{thm:covsetnpc} suggests that no algorithm for COV--SET can have complexity better than $O(2^{|T(s)|})$ unless there exists a better algorithm for HAMILTONIAN--PATH than the best algorithm known in the literature.  This seems to suggest that enumerating all the possible subsets of targets (corresponding to all the potential covering sets) and, for each of them, checking whether or not it is covering requires a complexity worse than $O(2^{|T(s)|})$. Surprisingly, we show in the next section that there is an algorithm with complexity $O(2^{|T(s)|})$  (neglecting polynomial terms) to enumerate all and only the covering sets and, for each of them, one covering route. Therefore, the complexity of our algorithm matches (neglecting polynomial terms) the complexity of the best--known algorithm for HAMILTONIAN--PATH.

Let us focus on Definition~\ref{def:interdom}.  Inter--Set dominance can be leveraged to introduce the concept of {\em maximal} covering sets which could enable a further reduction in the set of actions available to $\mathcal{D}$.

\begin{definition} [MAXIMAL COV--SET] Given a covering set $Q$ (where $Q=T(r)$ for some $r$), we say that $Q$ is \emph{maximal} if there is no route $r'$ such that $Q\subset T(r')$. 
\end{definition}

Furthermore, we say that $r$ such that $T(r)=Q$ is a maximal covering route. In the best case, when there is a route covering all the targets, the number of maximal covering sets is 1, while the number of covering sets (including the non--maximal ones) is $2^{|T(s)|}$. Thus, considering only maximal covering sets allows an exponential reduction of the payoffs matrix.
In the worst case, when all the possible subsets composed of $|T(s)|/2$ targets are maximal covering sets, the number of maximal covering sets is $O(2^{|T(s)|-2})$, while the number of covering sets is $O(2^{|T(s)|-1})$, allowing a reduction of the payoffs matrix by a factor of 2. Furthermore, if we knew {\em a priori} that $Q$ is a maximal covering set, we could avoid searching for covering routes for any set of targets that strictly contains $Q$. When designing  an algorithm to solve this problem, Definition~\ref{def:interdom} could then be exploited to introduce some kind of pruning technique to save average compute time. However, the following result shows that deciding if a covering set is maximal is hard.

\begin{definition}[MAX--COV--SET] The decision problem MAX--COV--SET is defined as:

\noindent INSTANCE: an instance of SRG--$v$ with a target set $T$ and a covering set $T' \subset T$;

\noindent QUESTION: is $T'$ maximal?
\end{definition}

\begin{theorem}
There is no polynomial--time algorithm for MAX--COV--SET unless $\mathcal{P}=\mathcal{NP}$.
\end{theorem}
\textit{Proof.} Assume for simplicity that $S=\{s_1\}$ and that $T(s_1)=T$. Initially, we observe that MAX--COV--SET is in co--$\mathcal{NP}$. Indeed, any covering route $r$ such that $T(r) \supset T'$ is a NO certificate for MAX--COV--SET, placing it in co--$\mathcal{NP}$. (Notice that, trivially, any covering route has length bounded by $O(|T|^2)$; also, notice that due to Theorem~2, having a covering set  would not suffice given that we cannot verify in polynomial time whether it is actually covering unless $\mathcal{P}=\mathcal{NP}$.) 

Let us suppose we have a polynomial--time algorithm for MAX--COV--SET, called $A$. Then (since $\mathcal{P} \subseteq \mathcal{NP} \cap \text{co--}\mathcal{NP}$) we have a polynomial algorithm for the complement problem, i.e., deciding whether all the covering routes for $T'$ are dominated. Let us consider the following algorithm: given an instance for COV--SET specified by graph $G=(V, E)$, a set of target $T$ with penetration times $d$, and a starting vertex $v$:
\begin{enumerate}
\item assign to targets in $T$ a lexicographic order $t_1, t_2, \ldots, t_{|T|}$;
\item for every $t \in T$, verify if $\{t\}$ is a covering set in $O(|T|)$ time by comparing $\omega^*_{v,t}$ and $d(t)$; if at least one is not a covering set, then output NO and terminate; otherwise set $\hat{T} = \{t_1\}$ and $k=2$;
\item apply algorithm $A$ on the following instance: graph $G=(V, E)$, target set $\{\hat{T} \cup \{t_k\} ,\hat{d}\}$ (where $\hat{d}$ is $d$ restricted to $\hat{T} \cup \{t_k\}$), start vertex $v$, and covering set $\hat{T}$;
\item if $A$'s output is YES (that is, $\hat{T}$ is not maximal) then set $\hat{T} = \hat{T} \cup \{t_k\}$, $k=k+1$ and restart from step 3; if $A$'s output is NO and $k=|T|$ then output YES; if $A$'s output is NO and $k<|T|$ then output NO;
\end{enumerate}
Thus, the existence of $A$ would imply the existence of a polynomial algorithm for COV--SET which (under $\mathcal{P} \neq \mathcal{NP}$) would contradict Theorem~2. This concludes the proof.\hfill$\Box$

Nevertheless, we show hereafter that there exists an algorithm enumerating all and only the maximal covering sets and one route for each of them (which potentially leads to an exponential reduction of the time needed for solving the linear program) with only an additional polynomial cost w.r.t. the enumeration of all the covering sets. Therefore, neglecting polynomial terms, our algorithm has a complexity of $O(2^{|T(s)|})$.

Finally, we focus on the complexity of approximating the best solution in an SRG--$v$. When $\mathcal{D}$ restricts its strategies to be pure, the problem is clearly not approximable in polynomial time even when the approximation ratio depends on $|T(s)|$. The basic intuition is that, if a game instance admits the maximal covering route that covers all the targets and the value of all the targets is 1, then either the maximal covering route is played returning a utility of 1 to $\mathcal{D}$ or any other route is played returning a utility of $0$, but no polynomial--time algorithm can find the maximal covering route covering all the targets, unless $\mathcal{P}=\mathcal{NP}$. On the other hand, it is interesting to investigate the case in which no restriction to pure strategies is present. We show that the problem keeps being hard. 
\begin{theorem}
The optimization version of $k$--SRG--$v$, say OPT--SRG--$v$, is $\mathcal{APX}$--hard even in the restricted case in which the graph is metric, there is only one signal~$s$, all targets $t\in T(s)$ have the same penetration time $d(t)$, and there is the maximal covering route covering all the targets.
\end{theorem}
\emph{Proof}. We produce an approximation--preserving reduction from TSP(1,2) that is known to be $\mathcal{APX}$--hard~\cite{papadimitriou1993tsp12}. For the sake of clarity, we divide the proof in steps.

\emph{TSP(1,2) instance}. An instance of undirected TSP(1,2) is defined as follows: 
\begin{itemize}
\item a set of vertices $V_{TSP}$,
\item a set of edges composed of an edge per pair of vertices,
\item a symmetric matrix $C_{TSP}$ of weights, whose values can be 1 or 2, each associated with an edge and representing the cost of the shortest path between the corresponding pair of vertices.
\end{itemize}
The goal is to find the minimum cost tour. Let us denote by $OPTSOL_{TSP}$ and $OPT_{TSP}$ the optimal solution of TSP(1,2) and its cost, respectively. Furthermore, let us denote by $APXSOL_{TSP}$ and $APX_{TSP}$ an approximate  solution of TSP(1,2) and its cost, respectively. It is known that there is no polynomial--time approximation algorithm with $APX_{TSP}/OPT_{APX}<\alpha$ for some $\alpha > 1$, unless $\mathcal{P}=\mathcal{NP}$~\cite{papadimitriou1993tsp12}. 

\emph{Reduction}. We map an instance of TSP(1,2) to a specific instance of SRG--$v$ as follows:
\begin{itemize}
\item there is only one signal $s$,
\item $T(s)=V_{TSP}$,
\item $w^*_{t,t'} = C_{TSP}(t,t')$, for every $t,t'\in T(s)$,
\item $\pi(t)=1$, for every $t\in T(s)$,
\item $w^*_{v,t}=1$, for every $t\in T(s)$,
\item $d(t)=\begin{cases} OPT_{TSP} & \textnormal{if $OPT_{TSP}=|V_{TSP}|$} \\ OPT_{TSP}-1 & \textnormal{if $OPT_{TSP}>|V_{TSP}|$}\end{cases}$, for every $t\in T(s)$.
\end{itemize}
In this reduction, we use the value of $OPT_{TSP}$ even if there is no polynomial--time algorithm solving exactly TSP(1,2), unless $\mathcal{P}=\mathcal{NP}$. We show below that with an additional polynomial--time effort we can deal with the lack of knowledge of $OPT_{TSP}$.

\emph{OPT--SRG--$v$ optimal solution}. By construction of the SRG--$v$ instance, there is a covering route starting from $v$ and visiting all the targets $t\in T(s)$, each within its penetration time. This route has a cost of exactly $d(t)$ and it is $\langle v, t_1, \ldots, t_{|T(s)|}\rangle$, where $\langle t_1, \ldots, t_{|T(s)|},t_1\rangle$ corresponds to $OPTSOL_{TSP}$ with the constraint that $w^*_{t_{|T(s)|},t_1}=2$ if $OPT_{TSP}>|V_{TSP}|$ (essentially, we transform the tour in a path by discarding one of the edges with the largest cost). Therefore, the optimal solution of SRG--$v$, say $OPTSOL_{SRG}$, prescribes to play the maximal route with probability one and the optimal value, say $OPT_{SRG}$, is 1. 

\emph{OPT--SRG--$v$ approximation}. Let us denote by $APXSOL_{SRG}$ and $APX_{SRG}$ an approximate solution of OPT--SRG--$v$ and its value, respectively. We assume there is a polynomial--time approximation algorithm with $APX_{SRG}/OPT_{SRG}\geq \beta$ where $\beta \in (0,1)$. Let us notice that $APXSOL_{SRG}$ prescribes to play a polynomially upper bounded number of covering routes with strictly positive probability. We introduce a lemma that characterizes such covering routes.

\begin{lemma}
The longest covering route played with strictly positive probability in $APXSOL_{SRG}$ visits at least $\beta |T(s)|$ targets.
\end{lemma}

\emph{Proof}. Assume by contradiction that the longest route visits $\beta |T(s)|-1$ targets. The best case in terms of maximization of the value of OPT--SRG--$v$ is, due to reasons of symmetry (all the targets have the same value), when there is a set of $|T(s)|$ covering routes of length $\beta |T(s)|-1$ such that each target is visited exactly by $\beta |T(s)|-1$ routes. When these routes are available, the best strategy is to randomize uniformly over the routes. The probability that a target is covered is $\beta - \frac{1}{|T(s)|}$ and therefore the value of $APX_{SRG}$ is $\beta - \frac{1}{|T(s)|}$. This leads to a contradiction, since the algorithm would provide an approximation strictly smaller than $\beta$.
\hfill$\Box$

\emph{TSP(1,2) approximation from OPT--SRG--$v$ approximation}. We use the above lemma to show that we can build a $(3-2\beta)$--approximation for TSP(1,2) from a $\beta$--approximation of OPT--SRG--$v$. Given an $APXSOL_{SRG}$, we extract the longest covering route played with strictly positive probability, say $\langle v, t_{1}, \ldots, t_{\beta |T(s)|}\rangle$. The route has a cost of at most $d(t)$, it would not cover $\beta |T(s)|$ targets otherwise. Any tour $\langle t_{1}, \ldots, t_{\beta |T(s)|}, t_{\beta |T(s)|+1},\ldots, t_{|T(s)|},t_{1}\rangle$ has a cost not larger than $d(t)-1+2(1-\beta)|T(s)|= OPT_{TSP} - 1 + 2(1-\beta)|V_{TSP}|$ (under the worst case in which all the edges in $\langle t_{\beta |T(s)|}, t_{\beta |T(s)|+1},\ldots, t_{|T(s)|},t_1\rangle$ have a cost of 2). Given that $OPT_{TSP}\geq |V_{TSP}|$, we have that such a tour has a cost not larger than $OPT_{TSP} - 1 + 2(1-\beta)|V_{TSP}|\leq OPT_{TSP} (3-2\beta)$. Therefore, the tour is a $(3-2\beta)$--approximation for TSP(1,2). Since TSP(1,2) is not approximable in polynomial time for any approximation ratio smaller than $\alpha$, we have the constraint that $3-2\beta\geq \alpha$, and therefore that $\beta \leq \frac{3-\alpha}{2}$. Since $\alpha>1$, we have that $\frac{3-\alpha}{2}<1$ and therefore that there is no polynomial--time approximation algorithm for OPT--SRG--$v$ when $\beta \in (\frac{3-\alpha}{2},1)$, unless $\mathcal{P}=\mathcal{NP}$.

\emph{$OPT_{TSP}$ oracle}. In order to deal with the fact that we do not know $OPT_{TSP}$, we can execute the approximation algorithm for OPT--SRG--$v$ using a guess over $OPT_{TSP}$. More precisely, we execute the approximation algorithm for every value in $\{|V_{TSP}|,\ldots, 2|V_{TSP}|\}$ and we return the best approximation found for TSP(1,2). Given that $OPT_{TSP}\in \{|V_{TSP}|,\ldots ,2|V_{TSP}|\}$, there is an execution of the approximation algorithm that uses the correct guess. 
\hfill$\Box$

We report some remarks to the above theorem.

\begin{remark}
The above result does not exclude the existence of constant--ratio approximation algorithms for OPT--SRG--$v$. We conjecture that it is unlikely. OPT--SRG--$v$ presents similarities with the (metric) DEADLINE--TSP, where the goal is to find the longest path of vertices each traversed before its deadline. The DEADLINE--TSP does not admit any constant--ratio approximation algorithm~\cite{bockenhauer07apxtsp} and the best--known approximation algorithm has logarithmic approximation ratio~\cite{bansal2004apxdeadlinetsp}. The following observations can be produced about the relationships between OPT--SRG--$v$ and DEADLINE--TSP:
\begin{itemize}
\item when the maximal route covering all the targets in the OPT--SRG--$v$ exists, the optimal solution of the OPT--SRG--$v$ is also optimal for the DEADLINE--TSP applied to the same graph;
\item when the maximal route covering all the targets in the OPT--SRG--$v$ does not exist, the optimal solutions of the two problems are different, even when we restrict us to pure--strategy solutions for the OPT--SRG--$v$;
\item approximating the optimal solution of the DEADLINE--TSP does not give a direct technique to approximate OPT--SRG--$v$, since we should enumerate all the subsets of targets and for each subset of targets we would need to execute the approximation of the DEADLINE--TSP, but this would require exponential time. We notice in addition that even the total number of sets of targets with logarithmic size is not polynomial, being $\Omega(2^{\log^2(|T|)})$, and therefore any algorithm enumerating them would require exponential time;
\item when the optimal solution of the OPT--SRG--$v$ is randomized, examples of optimal solutions in which maximal covering routes are not played can be produced, showing that at the optimum it is not strictly necessary to play maximal covering routes, but even approximations suffice.
\end{itemize}
\end{remark}

\begin{remark}
If it is possible to map DEADLINE--TSP instances to OPT--SRG--$v$ instances where the maximal covering route covering all the targets exists, then it trivially follows that OPT--SRG--$v$ does not admit any constant--approximation ratio. We were not able to find such a mapping and we conjecture that, if there is an approximation--preserving reduction from DEADLINE--TSP to OPT--SRG--$v$, then we cannot restrict to such instances. The study of instances of OPT--SRG--$v$ where mixed strategies may be optimal make the treatment very challenging.
\end{remark}

\subsection{Dynamic--programming algorithm}
\label{subsec:alg_dyn_prog}
We start by presenting two algorithms. The first one is exact, while the second one is an approximation algorithm. Both algorithms are based on a dynamic programming approach.

\subsubsection{Exact algorithm}
\label{subsec:alg_dyn_progexact}
Here we provide an algorithm based on the dynamic programming paradigm returning the set  of strategies available to $\mathcal{D}$ when it is in $v$ and receives a signal $s$. 
The algorithm we present in this section enumerates all the covering sets and, for each of them, it returns also the corresponding covering route. Initially, we observe that we can safely restrict our attention to a specific class of covering sets, that we call \emph{proper}, defined as follows.

\begin{definition}[Proper Covering Set]\label{def:propcovset}
Given a starting vertex $v$ and a signal $s$, a covering set $Q$ is proper if there is a route $r$ such that, once walked (along the shortest paths) over graph $G$, it does not traverse any target $t\in T(s)\setminus T(r)$.
\end{definition}

While in the worst case the number of proper covering sets is  equal to the number of covering sets (consider, for example, fully connected graphs with unitary edge costs) in realistic scenarios we expect that the number of proper covering sets is much smaller than the number of covering sets. As we show in Section~5, restricting to proper covering sets makes the complexity of our algorithm polynomial with respect to some special topologies: differently from the number of covering sets, the number of proper covering sets is polynomially upper bounded. Hereafter we provide the description of the algorithm.

Let us denote $C_{v,t}^k$ a collection of proper covering sets, where each set in this collection is denoted as $Q_{v,t}^k$. The proper covering set $Q_{v,t}^k$ has cardinality $k$ and admits a covering route $r$ whose starting vertex is $v$ and whose last covered target is $t$. Each $Q_{v,t}^k$ is associated with a cost $c(Q_{v,t}^k)$ representing the temporal cost of the shortest covering route for $Q_{v,t}^k$ that specifies $t$ as the $k$--th target to visit. Upon this basic structure, our algorithm iteratively computes proper covering sets collections and costs for increasing cardinalities, that is from $k=1$ possibly up to $k = |T(s)|$ including one target at each iteration. Using a dynamic programming approach, we assume to have solved up to cardinality $k-1$ and we specify how to complete the task for cardinality $k$. Detailed steps are reported in Algorithm~\ref{alg:dp}, while in the following we provide an intuitive description. Given $Q_{v,t}^{k-1}$, we can compute a set of targets $Q^+$ (Line~\ref{alg:Splus_intersect}) that is a subset of $T(s)$ such that for each target $t' \in Q^+$ the following properties hold:

\begin{itemize}
\item $t'\not \in Q_{v,t}^{k-1}$,
\item if $t'$ is appended to the shortest covering route for $Q_{v,t}^{k-1}$, it will be visited before $d(t')$,
\item the shortest path between $t$ and $t'$ does not traverse any target $t''\in T(s)\setminus Q_{v,t}^{k-1}$.
\end{itemize}

Function $ShortestPath(G,t,t')$ returns the shortest path on $G$ between $t$ and $t'$. For efficiency, we calculate (in polynomial time) all the shortest paths offline by means of the Floyd--Warshall algorithm~\cite{FM}. If $Q^+$ is not empty, for each $t' \in Q^+$, we extend $Q_{v,t}^{k-1}$ (Step~\ref{alg:extend}) by including it and naming the resulting covering set as $Q_{v,t'}^k$ since it has cardinality $k$ and we know it admits a covering route with last vertex $t'$. Such route can be obtained by appending $t'$ to the covering route for $Q_{v,t}^{k-1}$ and has cost $c(Q_{v,t}^{k-1}) + \omega^*_{t,t'}$. This value is assumed to be the cost of the extended proper covering set.---In Step~\ref{alg:search}, we make use of a procedure called $Search(Q,C)$ where $Q$ is a covering set and $C$ is a collection of covering sets. The procedure outputs $Q$ if $Q \in C$ and $\emptyset$ otherwise. We adopted an efficient implementation of such procedure which can run in $O(|T(s)|)$. More precisely, we represent a covering set $Q$ as a binary vector of length $|T(s)|$ where the $i$--th component is set to $1$ if target $t_i \in Q$ and $0$ otherwise. A collection of covering sets $C$ can then be represented as a binary tree with depth $|T(s)| $. The membership of a covering set $Q$ to collection $C$ is represented with a branch of the tree in such a way that if $t_i \in Q$ then we have a left edge at depth $i-1$ on such branch. We can easily determine if $Q \in C$ by checking if traversing a left (right) edge in the tree each time we read a $1$ ($0$) in $Q$'s binary vector we reach a leaf node at depth $|T(s)|$. The insertion of a new covering set in the collection can be done in the same way by traversing existing edges and expanding the tree where necessary.---If such extended proper covering set is not present in collection $C_{v,t'}^k$ or is already present with a higher cost (Step~\ref{alg:ifcost}), then collection and cost are updated (Steps~\ref{alg:updateCollection} and~\ref{alg:updateCost}).  After the iteration for cardinality $k$ is completed, for each proper covering set $Q$ in collection $C_{v,t}^k$, $c(Q)$ represents the temporal cost of the shortest covering route with $t$ as last target.

\begin{algorithm}\caption{DP--ComputeCovSets($v,s$)}\label{alg:dp}
\begin{scriptsize}
\begin{algorithmic}[1]


\State $\forall t \in T(s), k \in \{2, \ldots, |T(s)|\}$: $C_{v,t}^1 = \begin{cases} \{t\} & \textnormal{if $(v,t)\in E_R$} \\ \emptyset & \textnormal{otherwise} \end{cases}$, $C_{v,t}^k = \emptyset$

\State $\forall t \in T(s)$: $c(\{t\}) = \begin{cases} \omega^*_{v,t}  & \textnormal{if $C_{v,t}^1\neq \emptyset$}  \\ \infty & \textnormal{otherwise}\end{cases}$, $c(\emptyset) = \infty$

\ForAll{$k \in \{2 \ldots |T(s)|\}$}

	\ForAll{$t \in T(s)$} \label{alg:everyt}

		\ForAll{$Q_{v,t}^{k-1} \in C_{v,t}^{k-1}$}
		

		\State $Q^+ = \left \{t' \in T(s) \setminus Q_{v,t}^{k-1} :  \left(c(Q_{v,t}^{k-1}) + \omega^*_{t,t'} \le d(t') \right)\wedge \Big(\not \exists t'' \in T(s)\setminus Q: t''\in ShortestPath(G,t,t')\Big) \right\}$\label{alg:Splus_intersect}

			\ForAll{$t' \in Q^+$}

			\State $Q_{v,t'}^k = Q_{v,t}^{k-1} \cup \{t'\}$ \label{alg:extend}

			\State $U = Search(Q_{v,t'}^k,C_{v,t'}^k)$ \label{alg:search}

			\If {$c(U) > c(Q_{v,t}^{k-1}) + \omega^*_{t,t'}$} \label{alg:ifcost}
    				
				\State $C_{v,t'}^k = C_{v,t'}^k \cup \{Q_{v,t'}^k\}$ \label{alg:updateCollection}

				\State $c(Q_{v,t'}^k) = c(Q_{v,t}^{k-1}) + \omega^*_{t,t'}$ \label{alg:updateCost}
    
    			\EndIf				
			\EndFor
		\EndFor
	\EndFor
\EndFor

\State \textbf{return} $\{C_{v,t}^k: t\in T(s), k\leq |T(s)| \}$
\end{algorithmic}
\end{scriptsize}
\end{algorithm}

After Algorithm~\ref{alg:dp} completed its execution, for any arbitrary $T' \subseteq T$ we can easily obtain the temporal cost of its shortest covering route as 
\[
c^*(T') = \min_{Q \in Y_{|T'|}} c(Q)
\] 
where $Y_{|T'|}=\cup_{t \in T} \{Search(T',C^{|T'|}_{v,t}) \}$ (notice that if $T'$ is not a covering set then $c^*(T')=\infty$). For the sake of simplicity, Algorithm~\ref{alg:dp} does not specify how to carry out two sub--tasks we describe in the following.

The first one is the {\em annotation of dominated (proper) covering sets}. Each time Steps~\ref{alg:updateCollection} and~\ref{alg:updateCost} are executed, a covering set is added to some collection. Let us call it $Q$ and assume it has cardinality $k$. Each time a new $Q$ has to be included at cardinality $k$, we mark all the covering sets at cardinality $k-1$ that are dominated by $Q$ (as per Definition~\ref{def:interdom}). The number of sets that can be dominated is in the worst case $|Q|$ since each of them has to be searched in collection $C_{v,t}^{k-1}$ for each feasible terminal $t$ and, if found, marked as dominated. The number of terminal targets and the cardinality of $Q$ are at most $n$ and, as described above, the $Search$ procedure takes $O(|T(s)|)$. Therefore, dominated (proper) covering sets can be annotated with a $O({|T(s)|}^3)$ extra cost at each iteration of Algorithm~\ref{alg:dp}. We can only mark and not delete dominated (proper) covering sets since they can generate non--dominated ones in the next iteration. 

The second task is the {\em generation of routes}. Algorithm~\ref{alg:dp} focuses on proper covering sets and does not maintain a list of corresponding routes. In fact, to build the payoffs matrix for SRG--$v$ we do not strictly need covering routes since covering sets would suffice to determine payoffs. However, we do need them operatively since $\mathcal{D}$ should know in which order targets have to be covered to physically play an action. This task can be accomplished by maintaining an additional list of routes where each route is obtained by appending terminal vertex $t'$ to the route stored for $Q_{v,t}^{k-1}$ when set $Q_{v,t}^{k-1} \cup \{t'\}$ is included in its corresponding collection. At the end of the algorithm only routes that correspond to non--dominated (proper) covering sets are returned. Maintaining such a list introduces a $O(1)$ cost. 

\begin{theorem}
The worst--case complexity of Algorithm~\ref{alg:dp} is $O({|T(s)|}^22^{|T(s)|})$ since it has to compute proper covering sets up to cardinality $|T(s)|$. With annotations of dominances and routes generation the whole algorithm yields a worst--case complexity of $O({|T(s)|}^52^{|T(s)|})$.
\end{theorem}

\subsubsection{Approximation algorithm}
\label{subsec:alg_dyn_progapproximate}
The dynamic programming algorithm presented in the previous section cannot be directly adopted to approximate the maximal covering routes. We notice that even in the case we introduce a logarithmic upper bound over the size of the covering sets generated by Algorithm~\ref{alg:dp}, we could obtain a number of routes that is $O(2^{\log^2(|T(s)|)})$, and therefore exponential. Thus, our goal is to design a polynomial--time algorithm that generates a polynomial number of \textit{good} covering routes. We observe that if we have a total order over the vertices and we work over the complete graph of the targets where each edge corresponds to the shortest path, we can find in polynomial time the maximal covering routes subject to the constraint that, given any pair of targets $t,t'$ in a route, $t$ can precede $t'$ in the route only if $t$ precedes $t'$ in the order. We call \emph{monotonic} a route satisfying a given total order. Algorithm~\ref{alg:randomOrder} returns the maximal monotonic covering routes when the total order is lexicographic  (trivially, in order to change the order, it is sufficient to re--label the targets).

Algorithm~\ref{alg:randomOrder} is based on dynamic programming and works as follows. $R(k,l)$ is a matrix storing in each cell one route, while $L(k,l)$ is a matrix storing in each cell the maximum lateness of the corresponding route, where the lateness associated with a target $t$ is the difference between the (first) arrival time at $t$ along $r$ and $d(t)$ and the maximum lateness of the route is the maximum lateness of the targets covered by the route. The route stored in $R(k,l)$ is the one with the minimum lateness among all the monotonic ones covering $l$ targets where $t_k$ is the first visited target. Thus, basically, when $l=1$, $R(k,l)$ contains the route $\langle v,t_k\rangle$, while, when $l>1$, $R(k,l)$ is defined appending to $R(k,1)$ the best (in terms of minimizing the maximum lateness) route $R(k',l-1)$ for every $k'>k$, in order to satisfy the total order. The whole set of routes in $R$ are returned.\footnote{We notice that dominance can be applied to discard dominated routes. However, in this case, the improvement would be negligible since the total number of routes, including the non--dominated ones, is polynomial.} The complexity of Algorithm~\ref{alg:randomOrder} is $O(|T(s)|^3)$, except the time needed to find all the shortest paths. 

\begin{algorithm}\caption{MonotonicLongestRoute($v,s$)}\label{alg:randomOrder}
\begin{scriptsize}
\begin{algorithmic}[1]

\State $\forall k,k'  \in \{1, 2, \ldots, |T(s)|\}, R(k,k')  = \emptyset, L(k,k')=+\infty, C_R(k)=\emptyset,C_L(k)=+\infty$

\ForAll{$ \forall k \in \{|T(s)|, |T(s)|-1, \ldots, 1\}$} 
	
	\ForAll{$ \forall l \in \{1, 2, \ldots, |T(s)|\}$} 

			\If{$l = 1$} 
    				
				\State $R(k,l) = \langle v, t_{k} \rangle$

				\State $L(k,l) = w^*_{v,t_{k}} - d(t_{k})$

			\Else
    			
				\ForAll{$k'$ s.t. $|T(s)|\geq k' > |T(s)|-k$}    			
    			
					\State $C_R(k') = \langle R(k,1) , R(k',l-1)\rangle$
				    			
					\State $C_L(k') = \max \{L(k,1) , w^*_{v,t_{k}} + w^*_{t_{k},k'} - w^*_{v,k'} + L(k',l-1)\}$	     
			
				\EndFor
   			
   				\State $j = \arg \min_j \{C_L(j)\}$

				\If{$C_L(j)\leq 0$}
    					
					\State $R(k,l) \leftarrow C_R(j)$
	
					\State $L(k,l) \leftarrow C_L(j)$
	
				\EndIf
   			\EndIf				
	\EndFor
\EndFor

\State \textbf{return} $R$
\end{algorithmic}
\end{scriptsize}
\end{algorithm}

We use different total orders over the set of targets, collecting all the routes generated using each total order. The total orders we use are (where ties are broken randomly):
\begin{itemize}
\item increasing order $w^*_{v,t}$: the rationale is that targets close to $v$ will be visited before targets far from $v$;
\item increasing order $d_{v,t}$: the rationale is that targets with short deadlines will be visited before targets with long deadlines;
\item increasing order $d_{v,t}-w^*_{v,t}$: the rationale is that targets with short excess time will be visited before targets with long excess time.
\end{itemize}
In addition, we use a sort of random restart, generating random permutations over the targets.

\begin{theorem}\label{thm:approx1sun}
Algorithm~\ref{alg:randomOrder} provides an approximation with ratio $\Omega(\frac{1}{|T(s)|})$.
\end{theorem}
\emph{Proof sketch.} The worst case for the approximation ratio of our algorithm occurs when the covering route including all the targets exists and each covering route returned by our heuristic algorithm visits only one target. In that case, the optimal expected utility of $\mathcal{D}$ is 1. Our algorithm, in the worst case in which $\pi(t) = 1$ for every target $t$, returns an approximation ratio $\Omega(\frac{1}{|T(s)|})$. It is straightforward to see that, in other cases, the approximation ratio is larger.\hfill$\Box$

\subsection{Branch--and--bound algorithms}
\label{subsec:alg_b_and_b}
The dynamic programming algorithm presented in the previous section essentially implements a breadth--first search. In some specific situations, depth--first search could outperform breadth--first search, e.g., when penetration times are relaxed and good heuristics lead a depth--first search to find in a brief time the maximal covering route, avoiding to scan an exponential number of routes as the breadth--first search would do. In this section, we adopt the branch--and--bound approach to design both an exact algorithm and an approximation algorithm. In particular, in Section~\ref{subsubsec:alg_b_and_b_exact} we describe our exact algorithm, while in Section~\ref{subsubsec:alg_b_and_b_apx} we present the approximation one.

 
\subsubsection{Exact algorithm}
\label{subsubsec:alg_b_and_b_exact}
Our branch--and--bound algorithm (see Algorithm~\ref{alg:routes}) is a tree--search based algorithm working on the space of the covering routes and returning a set of covering routes $R$. It works as follows.

{\bf Initial step.} We exploit two global set variables, $CL_{min}$ and $CL_{max}$ initially set to empty (Steps~1--2 of Algorithm~\ref{alg:routes}). These variables contain {\em closed} covering routes, namely covering routes which cannot be further expanded without violating the penetration time of at least one target during the visit.  $CL_{max}$ contains the covering routes returned by the algorithm (Step~8 of Algorithm~\ref{alg:routes}), while $CL_{min}$ is used for pruning as discussed below. The update of $CL_{min}$ and $CL_{max}$ is driven by Algorithm~\ref{alg:close}, as discussed below. Given a starting vertex $v$ and a signal $s$, for each target $t \in T(s)$ such that $w^*_{v,t}\leq d(t)$ we generate a covering route $r$ with $r(0)=v$ and $r(1)=t$ (Steps~1--3 of Algorithm~\ref{alg:routes}). Thus, $\mathcal{D}$ has at least one covering route per target that can be covered in time from $v$. Notice that if, for some $t$, such minimal route does not exist, then target $t$ cannot be covered because we assume triangle inequality. This does not guarantee that $\mathcal{A}$ will attack $t$ with full probability since, depending on the values $\pi$, $\mathcal{A}$ could find more profitable to randomize over a different set of targets. The meaning of parameter $\rho$ is described below.

\begin{algorithm}[!h]\caption{Branch--and--Bound($v$, $s$, $\rho$)}\label{alg:routes}
\begin{algorithmic}[1]
\scriptsize
\State $CL_{max} \gets \emptyset$
\State $CL_{min} \gets \emptyset$

\ForAll{$t \in T(s)$}

\If{$w^*_{v,t} \le d(t)$} 

\State Tree--Search$(\lceil \rho \cdot  |T(s)| \rceil, \langle v, t \rangle)$

\EndIf

\EndFor

\State \textbf{return} $CL_{max}$
\end{algorithmic}
\end{algorithm}

{\bf Route expansions.} The subsequent steps essentially evolve on each branch according to a depth--first search with backtracking limited by $\rho$ (Step~4 of Algorithm~\ref{alg:routes}). The choice of $\rho$ directly influences the behavior of the algorithm and consequently its complexity. Each node in the search tree represents a route $r$ built so far starting from an initial route $\langle v, t \rangle$. At each iteration, route $r$ is expanded by inserting a new target at a particular position. 
We denote with $r^+(q,p)$ the route obtained by inserting target $q$ after the $p$--th target in $r$. Notice that every expansion of $r$ will preserve the relative order with which targets already present in $r$ will be visited. The collection of all the feasible expansions $r^+$s (i.e., the ones that are covering routes) is denoted by $R^+$ and it is ordered according to a heuristic that we describe below. Algorithm~\ref{alg:heuristic}, described below, is used to generate $R^+$ (Step~1 of Algorithm~\ref{alg:search_bb}). In each open branch (i.e., $R^+\neq \emptyset$), if the depth of the node in the tree is smaller or equal to $\lceil \rho \cdot |T(s)| \rceil$ then backtracking is disabled (Steps~7--11  of Algorithm~\ref{alg:search_bb}), while, if the depth is larger than such value, is enabled (Steps~5--6 of Algorithm~\ref{alg:search_bb}). This is equivalent to fix the relative order of the first (at most) $\lceil \rho\cdot |T(s)| \rceil$ inserted targets in the current route. In this case, with $\rho = 0$ we do not rely on the heuristics at all, full backtracking is enabled, the tree is fully expanded and the returned $R$ is complete, i.e., it contains all the non--dominated covering routes. Route $r$ is repeatedly expanded in a greedy fashion until no insertion is possible. As a result, Algorithm~\ref{alg:search_bb} generates at most $|T(s)|$ covering routes.

\begin{algorithm}[!h]\caption{Tree--Search($k, r$)}\label{alg:search_bb}
\begin{algorithmic}[1]
\scriptsize
\State $R^+=\{r^{(1)}, r^{(2)}, \ldots \} \gets \text{Expand}(r)$ 

\If{$R^+ = \emptyset$}

\State $\text{Close}(r)$

\Else

\If{$k > 0$} 

\State Tree--Search$(k-1,r^{(1)})$

\Else

\ForAll{$r^+ \in R^+$}

\State Tree--Search$(0,r^+)$
\State $\text{Close}(r^+)$

\EndFor

\EndIf

\EndIf
\end{algorithmic}
\end{algorithm}

{\bf Pruning.} 
Algorithm~\ref{alg:close} is in charge of updating $CL_{min}$ and $CL_{max}$ each time a route $r$ cannot be expanded and, consequently, the associated branch must be closed. We call $CL_{min}$ the {\em minimal} set of closed routes. This means that a closed route $r$ belongs to $CL_{min}$ only if $CL_{min}$ does not already contain another $r' \subseteq r$. Steps~1--6 of Algorithm~\ref{alg:close} implement such condition: first, in Steps~2--3 any route $r'$ such that $r' \supseteq r$ is removed from $CL_{min}$, then route $r$ is inserted in $CL_{min}$. Routes in $CL_{min}$ are used by Algorithm~\ref{alg:heuristic} in Steps~2 and~6 for pruning during the search. More precisely, a route $r$ is not expanded with a target $q$ at position $p$ if there exists a route $r' \in CL_{min}$ such that $r' \subseteq r^+(q,p)$. This pruning rule is safe since by definition if $r' \in CL_{min}$, then all the possible expansions of $r'$ are unfeasible and if $r' \subseteq r$ then $r$ can be obtained by expanding from $r'$. This pruning mechanism explains why once a route $r$ is closed is always inserted in $CL_{min}$ without checking the insertion against the presence in $CL_{min}$ of a route $r''$ such that $r'' \subseteq r$. Indeed, if such route $r''$ would be included in $CL_{min}$ we would not be in the position of closing $r$, having $r$ being pruned before by Algorithm~\ref{alg:heuristic} in Step~2 or Step~8.

We use $CL_{max}$ to maintain a set of the generated {\em maximal} closed routes. This means that a closed route $r$ is inserted here only if $CL_{max}$ does not already contain another $r'$ such that $r' \supseteq r$. This set keeps track of closed routes with maximum number of targets. Algorithm~\ref{alg:close} maintains this set by inserting a closed route $r$ in Step~12 only if no route $r' \supseteq r$ is already present in $CL_{max}$. Once the whole algorithm terminates, $CL_{max}$ contains the final solution.


\begin{algorithm}[!h]\caption{Close($r$)}\label{alg:close}
\begin{algorithmic}[1]
\scriptsize
\ForAll{$r' \in CL_{min}$}

\If{$r \subseteq r'$} 

\State $CL_{min} = CL_{min} \setminus \{r'\}$

\EndIf

\EndFor

\State $CL_{min} = CL_{min} \cup \{r\}$

\ForAll{$r' \in CL_{max}$}

\If{$r \subseteq r'$} 

\State \Return

\EndIf

\EndFor

\State $CL_{max} = CL_{max} \cup \{r\}$
\end{algorithmic}
\end{algorithm}

{\bf Heuristic function.} A key component of this algorithm is the heuristic function that drives the search. The heuristic function is defined as $h_r: \{T(s) \setminus T(r)\} \times \{1 \ldots |T(r)|\} \rightarrow \mathbb{Z}$, where $h_r(t',p)$ evaluates the cost of expanding $r$ by inserting target $t'$ after the $p$--th target of $r$. The basic idea, inspired by~\cite{savelsbergh1985local}, is to adopt a conservative approach, trying to preserve feasibility. Given a route $r$, let us define the {\em possible forward shift} of $r$ as the minimum temporal margin in $r$ between the arrival at a target $t$ and $d(t)$:
\begin{center}
$PFS(r) = \min_{t \in T(r)}(d(t) - A_r(t))$
\end{center}

The {\em extra mileage} $e_r(t',p)$ for inserting target $t'$ after position $p$ is the additional traveling cost to be paid:
\begin{center}
$
e_r(t',p) = (A_r(r(t')) + \omega^*_{r(p),t'} + \omega^*_{t',r(p+1)}) - A_r(r(p+1))
$
\end{center}

The {\em advance time} that such insertion gets with respect to $d(t')$  is defined as:
\begin{center}
$a_r(t',p) = d(t') - (A_r(r(p)) + \omega^*_{r(p),t'})$
\end{center}

Finally, $h_r(t',p)$ is defined as:
\begin{center}
$h_r(t',p) = \min \{a_r(t',p); (PFS(r) - e_r(t',p))\}$
\end{center}

\begin{algorithm}[!h]\caption{Expand($r$)}\label{alg:heuristic}
\begin{algorithmic}[1]
\scriptsize
\If{$T_{tight} \nsubseteq T(r)$}

\ForAll{$q \in T_{tight} \setminus T(r)$}\label{alg:for_positions_tight}

$P_q =\{\ p^{(1)}_{q},p^{(2)}_{q}, \ldots p^{(b)}_{q}\} \text{ s.t. } \forall i \in \{1, \ldots, b\},$ $\left\{\begin{array}{l}h_r(q,p^{(i)}_{q}) \ge h_r(q,p^{(i+1)}_{q}) \\  r^+(q, p^{(i)}_{q})  \text{ is a covering route} \\ \not \exists v'\in CL_{min}:r'\subseteq r^+(q,p_q^{(i)})\end{array}\right.$

\EndFor

\State $Q = \{ q^{(1)},q^{(2)}, \ldots, q^{(c)}\} \text{ s.t. } \forall i \in \{1, \ldots, c\},$
\label{alg:for_targets_tight}
 $h_r(q^{(i)},p^{(1)}_{q^{(i)}}) \ge h_r(q^{(i+1)},p^{(1)}_{q^{(i+1)}})$

\State $R^+ = \{ r^{(1)}, r^{(2)}, \ldots r^{(k)}\}\text{ where } \left\{\begin{array}{lll} r^{(1)} &=& r^+(q^{(1)}, p^{(1)}_{q^{(1)}})\\ \cdots & = & \cdots \\ r^{(k)} &=& r^+(q^{(c)}, p^{(b)}_{q^{(c)}}) \end{array}\right.$

\EndIf

\If{$T_{large} \nsubseteq T(r)$}

\ForAll{$u \in T_{large} \setminus T(r)$}\label{alg:for_targets_large}

$Q_p =\{\ q^{(1)}_{p},q^{(2)}_{p}, \ldots q^{(b)}_{p}\} \text{ s.t. } \forall i \in \{1, \ldots, b\}, \left\{ \begin{array}{l}h_r(q^{(i)}_{p},p) \ge h_r(q^{(i+1)}_{p},p) \\ r^+(q^{(i)}_{p},p) \text{ is a covering route} \\ 
\not\exists \; r' \in CL_{min} : r' \subseteq r^+(q^{(i)}_{p},p)\end{array}\right.$

\EndFor

\State $P = \{ p^{(1)},p^{(2)}, \ldots, p^{(c)}\} \text{ s.t. } \forall i \in \{1, \ldots, c\},$
\label{alg:for_positions_large}
$h_r(q^{(1)},p^{(i)}_{q^{(1)}}) \ge h_r(q^{(1)},p^{(i+1)}_{q^{(1)}})$

\State $R^+ = R^+ \cup \{ r^{(k+1)}, r^{(k+2)}, \ldots r^{(K)}\} \text{ where } \left\{\begin{array}{lll} r^{(k+1)} &=& r^+(q^{(1)}_{p}, p^{(1)})\\ \cdots & = & \cdots \\ r^{(K)} &=& r^+(q^{(b)}_{p}, p^{(c)}) \end{array}\right.$

\EndIf

\State \textbf{return} $R^+$

\end{algorithmic}
\end{algorithm}

We partition the set $T(s)$ in two sets $T_{tight}$ and $T_{large}$ where $t \in T_{tight}$ if $d(t) < \delta \cdot \omega^*_{v,t}$ and $t \in T_{large}$ otherwise ($\delta \in \mathbb{R}$ is a parameter). The previous inequality is a non--binding choice we made to discriminate targets with a tight penetration time from those with a large one. Initially, we insert all the tight targets and only subsequently we insert the non--tight targets. We use the two sets according to the following rules (see Algorithm~\ref{alg:heuristic}): 
\begin{itemize}
\item the insertion of a target belonging to $T_{tight}$ is always preferred to the insertion of a target belonging to $T_{large}$, independently of the insertion position;
\item insertions of $t \in T_{tight}$ are ranked according to $h$ considering first the insertion position and then the target;
\item insertions of $t \in T_{large}$ are ranked according to $h$ considering first the target and then the insertion position.
\end{itemize}
The rationale behind this rule is that targets with a tight penetration time should be inserted first and at their best positions. On the other hand, targets with a large penetration time can be covered later. Therefore, in this last case, it is less important which target to cover than when to cover it.

\begin{theorem}
Algorithm~\ref{alg:routes} with $\rho=0$ is an exact algorithm and has an exponential computational complexity since it builds a full tree of covering routes with worst--case size $O(|T(s)|^{|T(s)|})$.
\end{theorem}

\subsubsection{Approximation algorithm}
\label{subsubsec:alg_b_and_b_apx}
Since $\rho$ determines the \emph{completness degree} of the generated tree, we can exploit Algorithm~\ref{alg:routes} tuning $\rho$ to obtain an  approximation algorithm that is faster w.r.t. the exact one.

In fact, when $\rho < 1$ completeness is not guaranteed in favour of a less computational effort. In this case, the only guarantees that can be provided for each covering route $r \in CL_{max}$, once the algorithm terminates are:

\begin{itemize}
\item no other $r' \in CL_{max}$ dominates $r$;
\item no other $r' \notin CL_{max}$ such that $r \subseteq r'$ dominates $r$. Notice this does not prevent the existence of a route $r''$ not returned by the algorithm that visits targets $T(r)$ in a different order and that dominates $r$.
\end{itemize}
When $\rho$ is chosen as $\frac{k}{|T(s)|}$ (where $k \in \mathbb{N}$ is a parameter), the complexity of generating covering routes becomes polynomial in the size of the input. We can state the following theorem, whose proof is analogous to that one of Theorem~\ref{thm:approx1sun}.

\begin{theorem}
Algorithm~\ref{alg:search_bb} with $\rho=\frac{k}{|T(s)|}$ provides an approximation with ratio $\Omega(\frac{1}{|T(s)|})$ and runs in $O({|T(s)|}^3)$ given that heuristic $h_r$ can be computed in $O({|T(s)|}^2)$.
\end{theorem}

\subsection{Solving SRG--$v$}\label{sec:minmax}
Now we can formulate the problem of computing the optimal signal--response strategy for $\mathcal{D}$. Let us denote with $\sigma^{\mathcal{D}}_{v,s}(r)$ the probability with which $\mathcal{D}$ plays route $r$ under signal $s$ and with $R_{v,s}$ the set of all the routes available to $\mathcal{D}$ generated by some algorithm. We introduce function $U_{\mathcal{A}}(r,t)$, representing the utility function of $\mathcal{A}$ and defined as follows:

\begin{equation*}\label{eq:payoffs}
U_{\mathcal{A}}(r,t) = \begin{cases}
\pi(t) &\mbox{if } t \not \in r\\
0 & \mbox{otherwise}
\end{cases}.
\end{equation*}

The best $\mathcal{D}$ strategy (i.e., the maxmin strategy) can be found by solving the following linear mathematical programming problem:

\[
\begin{array}{rl}
\min\quad g_v	& 	\text{s.t.}										\\
\sum\limits_{s \in S(t)} p(s \mid t) \sum\limits_{r \in R_{v,s}} \sigma^{\mathcal{D}}_{v,s}(r) U_{\mathcal{A}}(r,t) \le g_v &\forall t \in T\\
\sum\limits_{r \in R_{v,s}}  \sigma^{\mathcal{D}}_{v,s}(r) = 1 & \forall s \in S\\
\sigma^{\mathcal{D}}_{v,s}(r) \ge 0 &\forall r \in R_{v,s}, s \in S
\end{array}
\]

The size of the mathematical program is composed of $|T|+|S|$ constraints (excluded $\geq 0$ constraints) and  $O(|V| |S| \max_{v,s}\{|R_{v,s}|\})$ variables. This shows that the hardness is due only to $\max_{v,s}\{|R_{v,s}|\}$, which, in its turn, depends only on $|T(s)|$. We provide the following remark.

\begin{remark}
We observe that the discretization of the environment as a graph is as accurate as the number of vertices is large, corresponding to reduce the size of the areas associated with each vertex, as well as to reduce the temporal interval associated with each turn of the game. Our algorithms show that increasing the accuracy of the model in terms of number of vertices requires polynomial time.
\end{remark}
\section{SRG--$v$ on special topologies}
\label{sec:srg_on_special_topologies}
In this section, we focus on special topologies, showing in Section~\ref{subsec:linear_graphs} the topologies with which solving a SRG--$v$ is computationally easy, those that are hard in Section~\ref{subsection:hard}, and the topologies for which the problem remains open in Section~\ref{subsection:borderline}.

\subsection{Easy topologies}
\label{subsec:linear_graphs}
In this section, we show that, with some special topologies, there exists an efficient algorithm to solve exactly the SRG--$v$.

\begin{figure}[H]
\begin{center}
\begin{tikzpicture}
  [scale=.8,auto=left,every node/.style={circle,fill=white,draw,minimum size = 1cm,font=\sffamily\large\bfseries}]
  \node (n1) at (0,0) {$t_1$};
  \node (n2) at (3,0) {$t_2$};
  \node (n3) at (6,0) {$v$};
  \node (n4) at (9,0) {$t_3$};
  \node (n5) at (12,0) {$t_4$};
  
  \path[every node/.style={font=\sffamily\small}]
    (n1) edge node [loop] {} (n2)
	(n2) edge node [loop] {} (n3)
	(n3) edge node [loop] {} (n4)
	(n4) edge node [loop] {} (n5);
	
\end{tikzpicture}
\caption[Linear graph]{Linear graph.}
\label{fig:lin-graph}
\end{center}
\end{figure}
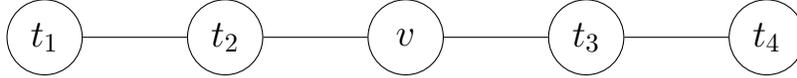

Let us consider a linear graph. An example is depicted in Figure~\ref{fig:lin-graph}. We state the following theorem.

\begin{theorem}
There is a polynomial--time algorithm solving OPT--SRG--$v$ with linear graphs.
\end{theorem}
\emph{Proof}. We show that Algorithm~\ref{alg:dp} requires polynomial time in generating all the pure strategies of $\mathcal{D}$. The complexity of Algorithm~\ref{alg:dp}, once applied to a given instance, depends on the number of proper covering sets (recall Definition~\ref{def:propcovset}). It can be shown that linear graphs have a polynomial number of proper covering sets. Given a starting vertex $v$, any proper covering set $Q$ can be characterized by two extreme targets of $Q$, the first being the farthest from $v$ on the left of $v$ (if any, and $v$ otherwise) and the second being the farthest from $v$ on the right of $v$ (if any, and $v$ otherwise). For example, see Figure~\ref{fig:lin-graph}, given proper covering set $Q=\{t_1,t_2,t_3\}$, the left extreme is $t_1$ and the right extreme is $t_3$. Therefore, the number of proper covering sets for each pair $v,s$ is $O(|T(s)|^2)$. Since the actions available to $\mathcal{D}$ are polynomially upper bounded, the time needed to compute the maxmin strategy is polynomial.\hfill$\Box$

\begin{figure}[h!]
\begin{center}
\begin{tikzpicture}
  [scale=.8,auto=left,every node/.style={circle,fill=white,draw,minimum size = 1cm,font=\sffamily\large\bfseries}]
  \node (n1) at (0,0) {$t_1$};
  \node (n2) at (3,0) {$t_2$};
  \node (n3) at (6,0) {$t_3$};
  \node (n4) at (6,-3) {$t_4$};
  \node (n5) at (6,-6) {$t_5$};
  \node (n6) at (3,-6) {$t_6$};
  \node (n7) at (0,-6) {$t_7$};
  \node (n8) at (0,-3) {$v$};
  
  \path[every node/.style={font=\sffamily\small}]
    (n1) edge node [loop] {} (n2)
	(n2) edge node [loop] {} (n3)
	(n3) edge node [loop] {} (n4)
	(n4) edge node [loop] {} (n5)
	(n5) edge node [loop] {} (n6)
	(n6) edge node [loop] {} (n7)
	(n7) edge node [loop] {} (n8)
	(n8) edge node [loop] {} (n1);
	
\end{tikzpicture}
\caption[Cycle graph]{Cycle graph.}
\label{fig:cycle-graph}
\end{center}
\end{figure}
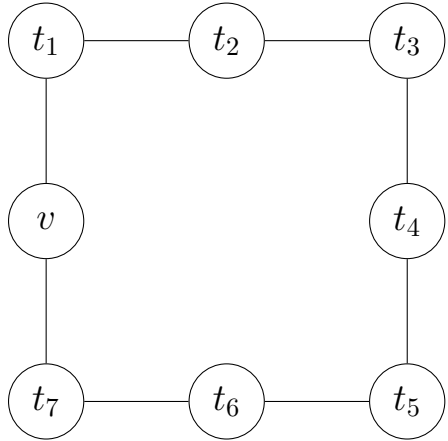

Let us consider a cycle graph. An example is depicted in Figure~\ref{fig:cycle-graph}. We can state the following theorem.

\begin{theorem}
There is a polynomial--time algorithm solving OPT--SRG--$v$ with cycle graphs (perimeters).
\end{theorem}

\emph{Proof}. The proof is analogous to that one of linear graphs. That is, each proper covering set can be characterized by two extremes: the left one and the right one. For example, see Figure~\ref{fig:cycle-graph}, given proper covering set $Q=\{t_1,t_2,t_4,t_5,t_6,t_7\}$, the left extreme is $t_4$ and the right extreme is $t_2$. As in linear graphs, the number of proper covering sets in a cycle graph is $O(|T(s)|^2)$.
\hfill$\Box$

The above results can be generalized to the case of tree graphs where the number of leaves is fixed. We can state the following theorem.
\begin{theorem}\label{theorem:easytreefixed}
There is a polynomial--time algorithm solving OPT--SRG--$v$ with tree graphs where the number of leaves is fixed.
\end{theorem}
\emph{Proof}. The proof is analogous to those of linear and cycle graphs. Here, each proper covering set can be characterized by a tuple of extremes, one for each path connecting $v$ to a leaf. The number of proper covering sets is $O(|T(s)|^n)$ where $n$ is the number of leaves of the tree.
\hfill$\Box$

The above results  show that Questions~2--4 are solvable in polynomial time with the above special topologies. We show in the next section that when the number of leaves in a tree is not fixed, the problem becomes hard. Finally, we provide a remark to the above theorem.

\begin{remark}
We already showed that, given an arbitrary topology, scaling the graph by introducing new vertices is possible with a polynomial--time cost. Theorem~\ref{theorem:easytreefixed} shows that with tree--based graphs this holds even when we introduce new targets. 
\end{remark}

\subsection{Hard topologies}
\label{subsection:hard}

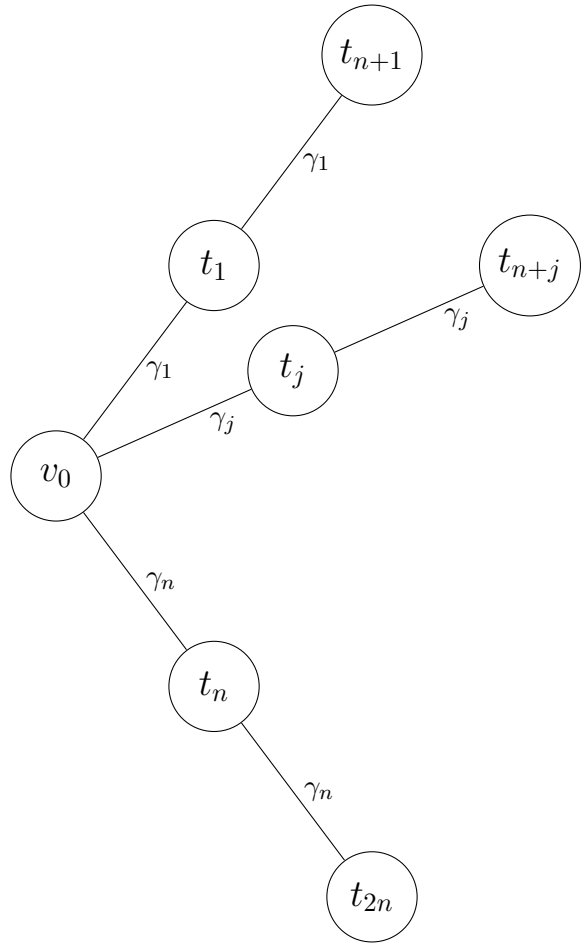
\begin{figure}[h!]
\begin{center}
\begin{tikzpicture}
  [scale=.7,auto=left,every node/.style={circle,fill=white,draw,minimum size = 1.2cm,font=\sffamily\large\bfseries}]
  \node (n1) at (0,8) {$v_0$};
  \node (n2) at (3,12) {$t_1$};
  \node (n3) at (6,16) {$t_{n+1}$};
  \node (n4) at (4.5,10) {$t_j$};
  \node (n5) at (9,12) {$t_{n+j}$};
  \node (n6) at (3,4) {$t_n$};
  \node (n7) at (6,0) {$t_{2n}$};
  
  \path[every node/.style={font=\sffamily\small}]
    (n1) edge node [loop right] {$\gamma_1$} (n2)
    	 edge node [loop right] {~~~$\gamma_j$} (n4)
    	 edge node [loop right] {$\gamma_n$} (n6)
	(n2) edge node [loop right] {$\gamma_1$} (n3)
	(n4) edge node [loop right] {~~~$\gamma_j$} (n5)  
	(n7) edge node [loop right] {$\gamma_n$} (n6);
\end{tikzpicture}
\caption[Special 2--level star graph]{Special 2--level star graph.}
\label{fig:two-level_star_graph}
\end{center}
\end{figure}

Let us consider a special topology, as shown in Figure~\ref{fig:two-level_star_graph} and defined in the following.
\begin{definition}[S2L--STAR] Special 2--level star graph instances (S2L--STAR) are:
\begin{itemize}
\item $V = \{v_0, v_1, v_2, \ldots, v_{2n}\}$, where $v_0$ is the starting position;
\item $T=V\setminus \{v_0\}$, where vertices $v_i$ with $i\in \{1,\ldots,n\}$ are called \emph{inner targets}, while vertices $v_i$ with $i\in \{n+1,\ldots,2n\}$ are called \emph{outer targets};
\item $E=\{(v_0, v_i),(v_i, v_{n+i}):\forall i \in \{1,\ldots,n\}\}$ and we call $i$--th branch the pair of edges $((v_0,v_i), (v_i,v_{n+i}))$;
\item travel costs are $c(v_0, v_i) = c(v_i, v_{n+i}) = \gamma_i$ for every $i\in \{1,\ldots,n\}$, where $\gamma_i \in \mathbb{N}^+$;
\item penetration times are, for $i\in \{1,\ldots,n\}$, $d(t)=\begin{cases} 6H - 3\gamma_i\, & t= v_i \\ 10H - 2\gamma_i, & t= v_{n+i} \end{cases}$, where $H = \frac{\sum_{i =1}^n \gamma_i}{2}$;
\item $\pi(t)=1$ for every $t \in T$.
\end{itemize}
\end{definition}
%

Initially, we show a property of S2L--STAR instances that we shall use below.
\begin{lemma}\label{lemma:S2L--STAR}
If an instance of S2L--STAR admits a maximal covering route $r$ that covers all the targets, then the branches can be partitioned in two sets $C_1$ and $C_2$ such that:
\begin{itemize}
\item all the branches in $C_1$ are visited only once while all the branches in $C_2$ are visited twice, and
\item $\sum\limits_{i\in C_1}\gamma_i = \sum\limits_{i\in C_2}\gamma_i=H$.
\end{itemize}
\end{lemma}

\textit{Proof.}  Initially, we observe that, in a feasible solution, the visit of a branch can be of two forms. If branch $i$ is visited once, then $\mathcal{D}$  will visit the inner target before time $6H-3\gamma_i$ and immediately after the outer target. $C_1$ denotes the set of all the branches visited according to this form. If branch $i$ is visited twice, then $\mathcal{D}$ will visit at first the inner target before time $6H-3\gamma_i$, coming back immediately after to $v_0$, and subsequently in some time after $6H-3\gamma_i$, but before $10H-2\gamma_i$, $\mathcal{D}$ will visit again the inner target and immediately after the outer target. $C_2$ denotes the set of all the branches that are visited according to this form. All the other forms of visits (e.g., three or more visits, different order visits, and visits at different times) are useless and any route in which some branch is not neither in $C_1$ nor in $C_2$ can be modified such that all the branches are either in $C_1$ or in $C_2$ strictly decreasing the cost of the solution as follows:
\begin{itemize}
\item if branch $i$ is visited only once and the visit of the inner target is after time $6H-3\gamma_i$, then the solution is not feasible;
\item if branch $i$ is visited twice and the first visit of the inner target is after time $6H-3\gamma_i$, then the solution is not feasible;
\item if branch $i$ is visited twice and the second visit of the inner target is before time $6H-3\gamma_i$, then the first visit of the branch can be omitted saving $2\gamma_i$;
\item if branch $i$ is visited twice and the outer target is visited during the first visit, then the second visit of the branch can be omitted saving $\geq 2\gamma_i$;
\item if branch $i$ is visited three or more times, all the visits except the first one in which the inner target is visited and the first one in which the outer target is visited can be omitted saving $\geq 2\gamma_i$.
\end{itemize}

We assume that, if there is a maximal covering route $r$ that covers all the targets, then the visits are such that $C_1\cup C_2 = \{1,\ldots,n\}$ and therefore that each branch is visited either once or twice as discussed above. We show below that in S2L--STAR instances such an assumption is always true. Since $r$ covers all the targets, we have that the following conditions are satisfied:
%
%
%
%
%
\begin{align}
2\sum_{i \in C_2} \gamma_i + 4\sum_{i \in C_1} \gamma_i &\leq 6H 				\label{firstdeadlinecon}		\\ 
6\sum_{i \in C_2} \gamma_i + 4\sum_{i \in C_1} \gamma_i &\leq 10H				\label{lastdeadlinecon}
\end{align}
Constraint~(\ref{firstdeadlinecon}) requires that the cost of visiting entirely all the branches in $C_1$ and partially (only the inner target) all the branches in $C_2$ is not larger than the penetration times of the inner targets. Notice that this holds only when the last inner target is first--visited on a branch in $C_1$. We show below that such assumption is always verified. Constraint~(\ref{lastdeadlinecon}) requires that the cost of visiting entirely all the branches in $C_1$ and at first partially and subsequently entirely all the branches in $C_2$ is not larger than the penetration times of the outer targets. We can simplify the above pair of constraints as follows:
\begin{align*}
\underbrace{2\sum_{i \in C_2} \gamma_i + 2\sum_{i \in C_1} \gamma_i}_{4H} + 2\sum_{i \in C_1} \gamma_i &\leq 6H \\ 
2\sum_{i \in C_2} \gamma_i+ \underbrace{4\sum_{i \in C_2} \gamma_i + 4\sum_{i \in C_1} \gamma_i}_{8H} &\leq 10H
\end{align*}
obtaining:
\begin{align*}
\sum_{i \in C_1} \gamma_i  &\leq H \\ 
\sum_{i \in C_2} \gamma_i &\leq H
\end{align*}
since, by definition, $\sum_{i \in C_1} \gamma_i+\sum_{i \in C_2} \gamma_i=2H$, it follows that:
\[
\sum_{i \in C_1} \gamma_i = \sum_{i \in C_2} \gamma_i = H.
\]
Therefore, if $r$ covers all the targets and it is such that all the branches belong either to $C_1$ or to $C_2$, we have that $r$ visits the last outer target exactly at its penetration time. This is because Constraints~(\ref{firstdeadlinecon}) and~(\ref{lastdeadlinecon}) hold as equalities. Thus, as shown above, in any route in which a branch is not neither in $C_1$ nor in $C_2$ we can change the visits such that all the branches are in either $C_1$ or $C_2$, strictly reducing the total cost. It follows that no route with at least one branch that is not neither in $C_1$ nor in $C_2$ can have a total cost equal to or smaller than the penetration time of the outer targets. Similarly, from the above equality it follows that any solution where the last inner target is first--visited on a $C_2$ branch can be strictly improved by moving such branch to $C_1$ and therefore no route in which the last inner target is first--visited on a $C_2$ branch can have a total cost equal to or smaller than the penetration time of the outer targets.
\hfill $\Box$

\begin{definition}[PARTITION] The decision problem PARTITION is defined as:

\noindent INSTANCE: A finite set $I = \{1, 2, \ldots, l\}$, a size $a_i \in \mathbb{N}^+$ for each $i \in I$, and a bound $B \in \mathbb{N}^+$ such that $\sum_{i\in I} a_i = 2B$.

\noindent QUESTION: Is there any subset $I' \subseteq I$ such that $\sum_{i \in I'} a_i = \sum_{i \in I\setminus I'} a_i = B$?
\end{definition}

%
We can now state the following theorem:

\begin{theorem}\label{thm:max-cov-np}
$k$--SRG--$v$ is $\mathcal{NP}$--hard even when restricted to S2L--STAR instances.
\end{theorem}
\textit{Proof.} We provide a reduction from PARTITION that is known to be weakly $\mathcal{NP}$--hard. For the sake of clarity, we divide the proof in  steps.

\textit{Reduction}. We map an instance of PARTITION to an instance of $k$--SRG--$v$ on S2L--STAR graphs as follows
\begin{itemize}
\item $S = \{s\}$,
\item $n=l$ (i.e., the number of branches in S2L--STAR equals the number of elements in PARTITION);
\item $\gamma_i = a_i$ for every $i\in I$;
\item $H = B$,
\item $k=0$.
\end{itemize}
The rationale is that there is a feasible solution for PARTITION if and only if there is the maximal covering route that covers all the targets in a $k$--SRG--$v$ on a S2L--STAR graph.

\textit{If.} From Lemma~\ref{lemma:S2L--STAR} we know that, if there is the maximal covering route that covers all the targets in a $k$--SRG--$v$ on a S2L--STAR graph, then the branches can be partitioned in two sets $C_1,C_2$ such that $\sum_{i\in C_1}\gamma_i = \sum_{i\in C_2}\gamma_i = H$. By construction $\gamma_i = a_i$ and $H= B$. So, if there is the maximal covering route that covers all the targets in a $k$--SRG--$v$ on a S2L--STAR graph, then there is partition of set $I$ in two subsets $I'=C_1$ and $I''=C_2$ such that $\sum_{i\in C_1}\gamma_i = \sum_{i\in I'}a_i = H = B = \sum_{i\in C_2}\gamma_i = \sum_{i\in I''}a_i$.

\textit{Only if.} If PARTITION admits a feasible solution, then, once assigned $I'=C_1$ and $I''=C_2$, it is straightforward to see that the route visits all the targets by their penetration times and therefore that the route is a maximal covering route. 
\hfill $\Box$



Let us notice that the above reduction, differently from that of Theorem~\ref{thm:SRG-npc}, does not exclude the existence of an FPTAS, i.e., Fully Polynomial Time Approximation Scheme. This may hold since PARTITION admits an FPTAS. Furthermore, we observe that S2L--STAR graphs are special kinds of trees and therefore $k$--SRG--$v$ on trees is $\mathcal{NP}$--hard. Finally, we observe that the above result shows that it is unlikely that there is a polynomial--time algorithm solving Questions~1--4.

\subsection{Borderline topologies}
\label{subsection:borderline}
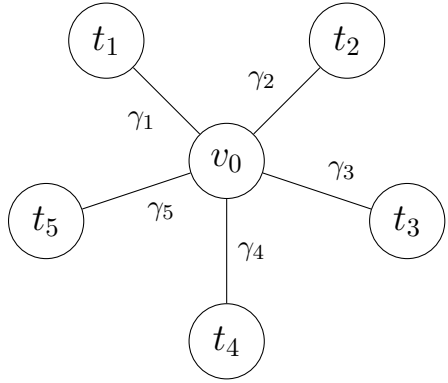
\begin{figure}[H]
\begin{center}
\begin{tikzpicture}
  [scale=.8,auto=left,every node/.style={circle,fill=white,draw,minimum size = 1cm,font=\sffamily\large\bfseries}]
  \node (n1) at (0,0) {$v_0$};
  \node (n2) at (-2,2) {$t_1$};
  \node (n3) at (2,2) {$t_2$};
  \node (n4) at (3,-1) {$t_3$};
  \node (n5) at (0,-3) {$t_4$};
  \node (n6) at (-3,-1) {$t_5$};

  \path[every node/.style={font=\sffamily\small}]
    (n1) edge node [loop] {$\gamma_1$} (n2)
	(n1) edge node [loop] {$\gamma_2$} (n3)
	(n1) edge node [loop] {$\gamma_3$} (n4)
	(n1) edge node [loop] {$\gamma_4$} (n5)
	(n1) edge node [loop] {$\gamma_5$} (n6);
	
\end{tikzpicture}
\caption[Star graph]{Star graph.}
\label{fig:star-graph}
\end{center}
\end{figure}

Let us consider a \textit{star graph}, as shown in Figure~\ref{fig:star-graph}, defined as follows.

\begin{definition}[SIMPLE--STAR] Simple star graph instances (SIMPLE--STAR) are:
\begin{itemize}
\item $V = \{v_0, v_1, v_2, \ldots, v_{n}\}$, where $v_0$ is the starting vertex of $\mathcal{D}$;
\item $T = V\setminus\{v_0\}$;
\item $E=\{(v_0, v_i), \forall i \in \{1,\ldots, n\}\}$;
\item travel costs are $c(v_0, v_i) = \gamma_i$, where $\gamma_i \in \mathbb{N}^+$;
\item penetration times $d_i$ and values $\pi(v_i)$ can be any.
\end{itemize}
\end{definition}

We can state the following theorem.
\begin{theorem}
If the maximal covering route $r$ covering all the targets exists, the Earliest Due Date algorithm returns $r$ in polynomial time  once applied to SIMPLE--STAR graph instances.
\end{theorem}
\emph{Proof.} The Earliest Due Date~\cite{EDD} (EDD) algorithm is an optimal algorithm for synchronous (i.e., without release times) aperiodic scheduling with deadlines. It executes (without preemption) the tasks in ascending order according to the deadlines, thus requiring polynomial complexity in the number of tasks. Any SIMPLE--STAR graph instance can be easily mapped to a synchronous aperiodic scheduling problem: each target $t_i$ is an aperiodic task $J_i$, the computation time of $J_i$ is equal to $2\gamma_i$, the deadline of task $J_i$ is $d(t_i)+\gamma_i$. It is straightforward to see that, if EDD returns a feasible schedule, then there is the maximal covering route, and, if EDD returns a non--feasible schedule, then there is not any maximal covering route.
\hfill$\Box$

The above result shows that Question~2 can be answered in polynomial time. We show that also Question~3 can be answered in polynomial time be means of a simple variation of EDD algorithm. 
\begin{theorem}
Given a signal $s$, the best pure strategy of $\mathcal{D}$ in an SRG--$v$ game on SIMPLE--STAR graph instances can be found in polynomial time.
\end{theorem}
\emph{Proof}. Given a signal $s$, the algorithm that finds the best pure strategy is a variation of the EDD algorithm. For the sake of clarity, we describe the algorithm in the simplified case in which there is only one signal $s$. The extension to the general case is straightforward. The algorithm works as follows:
\begin{enumerate}
\item apply EDD,
\item if the maximal covering route exists, then return it,
\item else remove the target $t$ with the smallest $\pi(t)$ from $T(s)$,
\item go to Point 1.
\end{enumerate}
Essentially, the algorithm returns the subset of targets admitting a covering route minimizing the maximum value among all the non--covered targets.
\hfill$\Box$

Although the treatment of SIMPLE--STAR graph instances in pure strategies is computationally easy, it is not clear if the treatment keeps being easy when $\mathcal{D}$ is not restricted to play pure strategies. We just observe that Algorithm~1 requires exponential complexity, the number of proper covering sets being exponential. Thus, the complexity of solving Questions~1 and~4 remains unaddressed. 

\section{Patrolling game}
\label{sec:patrolling_game}
In this section, we focus on the PG. Specifically, in Section~\ref{subsec:theorem} we state our main result showing that patrolling is not necessary when an alarm system is present, in Section~\ref{subsec:algorithm} we propose the algorithm to deal with the PG, in Section~\ref{subsec:summary} we summarize the complexity results about Questions~1--4. 
 
\subsection{Stand still}
\label{subsec:theorem}
We focus on the problem of finding the best patrolling strategy given that we know the best signal--response strategy for each vertex $v$ in which $\mathcal{D}$ can place.  Given the current vertex of $\mathcal{D}$ and the sequence of the last, say $n$, vertices visited by $\mathcal{D}$ (where $n$ is a tradeoff between effectiveness of the solution and computational effort), a patrolling strategy is usually defined as a randomization over the next adjacent vertices~\cite{BasilicoGA12}. We define $v^* = \argmin_{v \in V}\{g_v\}$, where $g_v$ is the value returned by the optimization problem described in Section~3.3, as the vertex that guarantees the maximum expected utility to $\mathcal{D}$ over all the SRG--$v$s. We show that the maxmin equilibrium strategy in PG prescribes that $\mathcal{D}$ places at $v^*$, waits for a signal, and responds to it.

\begin{theorem}\label{thm:nopatrol}
Without false positives and missed detections, if $\forall t \in T$ we have that $|S(t)| \ge 1$, then any patrolling strategy is dominated by the placement in $v^*$.
\end{theorem}

\noindent
\textit{Proof.} Any patrolling strategy different from the placement in $v^*$ should necessarily visit a vertex $v' \neq v^*$.  Since the alarm system is not affected by missed detections, every attack will raise a signal which, in turn, will raise a response yielding an utility of $g_x$ where $x$ is the current position of $\mathcal{D}$ at the moment of the attack. Since $\mathcal{A}$ can observe the current position of $\mathcal{D}$ before attacking, $x = \argmax_{v \in P}\{g_{v}\}$ where $P$ is the set of the vertices patrolled by $\mathcal{D}$. Obviously, for any $P \supseteq \{v^*\}$ we would have that $g_x \ge g_{v^*}$ and therefore placing at $v^*$ and waiting for a signal is the best strategy for $\mathcal{D}$.\hfill$\Box$

The rationale is that, if the patrolling strategy of $\mathcal{D}$ prescribes to patrol a set of vertices, say $V'$, then, since $\mathcal{A}$ can observe the position of $\mathcal{D}$, the best strategy of $\mathcal{A}$ is to wait for $\mathcal{D}$ being in $v'=\argmax_{v \in V'}\{g_v\}$ and then to attack. Thus, by definition of $g_{v^*}$, if $\mathcal{D}$ leaves $v^*$ to patrol additional vertices the expected utility it receives is no larger than that it receives from staying in~$v^*$.

A deeper analysis of Theorem~\ref{thm:nopatrol} can show that its scope does include cases where missed detections are present up to a non--negligible extent. For such cases, placement--based strategies keep being optimal even in the case when the alarm systems fails in detecting an attack. We encode the occurrence of this robustness property in the following proposition, which we shall prove by a series of examples.

\begin{proposition}\label{prop:robustness}
There exist Patrolling Games where staying in a vertex, waiting for a signal, and responding to it is the optimal patrolling strategy for $\mathcal{D}$ even with a missed detection rate $\alpha = 0.5$.
\end{proposition}
\textit{Proof.} The expected utility for $\mathcal{D}$ given by the placement in $v^*$ is $(1-\alpha)(1-g_{v^*})$, where $(1-\alpha)$ is the probability with which the alarm system correctly generates a signal upon an attack and $(1-g_{v^*})$ denotes $\mathcal{D}$'s payoff when placed in $v^*$. A non--placement--based patrolling strategy will prescribe, by definition, to move between at least two vertices. From this simple consideration, we observe that an upper bound to the expected utility of any non--placement strategy is entailed by the case where $\mathcal{D}$ alternately patrols vertices $v^*$ and $v^*_2$, where $v^*_2$ is the second best vertex in which $\mathcal{D}$ can statically place. Such scenario gives us an upper bound over the expected utility of non--placement strategies, namely $1-g_{v^*_2}$. 
It follows that a sufficient condition for the placement in $v^*$ being optimal is given by the following inequality:

\begin{equation}\label{eq_false_negatives}
(1-\alpha)(1-g_{v^*})>(1-g_{v^*_2}).
\end{equation}

\begin{figure}[h]
\centering
\subfigure[Equation~\ref{eq_false_negatives} holds for $\alpha \le 0.25$.]{
\begin{tabular}{c}
\begin{tikzpicture}
  [scale=.8,auto=left,every node/.style={circle,fill=white,draw,minimum size = 1cm,font=\sffamily\large\bfseries}]
  \node (n2) at (0,2) {$t_2$};
  \node (n1) at (2,0) {$t_1$};
  \node (n3) at (-2,0) {$t_3$};
  \node (n4) at (0,-2) {$t_4$};

  \path[every node/.style={font=\sffamily\small}]
    (n1) edge node [loop] {} (n2)
	(n1) edge node [loop] {} (n3)
	(n1) edge node [loop] {$2$} (n4)
	(n3) edge node [loop] {} (n4);	
\end{tikzpicture}
\end{tabular}
\begin{tabular}{ c | c | c | c }
    $t$ & $\pi(t)$ & $d(t)$ & $p(s_1 \mid t)$ \\ \hline
    $t_1$ & 0.5 & 1  & 1.0   \\ \hline
    $t_2$ & 0.5 & 3 & 1.0 \\ \hline
    $t_3$ & 0.5 & 2 & 1.0 \\ \hline
    $t_4$ & 0.5 & 2 & 1.0 
\end{tabular}}


\subfigure[Equation~\ref{eq_false_negatives} holds for $\alpha \le 0.5$.]{
\begin{tabular}{c}
\begin{tikzpicture}
  [scale=.8,auto=left,every node/.style={circle,fill=white,draw,minimum size = 1cm,font=\sffamily\large\bfseries}]
  \node (n2) at (0,2) {$t_2$};
  \node (n1) at (2,0) {$t_1$};
  \node (n3) at (-2,0) {$t_3$};
  \node (n4) at (0,-2) {$t_4$};

  \path[every node/.style={font=\sffamily\small}]
    (n1) edge node [loop] {} (n2)
	(n1) edge node [loop] {} (n3)
	(n1) edge node [loop] {$2$} (n4)
	(n3) edge node [loop] {} (n4);
	
\end{tikzpicture}
\end{tabular}
  \begin{tabular}{ c | c | c | c }
    $t$ & $\pi(t)$ & $d(t)$ & $p(s_1 \mid t)$ \\ \hline
    $t_1$ & 1.0 & 1  & 1.0   \\ \hline
    $t_2$ & 1.0 & 3 & 1.0  \\ \hline
    $t_3$ & 1.0 & 2 & 1.0 \\ \hline
    $t_4$ & 1.0 & 2 & 1.0 
  \end{tabular}}
\caption{Two examples proving Proposition~\ref{prop:robustness}.}
\label{fig:graphs_50}
\end{figure}
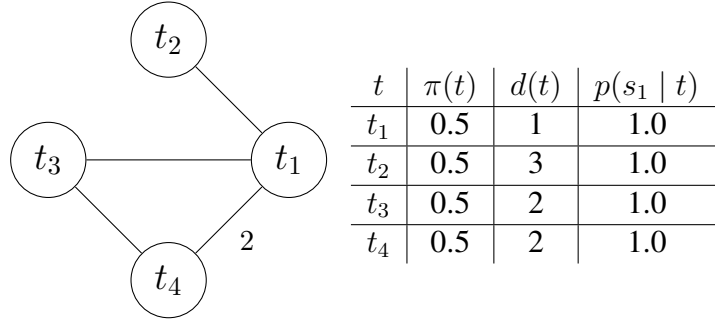
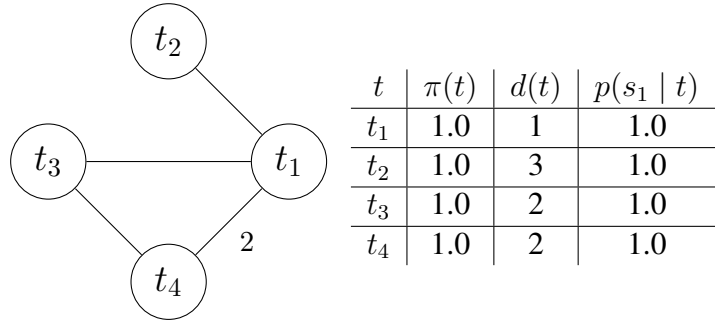

To prove Proposition~\ref{prop:robustness}, it then suffices to provide a Patrolling Game instance where Equation~\ref{eq_false_negatives} holds under some non--null missed detection rate $\alpha$. In Fig.~\ref{fig:graphs_50}(a) and Fig.~\ref{fig:graphs_50}(b), we report two of such examples. The depicted settings have unitary edges except where explicitly indicated. For both, without missed detections, the best patrolling strategy is a placement $v^* = 4$. When allowing missed detections, in Fig.~\ref{fig:graphs_50}(a) it holds that $g_{v^*} = 0$ and $g_{v^*_2} = 0.75$, where $v^* = 4$ and $v^*_2 = 1$. Thus, by Equation~\ref{eq_false_negatives}, placement $v^* = 4$ is the optimal strategy for $\alpha \le 0.25$. Under the same reasoning scheme, in Fig.~\ref{fig:graphs_50}(b) we have that $g_{v^*} = 0$ and $g_{v^*_2} = 0.5$, making the placement $v^* = 4$ optimal for any $\alpha \le 0.5$. \hfill$\Box$

It is reasonable to expect that a similar result holds also for the case with false positives. However, dealing with false positives is much more intricate than handling false negative and requires new models, e.g., $\mathcal{D}$ could respond to an alarm signal only with a given probability and with the remaining probability could stay in the current vertex. For this reason, we leave the treatment of false positives and a more accurate treatment of false negatives to future works.

\subsection{Computing the best placement}
\label{subsec:algorithm}
Under the absence of false positives and missed detections, Theorem~\ref{thm:nopatrol} simplifies the computation of the patrolling strategy by reducing it to the problem of finding $v^*$. To such aim, we must solve a SRG--$v$ for each possible starting vertex $v$ and select the one with the maximum expected utility for $\mathcal{D}$. Algorithm~\ref{alg:bestplace} depicts the solving algorithm. Function $SolveSRG(v)$ returns the optimal value $1-g_{v^*}$. The complexity is linear in $|V|$, once $g_v$ has been calculated for every $v$.

\begin{algorithm}[!h]\caption{BestPlacement($G$, $s$)}\label{alg:bestplace}
\begin{algorithmic}[1]
\scriptsize
\State $U(v) \leftarrow 0$ for every $v \in V$

\ForAll{$v \in V$}

\State $U(v)\leftarrow SolveSRG(v)$

\EndFor

\State \textbf{return} $\max(U)$
\end{algorithmic}
\end{algorithm}

Since all the vertices are possible starting points, we should face this hard problem (see Theorem~\ref{thm:SRG-npc}) $|V|$ times, computing, for each signal, the covering routes from all the vertices. To avoid this issue, we ask whether there exists an algorithm that in the worst case allows us to consider a number of iterations such that solving the problem for a given starting vertex $v$ could help us finding the solution for another starting vertex $v'$. In other words, considering a specific set of targets, we wonder whether a solution for COV--SET with starting vertex $v$ can be used to derive, in polynomial time, a solution to COV--SET for another starting vertex $v'$. This would allow us to solve an exponential--time problem only once instead of solving it for each vertex of the graph. To answer this question, we resort to hardness results for reoptimization, also called \emph{locally modified} problems~\cite{bockenhauer2006reusing}. We show that, even if we know all the covering routes from a starting vertex, once we changed the starting vertex selecting an adjacent one, finding the covering routes from the new starting vertex is hard. 

\begin{definition}[{\scshape lm}--COV--ROUTE] A locally modified covering route ({\scshape lm}--COV--ROUTE) problem is defined as follows:

\noindent INSTANCE:  graph $G=(V,E)$, a set of targets $T$ with penetration times $d$, two starting vertices $v_1$ and $v_2$ that are adjacent, and a covering route $r_1$ with $r_1(0)=v_1$ such that $T(r_1)=T$.

\noindent QUESTION: is there $r_2$ with $r_2(0)=v_2$ and $T(r_2)=T$?
\end{definition}

\begin{theorem}
{\scshape lm}--COV--ROUTE is $\mathcal{NP}$--complete.\label{theorem:reoptimization}
\end{theorem}

\noindent
%
\textit{Proof}. We divide the proof in two steps, membership and hardness.

\emph{Membership}. Given a YES certificate constitutes by a route, the verification is easy, requiring one to apply the route and check whether each target is visited by its deadline. It requires linear time in the number of targets.

\emph{Hardness}. Let us consider the Restricted Hamiltonian Circuit problem (RHC) which is known to be $\mathcal{NP}$--complete. RHC is defined as follows: given a graph $G_H=(V_H,E_H)$ and an Hamiltonian path $P=\langle h_1, \ldots, h_n\rangle$ for $G_H$ such that $h_i \in V_H$ and $(h_1,h_n) \notin E_H$, find an Hamiltonian cycle for $G_H$.  From such instance of RHC, following the approach of~\cite{bockenhauer2006reusing}, we build the following instance for {\scshape lm}--COV--ROUTE: 

\begin{itemize}
\item $V = V_H\cup\{v_1,v_2,v_t\}$;
\item $T = V_H\cup\{v_t\}$;
\item $E = E_H\cup \{(h_n,v_t),(h_i,v_s):i\in \{1,\ldots,n\}\}$;
\item $d(v_t)=n+1$ and $d(t)=n$ for any $t \in T$ with $t\neq v_t$;
\item $w_{v,v'}=
\begin{cases} 
		1			&	\textnormal{if } v = h_n, v' = v_t									\\
		1			&	\textnormal{if } v = h_i, v' = h_j, \forall i,j \in \{1,\ldots n\}				\\
		1			&	\textnormal{if } v = v_1, v' = h_1									\\
		2			&	\textnormal{if } v = v_1, v' = h_{n-1}								\\
		\geq 2		&	\textnormal{if } v = v_1, v' = h_i, \forall i\in \{1,\ldots n-2, n\}			\\
		\geq 2		&	\textnormal{if } v = v_1, v' = v_t									\\
		2			&	\textnormal{if } v = v_2, v' = h_1									\\
		1			&	\textnormal{if } v = v_2, v' = h_{n-1}								\\
		\geq 2		&	\textnormal{if } v = v_2, v' = h_i, \forall i\in \{1,\ldots n-2, n\}			\\
		\geq 2		&	\textnormal{if } v = v_2, v' = v_t
\end{cases}$;
\item $r_1=\langle v_1, h_1, \cdots, h_n, v_t \rangle$.
\end{itemize}
Basically, given $G_H$ we introduce three vertices $v_1,v_2,v_t$, where $v_1,v_2$ are adjacent starting vertices and $v_t$ is a target. We know the covering routes from $v_1$, and we aim at finding the covering routes from $v_2$. The new starting vertex ($v_2$) is closer to $h_{n-1}$ than the previous one ($v_1$) by 1 and farther from $h_1$ than previous one ($v_1$) by 1. There is no constraint over the distances between the starting vertices and the other targets except that they are larger than or equal to 2. We report in Figure~\ref{figure:locallymodified} an example of the above construction. Notice that by construction, if the maximal covering route $r_2$ with $r_2(0)=v_2$ and $T(r_2)=T$ exists, then $v_t$ must be the last visited target. Route $r_1$ is covering since $\langle h_1, \ldots, h_n\rangle$ is a Hamiltonian path for $G_H$. We need to show that route $r_2$ with $r_2(0)=v_2$ and $T(r_2)=T$ exists if and only if $G_H$ admits a Hamiltonian cycle. It can observed that, if $r_2$ exists, then it must be such that $r_2= \langle v_2, h_{n-1},\ldots, h_n,v_t\rangle$ and therefore $\langle h_{n-1}, \ldots, h_n\rangle$ must be a Hamiltonian path for $G_H$. Since we know, by $r_1$, that $(h_{n-1},h_n) \in E_H$, it follows that $\langle h_{n-1}, \ldots, h_n, h_{n-1}\rangle$ is a Hamiltonian cycle. This concludes the proof.
\hfill$\Box$

This shows that iteratively applying Algorithm~\ref{alg:dp} to SRG--$v$ for each starting vertex $v$ and then choosing the vertex with the highest utility is the best we can do in the worst case.

\begin{figure}
\begin{center}
\begin{tikzpicture}
  [scale=.4,auto=left,every node/.style={circle,fill=white,draw,minimum size = 1cm,font=\sffamily\large\bfseries}]
  \node (n1) at (3,0) {$h_1$};
  \node (n2) at (0,4) {$h_2$};
  \node (n3) at (3,8) {$h_3$};
  \node (n4) at (6,4) {$h_4$};
  \node (n5) at (9,0) {$v_1$};
  \node (n6) at (15,0) {$v_2$};
  \node (n7) at (12,4) {$h_5$};
  \node (n8) at (12,8) {$h_6$};
  \node (n9) at (21,0) {$h_9$};
  \node (n10) at (18,4) {$h_8$};
  \node (n11) at (21,8) {$h_7$};
  \node (n12) at (24,4) {$h_{10}$};
  \node (n13) at (28,4) {$v_t$};
  
  \path[every node/.style={font=\sffamily\small}]
    (n1) edge node [loop right] {} (n2)
    	 edge node [loop right] {} (n4)
    	 edge node [loop right] {} (n5)
	(n2) edge node [loop right] {} (n3)
	(n2) edge node [loop right] {} (n4)
	(n3) edge node [loop right] {} (n4)
		 edge node [loop right] {} (n8)  
	(n4) edge node [loop right] {} (n7)
	(n5) edge node [loop right] {} (n6)
	(n6) edge node [loop right] {} (n9)
	(n7) edge node [loop right] {} (n8)
	(n7) edge node [loop right] {} (n10)
	(n8) edge node [loop right] {} (n11)
	(n9) edge node [loop right] {} (n10)
	(n9) edge node [loop right] {} (n12)
	(n9) edge node [loop right] {} (n10)
	(n10) edge node [loop right] {} (n11)
	(n10) edge node [loop right] {} (n12)
	(n11) edge node [loop right] {} (n12)
	(n12) edge node [loop right] {} (n13);
		
\end{tikzpicture}
\caption{Example of construction used in the proof of Theorem~\ref{theorem:reoptimization}: the Hamiltonian path $\langle h_1,h_2,h_3,h_4,h_5,h_6,h_7,h_8,h_9,h_{10}\rangle$ on $G_H$ is given, as well as covering route $r_1$ with $r_1(0)=v_1$ and $T(r_1)=T$. It can be observed that there is another Hamiltonian path for $G_H$, i.e., $\langle h_9,h_8,h_5,h_4,h_1,h_2,h_3,h_6,h_7,h_{10}\rangle$, allowing covering route $r_2$ with $r_2(0)=v_2$ and $T(r_2)=T$. Notice that, if we remove the edge $(h_5,h_8)$, then covering route $r_2$ such that $r_2(0)=v_2$ and $T(r_2)=T$ does not exist.}
\label{figure:locallymodified}
\end{center}
\end{figure}

\subsection{Summary of results}
\label{subsec:summary}
We summarize our computational results about Questions~1--4 in Table~\ref{tbl:questions_comp_compl}, including also results about the resolution of the PG. We use `?' for the problems remained open in this paper. 

\begin{table}[h]
\begin{center}
\begin{tabular}{ | c | l | l | l | l | l |}\hline
\backslashbox[2mm]{Question}{Topology} & Linear & Cycle & Star & Tree & Arbitrary  \\ \cline{1-6}
Question 1 					& $\mathcal{FP}$ 	& $\mathcal{FP}$ 	& ? 				& $\mathcal{FNP}$--hard 		& $\mathcal{APX}$--hard		\\ \cline{1-6}
Question 2 					& $\mathcal{P}$ 	& $\mathcal{P}$ 	& $\mathcal{P}$ 	& $\mathcal{NP}$--hard 		& $\mathcal{NP}$--hard 		\\ \cline{1-6}
Question 3 					& $\mathcal{P}$ 	& $\mathcal{P}$ 	& $\mathcal{P}$ 	& $\mathcal{NP}$--hard 		& $\mathcal{NP}$--hard 		\\ \cline{1-6}
Question 4 					& $\mathcal{P}$ 	& $\mathcal{P}$ 	& ? 				& $\mathcal{NP}$--hard 		& $\mathcal{NP}$--hard 		\\ \cline{1-6}
Question 2 			 		& $\mathcal{FP}$ 	& $\mathcal{FP}$ 	& ? 				& ? 						& $\mathcal{NP}$--hard		\\ 
Reoptimization					&				& 				&				&						& 						\\ \cline{1-6}
\end{tabular}
\end{center}
\caption{Computational complexity of discussed questions.}
\label{tbl:questions_comp_compl}
\end{table}


%
\section{Experimental evaluation}
\label{sec:experimental_evaluation}
In this section, we experimentally evaluate our algorithms. We implemented our algorithms in MATLAB and we used a 2.33GHz LINUX machine to run our experiments. For a better analysis, we provide two different experimental evaluations. In Section~\ref{subsection:worstcaseexp}, we apply our algorithms to worst--case instances suggested by our $\mathcal{NP}$--hardness reduction, in order to evaluate the worst--case performance of the algorithms and to investigate experimentally the gap between our $\mathcal{APX}$--hardness result and the theoretical guarantees of our approximation algorithms. In Section~\ref{subsection:realisticexp}, we apply our algorithms to a specific realistic instance we mentioned in Section~\ref{sec:introduction}, Expo 2015.

\subsection{Worst--case instances analysis}
\label{subsection:worstcaseexp}
We evaluate the scalability of Algorithm~\ref{alg:dp} and the quality of the solution returned by  our approximation algorithms for a set of instances of SRG--$v$. We do not include results on the evaluation of the algorithm to solve completely a PG, given that it trivially requires asymptotically $|V|$ times the effort required by the resolution of a single instance of SRG--$v$. In the next section we describe our experimental setting, in Section~\ref{ex:exact} we provide a quantitative analysis of the exact algorithms while in Section~\ref{ex:approx} we evaluate the quality of our approximations.

\subsubsection{Setting}
As suggested by the proof of Theorem~\ref{thm:covsetnpc}, we can build hard instances for our problem from instances of HAMILTONIAN--PATH. More precisely, our worst--case instances are characterized by: 
\begin{itemize}
\item all the vertices are targets, 
\item edge costs are set to $1$, 
\item there is only one signal, 
\item penetration times are set to $|T| - 1$, 
\item values are drawn from $(0,1]$ with uniform probability for all the targets,  
\item the number of edges is drawn from a normal distribution with mean $\epsilon$, said \emph{edge density} and defined as $\epsilon = |E|/\frac{|T|(|T|-1)}{2}$, and 
\item starting vertex $v$ is drawn among the targets of $T$ with uniform probability. 
\end{itemize}
We explore two parameter dimensions: the number of targets $|T|$ and the value of edge density $\epsilon$. In particular, we use the following values: 
\begin{align*}
|T|			&	\in \{6,8,10,12,14,16,20,25,30,35,40,45,50\},	\\ 
\epsilon 		&	\in \{0.05,0.10,0.25,0.50,0.75,1.00\}.
\end{align*} 
For each combination of values of $|T|$ and $\epsilon$, we randomly generate 100 instances with the constraint that, if $\epsilon \frac{|T|^2}{2}< |T|$, we introduce additional edges in order to assure the graph connectivity. The suitability of our worst--case analysis is corroborated by the results obtained with a realistic setting (see Section~\ref{subsection:realisticexp}) which present hard subproblems characterized by the features listed above.

\subsubsection{Exact algorithms scalability}\label{ex:exact}
\begin{figure}[h]
\hspace{-0.75cm}
\begin{center}
\includegraphics[scale=0.7]{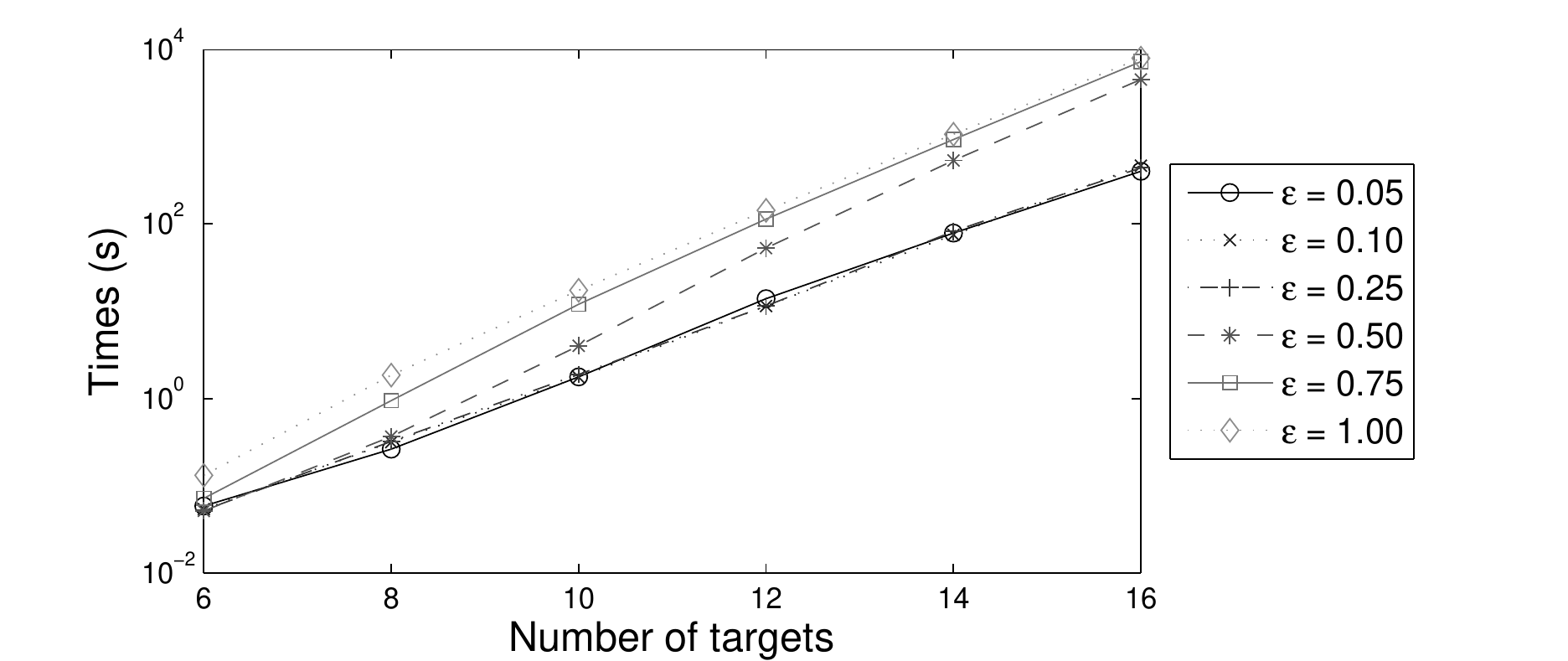}
\end{center}
\caption{Compute times in seconds of our exact dynamic programming algorithm (Algorithm~\ref{alg:dp}), with the annotation of dominated (proper) covering sets and the generation of the routes,  as $|T|$ and $\epsilon$ vary.}
\label{fig:exacttimes}
\end{figure}

We report in Figure~\ref{fig:exacttimes} the compute time (averaged over 100 SRG--$v$ instances) required by our exact dynamic programming algorithm (Algorithm~\ref{alg:dp}), with the annotation of dominated (proper) covering sets and the generation of the routes,  as $|T|$ and $\epsilon$ vary. We report in~\ref{appendix:boxplots},  the boxplots showing the statistical significance of the results. It can be observed that the compute times are exponential in $|T|$, the curves being lines in a semilogarithmic plot, and the value of $\epsilon$ determines the slope of the line. Notice that with $\epsilon \in \{0.05,0.10, 0.25\}$ the number of edges is almost the same when $|T|\leq 16$ due to the constraint of connectivity of the graph, leading thus to the same compute times. Beyond 16 targets, the compute times of our exact dynamic programming algorithm are excessively long (with only $\epsilon=0.25$, the compute time when $|T|=20$ is lower than$10^4$ seconds). Interestingly, the compute time monotonically decreases as $\epsilon$ decreases. This is thanks to the fact that the number of proper covering sets dramatically reduces as $\epsilon$ reduces and that Algorithm~\ref{alg:dp} enumerates only the proper covering sets.

We do not report any plot of the compute times of our exact branch--and--bound algorithm, since it requires more than $10^4$ seconds when $|T|>8$ even with $\epsilon=0.25$, resulting thus non--applicable in practice. This is because the branch--and--bound algorithm has a complexity $O(|T|^{|T|})$, while the dynamic programming algorithm has a complexity $O(2^{|T|})$.

\begin{figure}[h]
\hspace{-1.5cm}
\begin{center}
\includegraphics[scale=0.7]{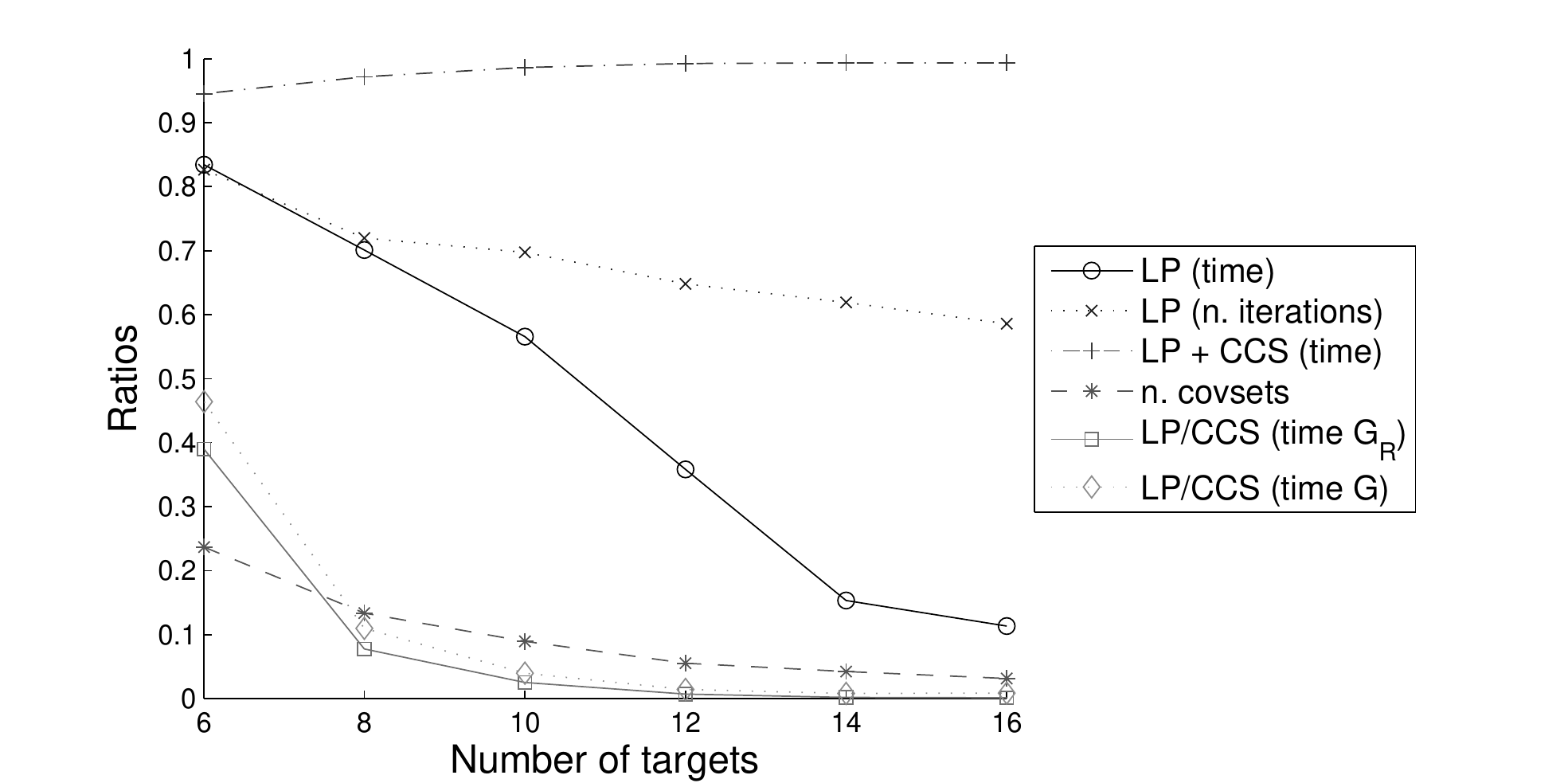}
\end{center}
\caption{Ratios evaluating dominances with $\epsilon = 0.25$ as $|T|$ varies.}
\label{fig:ratios_dominances}
\end{figure}

Figure~\ref{fig:ratios_dominances} shows the impact of discarding dominated actions from the game when $\epsilon = 0.25$. It depicts the trend of some performance ratios for different metrics. We shall call $\mathcal{G}$ the complete game including all $\mathcal{D}$'s dominated actions and $\mathcal{G}_R$ the reduced game; CCS will denote the full version of Algorithm~\ref{alg:dp} and LP will denote the linear program to solve SRG--$v$. Each instance is solved for a random starting vertex $v$; we report average ratios for $100$ instances. ``n. covsets'' is the ratio between the number of covering sets in $\mathcal{G}_R$ and in $\mathcal{G}$. Dominated actions constitute a large percentage, increasing with the number of targets. This result indicates that the structure of the problem exhibits a non-negligible degree of redundancy. LP times (iterations) report the ratio between $\mathcal{G}_R$ and $\mathcal{G}$ for the time (iterations) required to solve the maxmin linear program. A relative gain directly proportional to the percentage of dominated covering sets is observable (LP has less variables and constraints). A similar trend is not visible when considering the same ratio for the total time, which includes CCS. Indeed, the time needed by CCS largely exceed LP's and removal of dominated actions determines a polynomial additional cost, which can be seen in the slightly increasing trend of the curve. The relative gap between LP and CCS compute times can be assessed by looking at the LP/CCS curve: when more targets are considered the time taken by LP is negligible w.r.t. CCS's. This shows that removing dominated actions is useful, allowing a small improvement in the average case, and assuring an exponential improvement in the worst case.

\begin{figure}[h]
\hspace{-0.75cm}
\begin{center}
\includegraphics[scale=0.7]{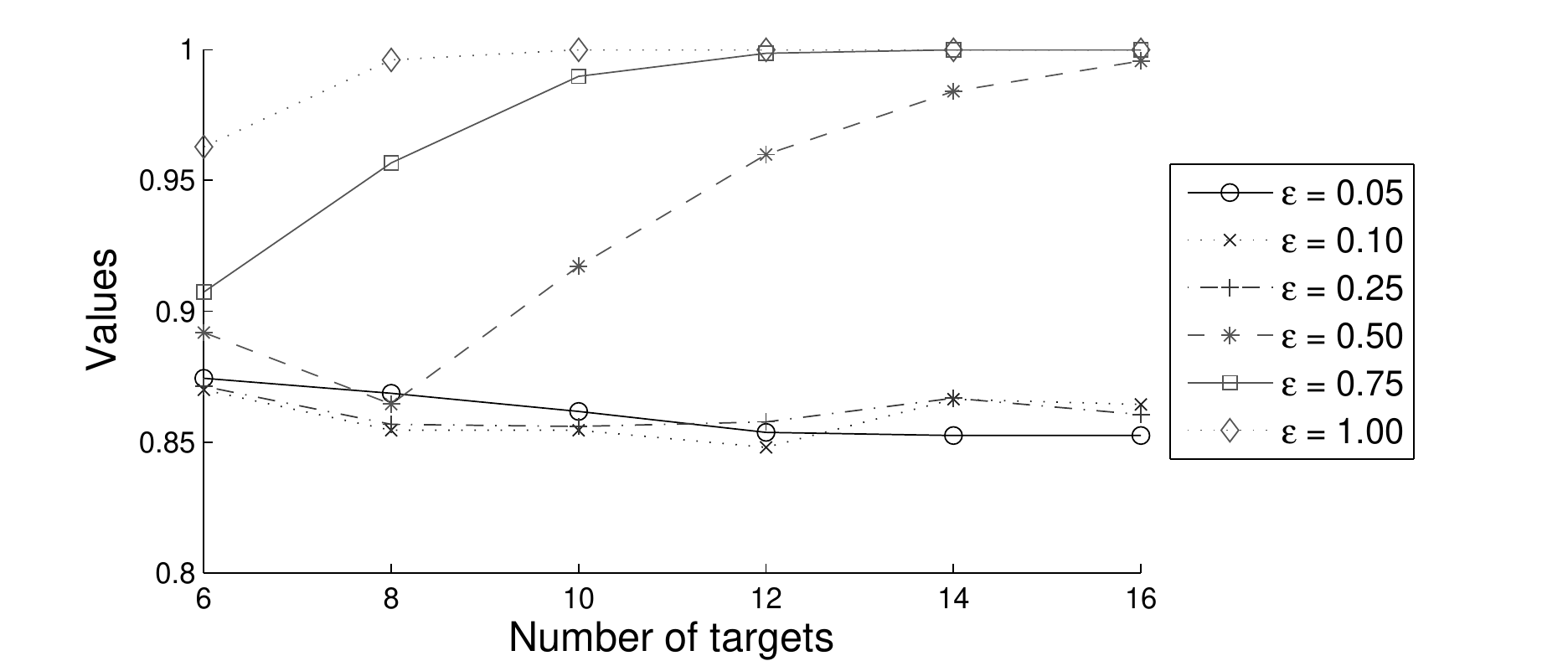}
\end{center}
\caption{Optimal game values as $|T|$ and $\epsilon$ vary.}
\label{fig:exactvalues}
\end{figure}

Figure~\ref{fig:exactvalues} shows the game value for $\mathcal{D}$, $1-g_v$, as $|T|$ and $\epsilon$ vary  (averaged over 100 instances). It can be observed that the game value is almost constant as $|T|$ varies for $\epsilon \in \{0.05,0.10,0.25\}$ and it is about 0.87. This is because all these instances have a similar number of edges, very close to the minimum number necessary for having connected graphs. With a larger number of edges, the game value increases. Interestingly, fixed a value of $\epsilon$, there is a threshold of $|T|$ such that beyond the threshold the game value increases as $|T|$ increases. This suggests that the minimum game value is obtained for connected graphs with the minimum number of edges.

In Tab.~\ref{tbl:times}, we report compute times with multiple signals, where the targets covered by a signal and the probability that a target triggers a signal are randomly chosen according to a uniform distribution. Values are averages over $100$ random instances and give insights on the computation effort along the considered dimensions.  The results show that the problem is computationally challenging even for a small number of targets and signals.

\begin{table}[h]
\begin{center}
\begin{tabular}{|c|c|c|c|}\hline
\backslashbox[1mm]{$m$}{$|T(s)|$} & $5$ & $10$ & $15$  \\ \cline{1-4}
2 & - & 17.83 & 510.61 \\ \cline{1-4}
3 & - & 33.00 & 769.30 \\ \cline{1-4}
4 & 0.55 & 35.35 & 1066.76  \\ \cline{1-4}
5 & 0.72 & 52.43 & 1373.32  \\ \cline{1-4}
\end{tabular}
\end{center}
\caption{Compute times (in seconds) for multi--signal instances.}
\label{tbl:times}
\end{table}

\subsubsection{Approximation algorithms}\label{ex:approx}
We evaluate the actual approximation ratios obtained with our approximation algorithms as $(1-\hat{g}_v) / (1-g_v)$, where $g_v$ is the expected utility of $\mathcal{A}$ at the equilibrium considering all the covering sets and $\hat{g}_v$ is the expected utility of $\mathcal{A}$ at the equilibrium when covering sets are generated by our heuristic algorithm. We execute our approximation dynamic programming algorithm with a different number, say RandRes, of randomly generated orders from $\{10,20,30,40,50\}$, in addition to the 3 heuristics discussed in Section~\ref{subsec:alg_dyn_progapproximate}. We executed our approximation branch and bound algorithm with constant values of $\rho$ from $\{0.25, 0.50, 0.75, 1.00\}$ (we recall that with $\rho = 1.00$ backtracking is completely disabled). 

\begin{figure}[!htbp]
\scriptsize
\begin{center}
\begin{tabular}{|c|c|c|}\hline
\begin{sideways}\hspace{2.25cm}$\epsilon = 0.05$\end{sideways}	& \hspace{-0.50cm}	\includegraphics[scale=0.52]{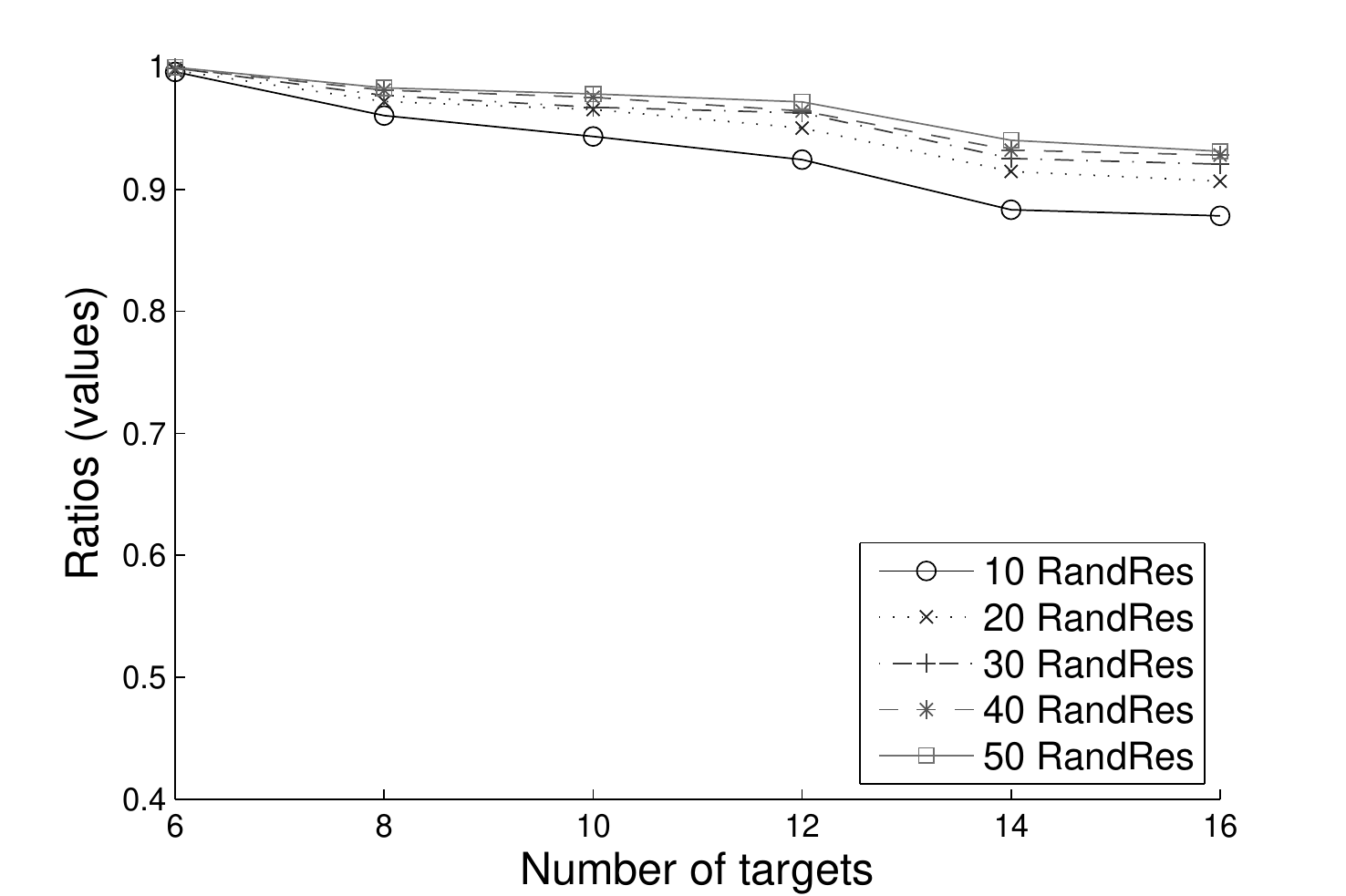}			\hspace{-0.55cm}	&	\hspace{-0.55cm}	\includegraphics[scale=0.52]{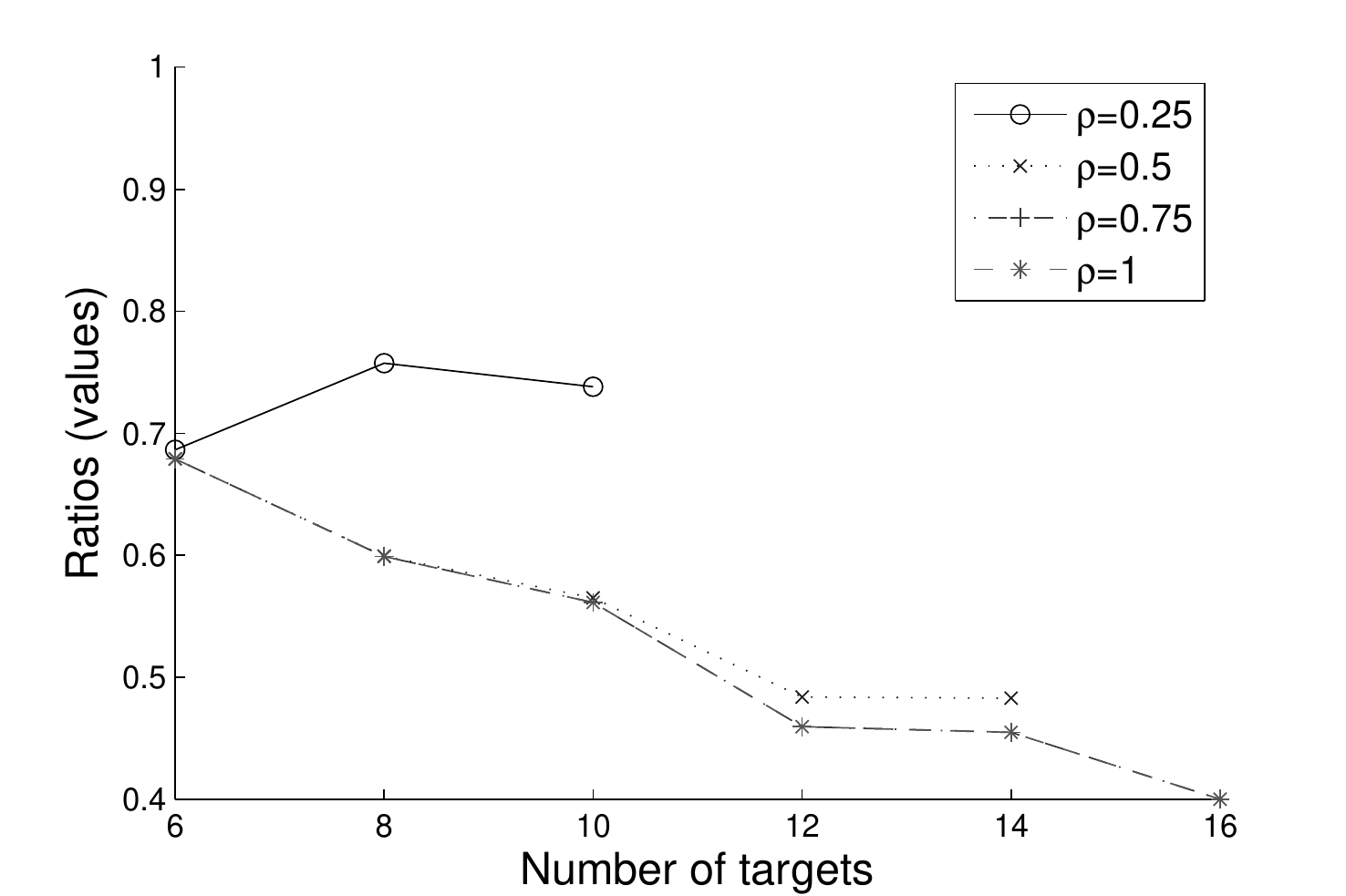}	\hspace{-0.8cm}	\\\hline
\begin{sideways}\hspace{2.25cm}$\epsilon = 0.10$\end{sideways}	& \hspace{-0.50cm}	\includegraphics[scale=0.52]{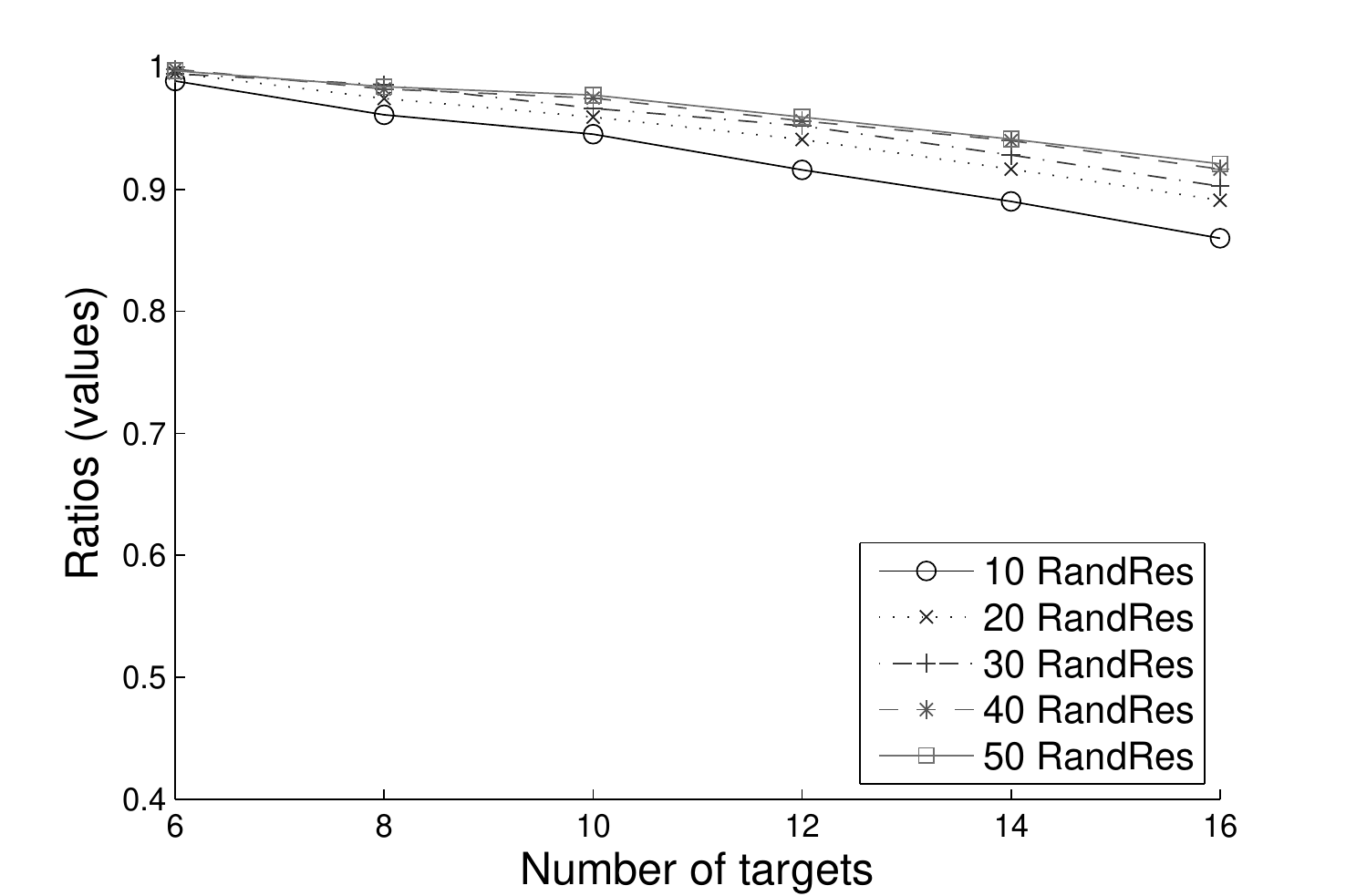}			\hspace{-0.55cm}	&	\hspace{-0.55cm}	\includegraphics[scale=0.52]{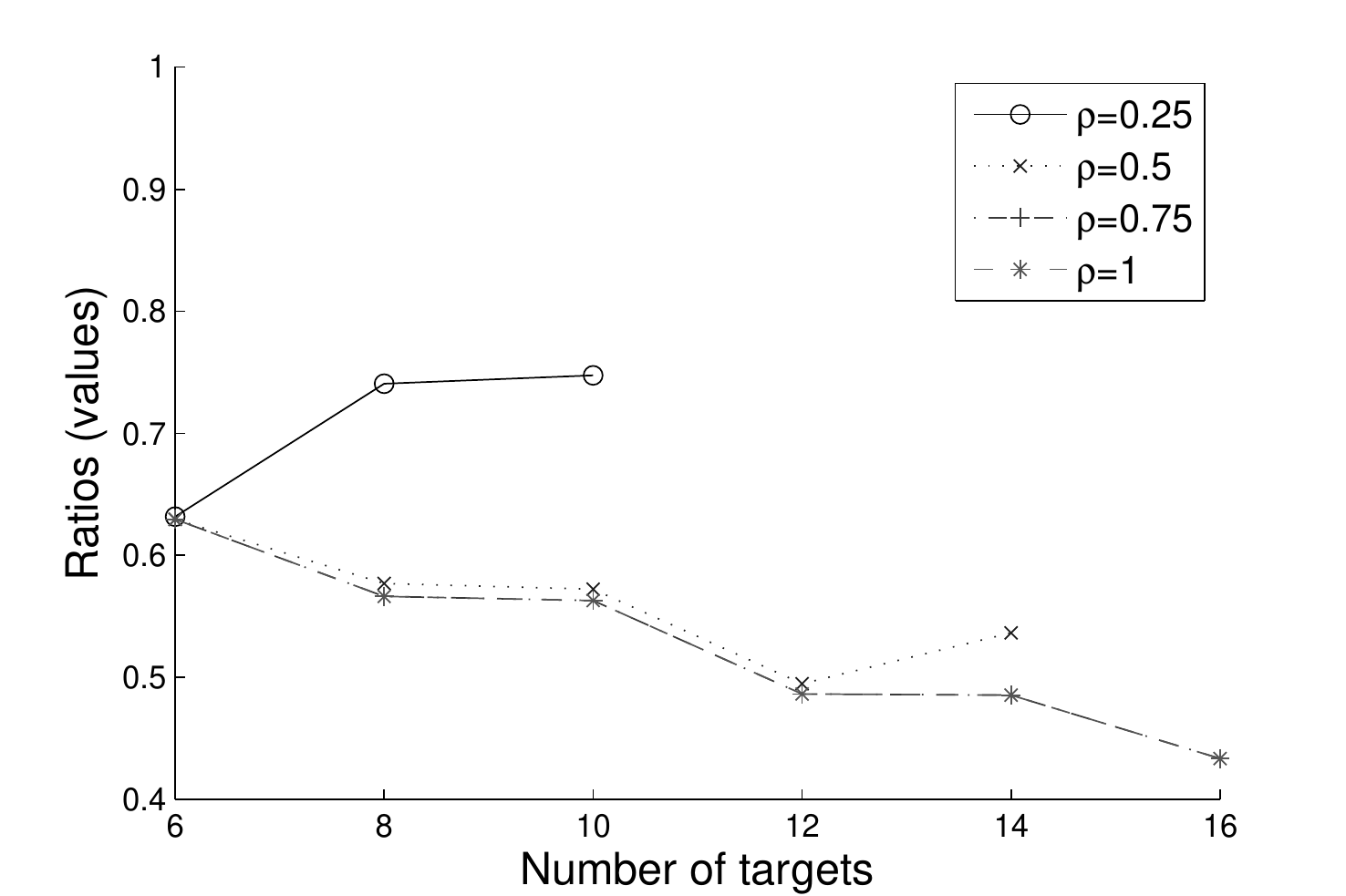}	\hspace{-0.8cm}	\\\hline
\begin{sideways}\hspace{2.25cm}$\epsilon = 0.25$\end{sideways}	& \hspace{-0.50cm}	\includegraphics[scale=0.52]{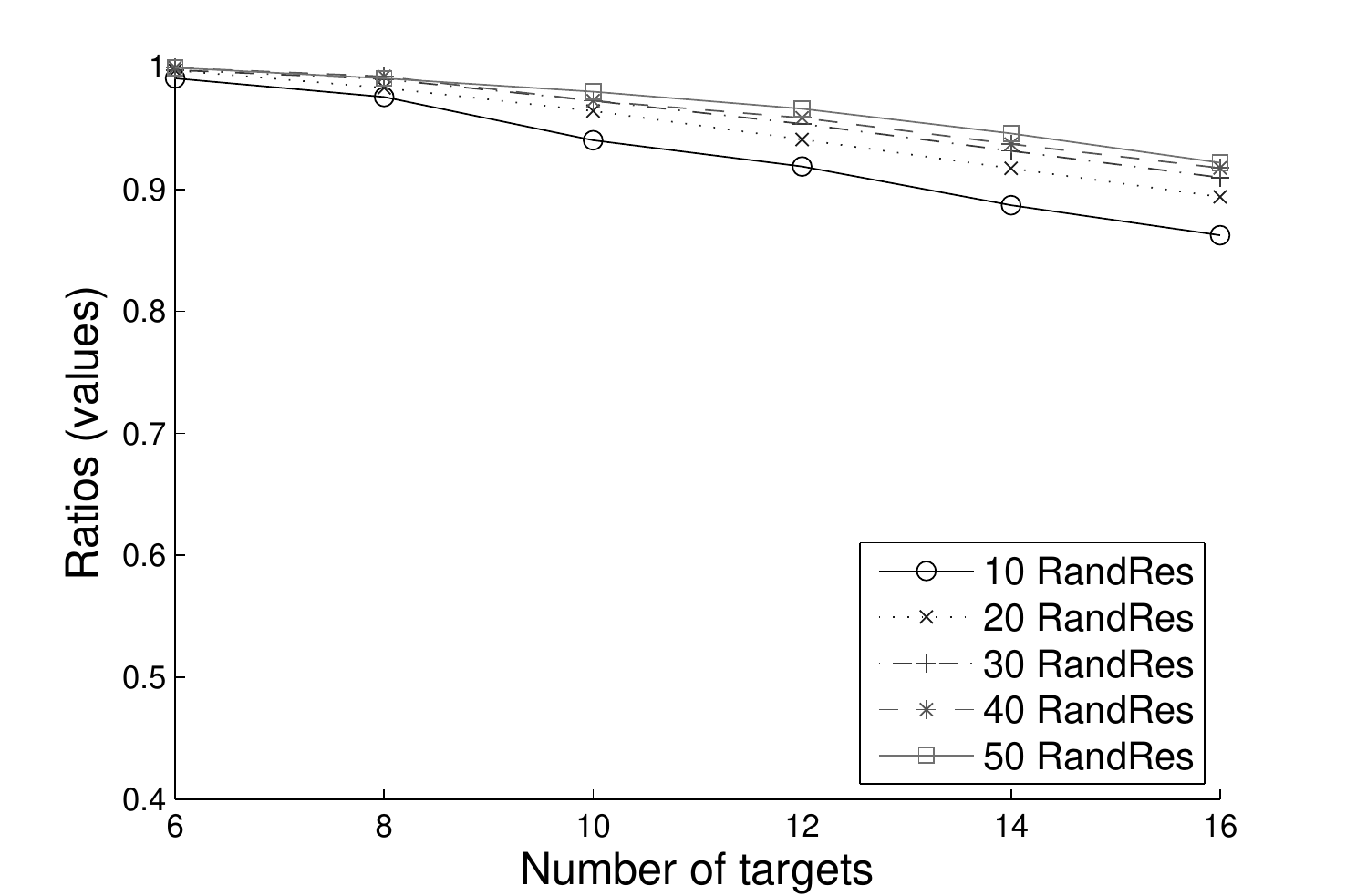}			\hspace{-0.55cm}	&	\hspace{-0.55cm}	\includegraphics[scale=0.52]{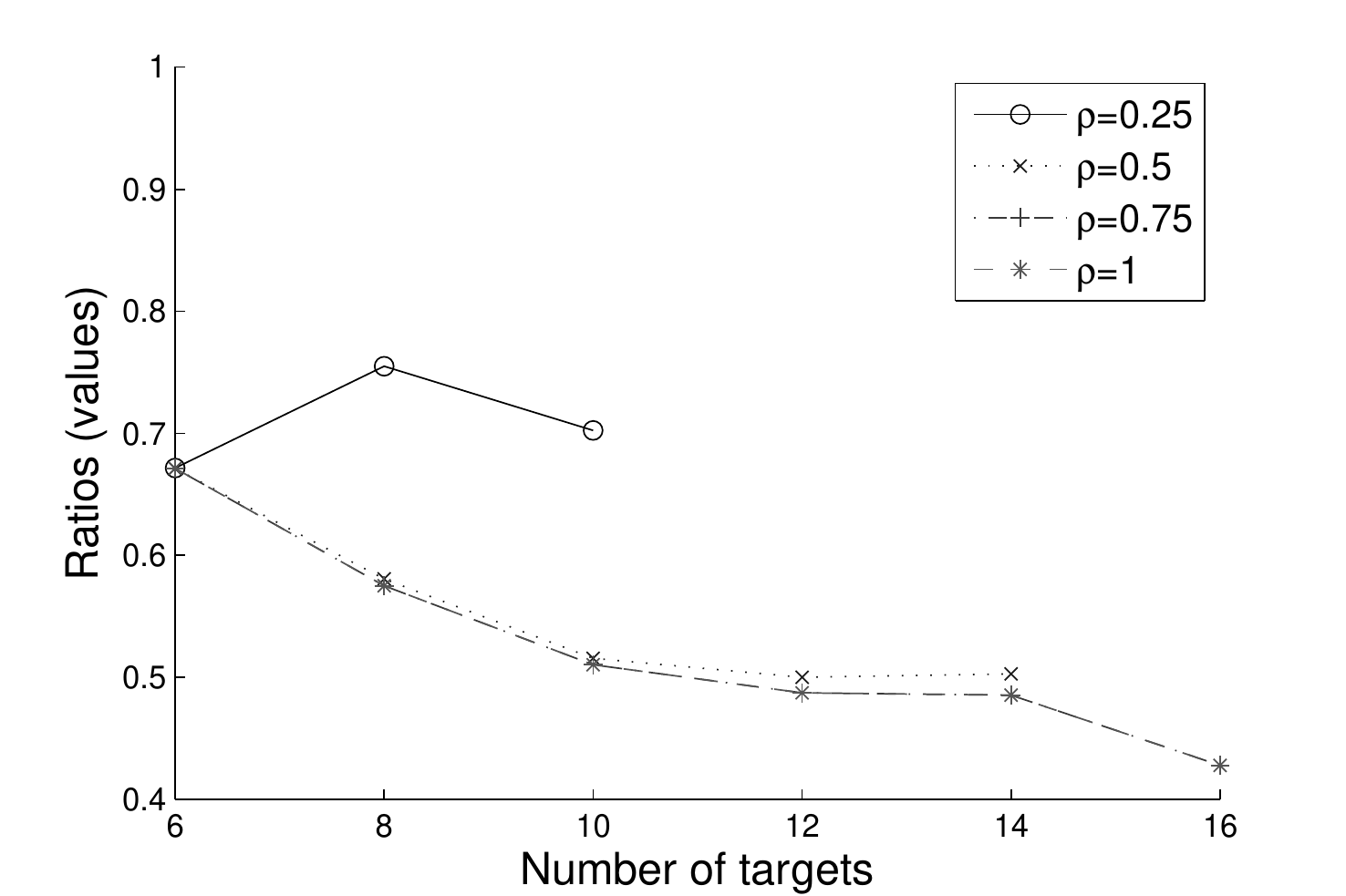}	\hspace{-0.8cm}	\\ \hline
													& 																							&									\\ 
													& Dynamic programming based approximation															&	Branch and bound based approximation	\\ 
													&																							&									\\\hline
\end{tabular}
\end{center}
\caption{Approximation ratios as $|T|$ varies.}
\label{fig:apx1}
\end{figure}

\begin{figure}[!htbp]
\scriptsize
\begin{center}
\begin{tabular}{|c|c|c|}\hline
\begin{sideways}\hspace{2.25cm}$\epsilon = 0.50$\end{sideways}	& \hspace{-0.50cm}	\includegraphics[scale=0.52]{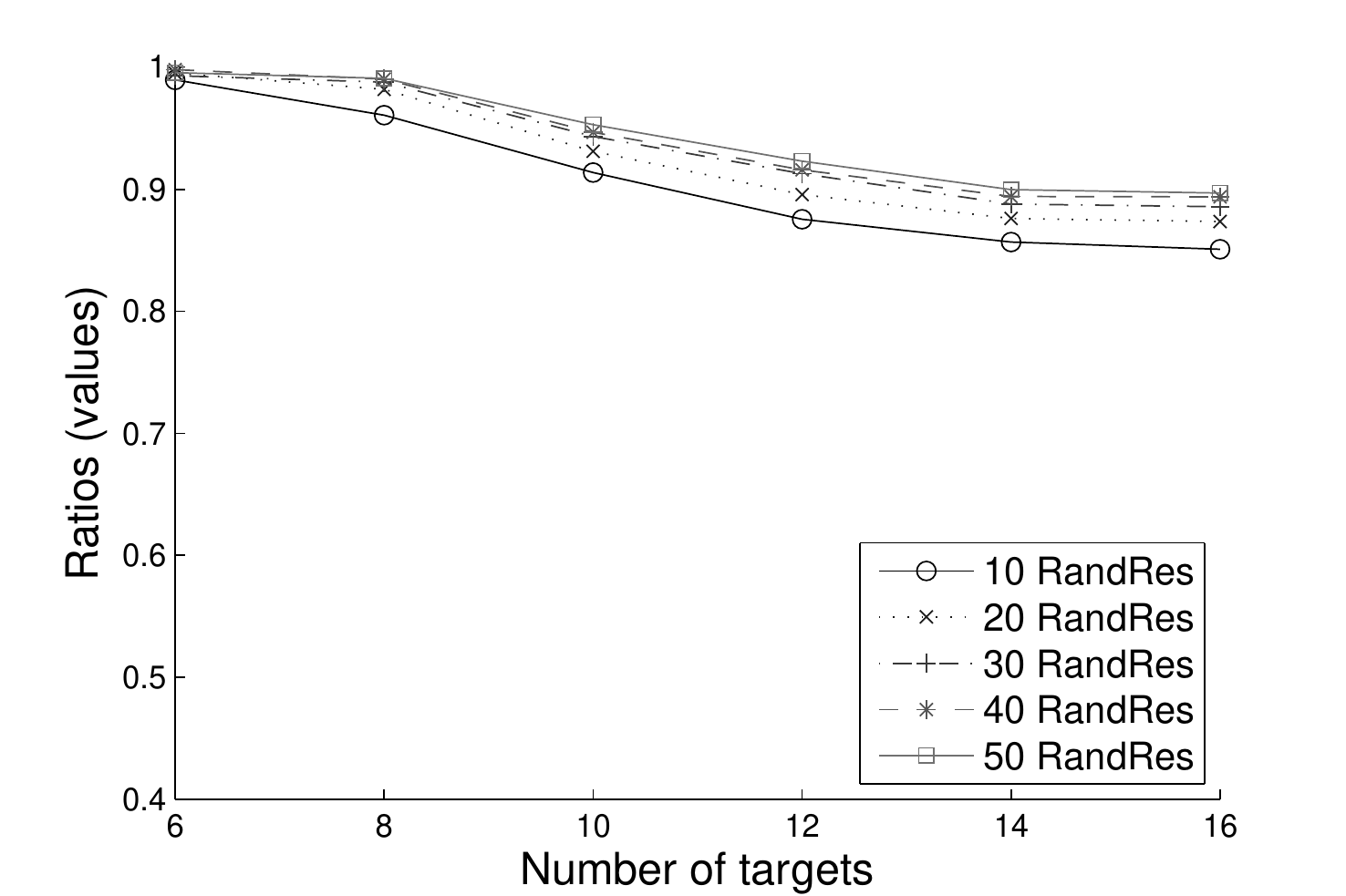}			\hspace{-0.55cm}	&	\hspace{-0.55cm}	\includegraphics[scale=0.52]{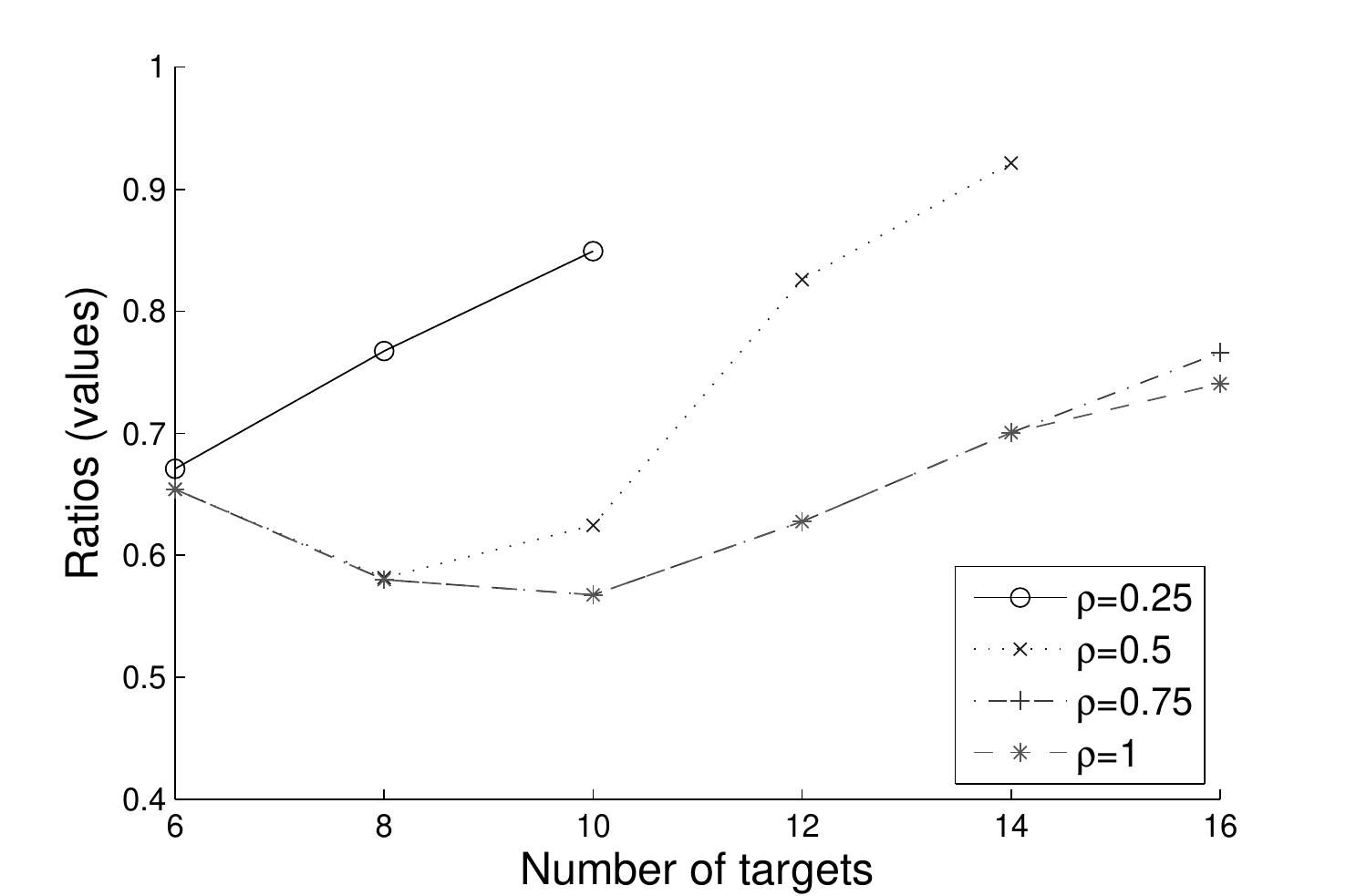}	\hspace{-0.8cm}	\\\hline
\begin{sideways}\hspace{2.25cm}$\epsilon = 0.75$\end{sideways}	& \hspace{-0.50cm}	\includegraphics[scale=0.52]{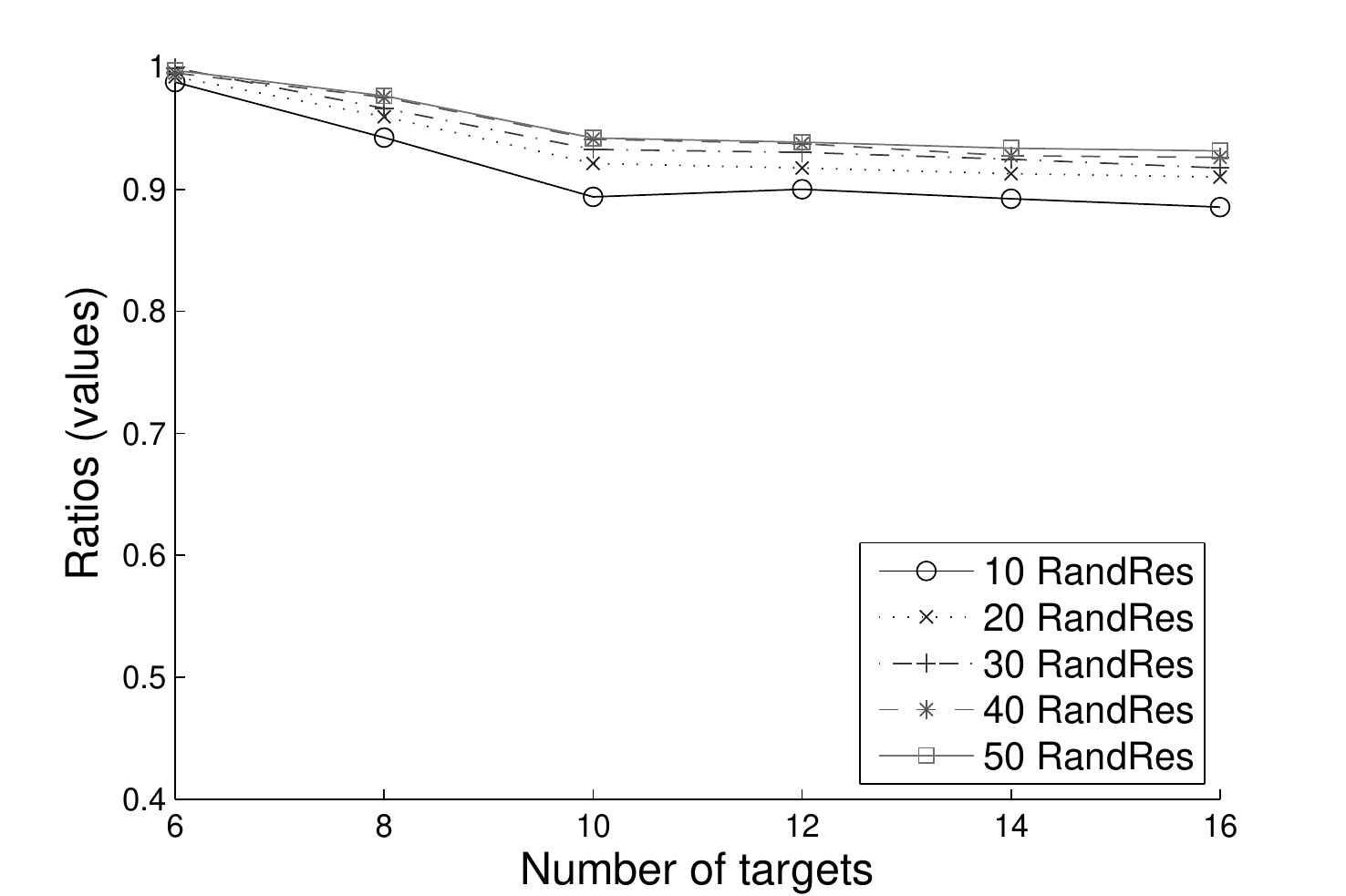}			\hspace{-0.55cm}	&	\hspace{-0.55cm}	\includegraphics[scale=0.52]{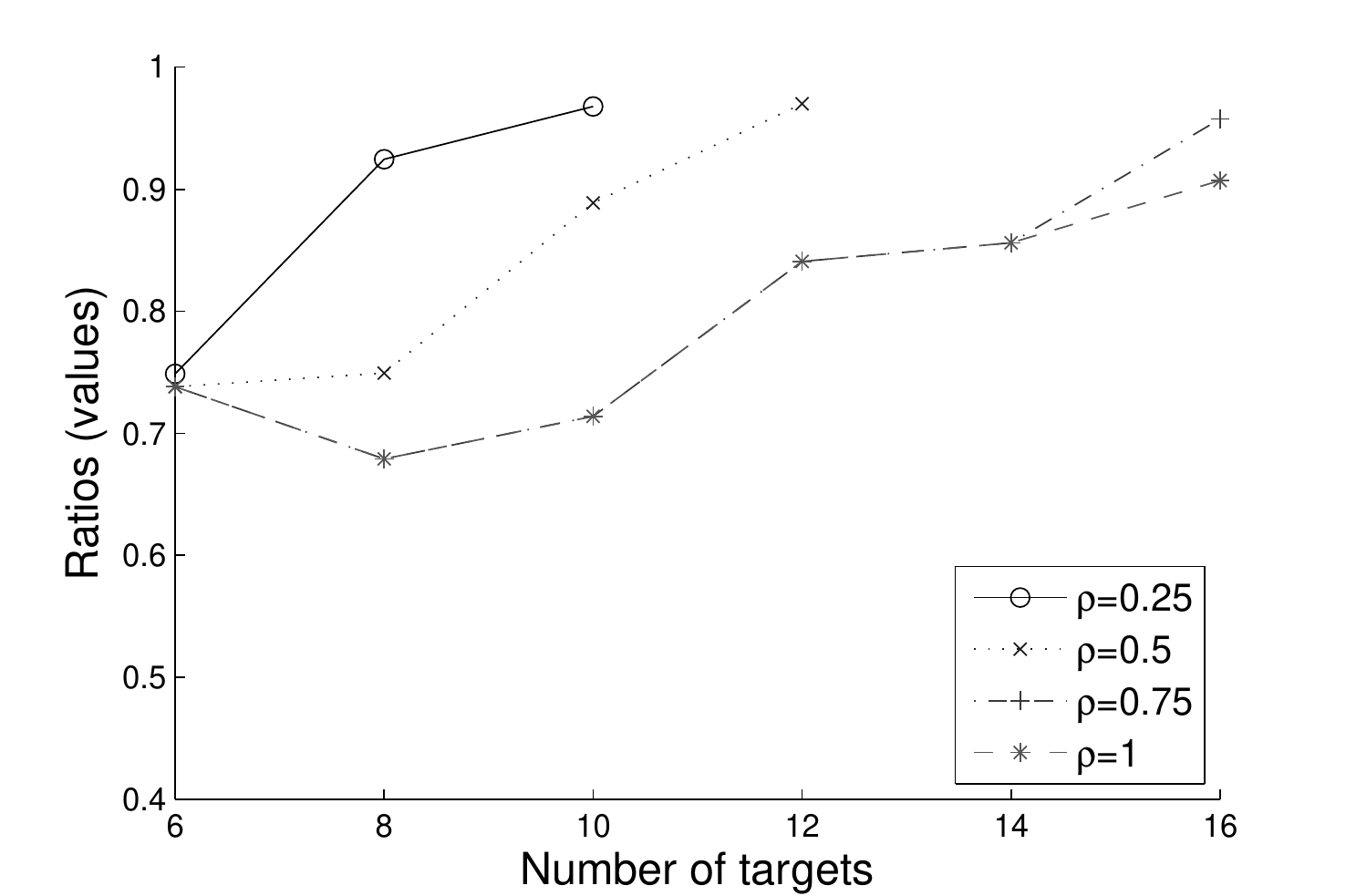}	\hspace{-0.8cm}	\\\hline
\begin{sideways}\hspace{2.25cm}$\epsilon = 1.00$\end{sideways}	& \hspace{-0.50cm}	\includegraphics[scale=0.52]{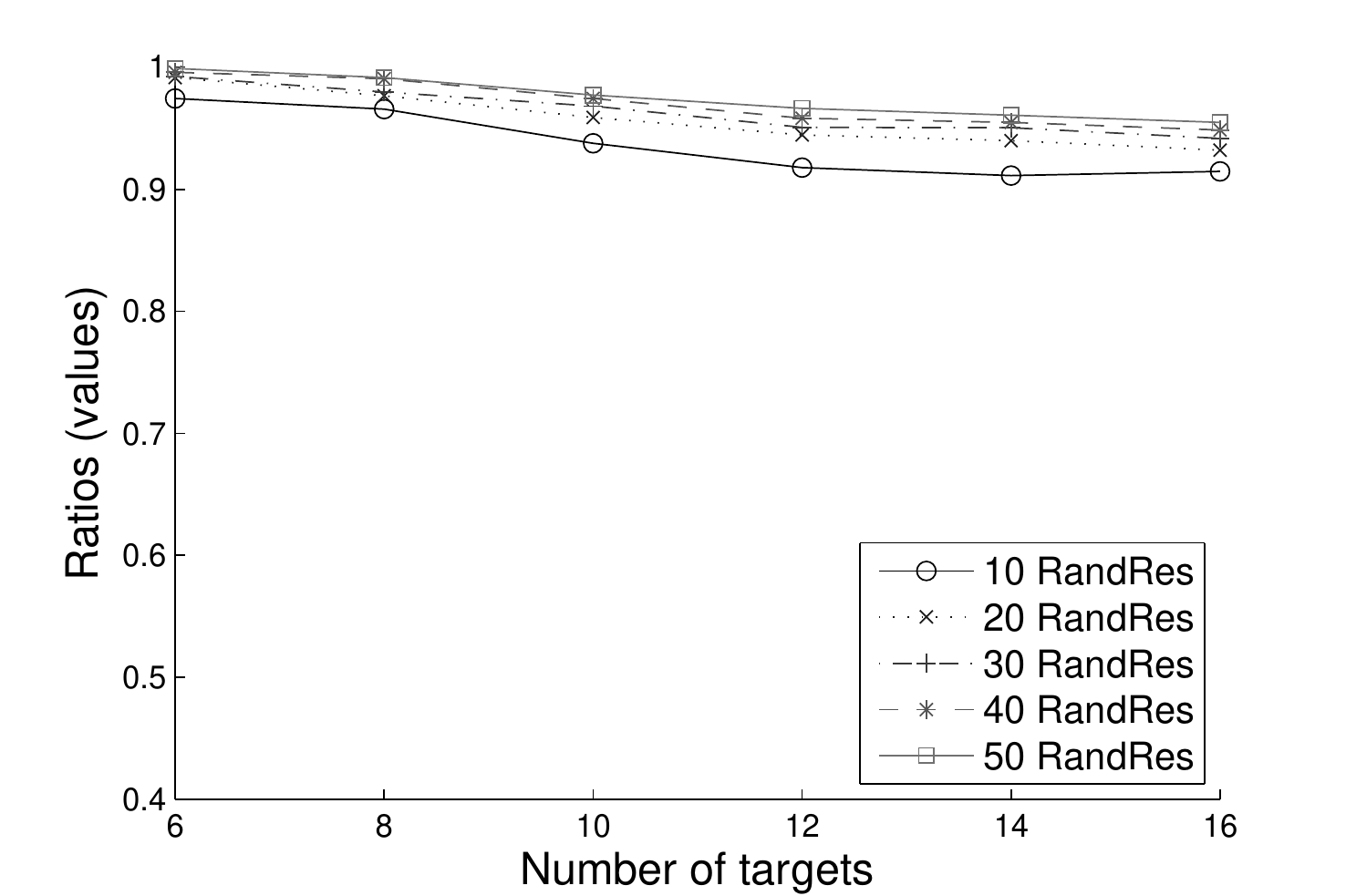}		\hspace{-0.55cm}	&	\hspace{-0.55cm}	\includegraphics[scale=0.52]{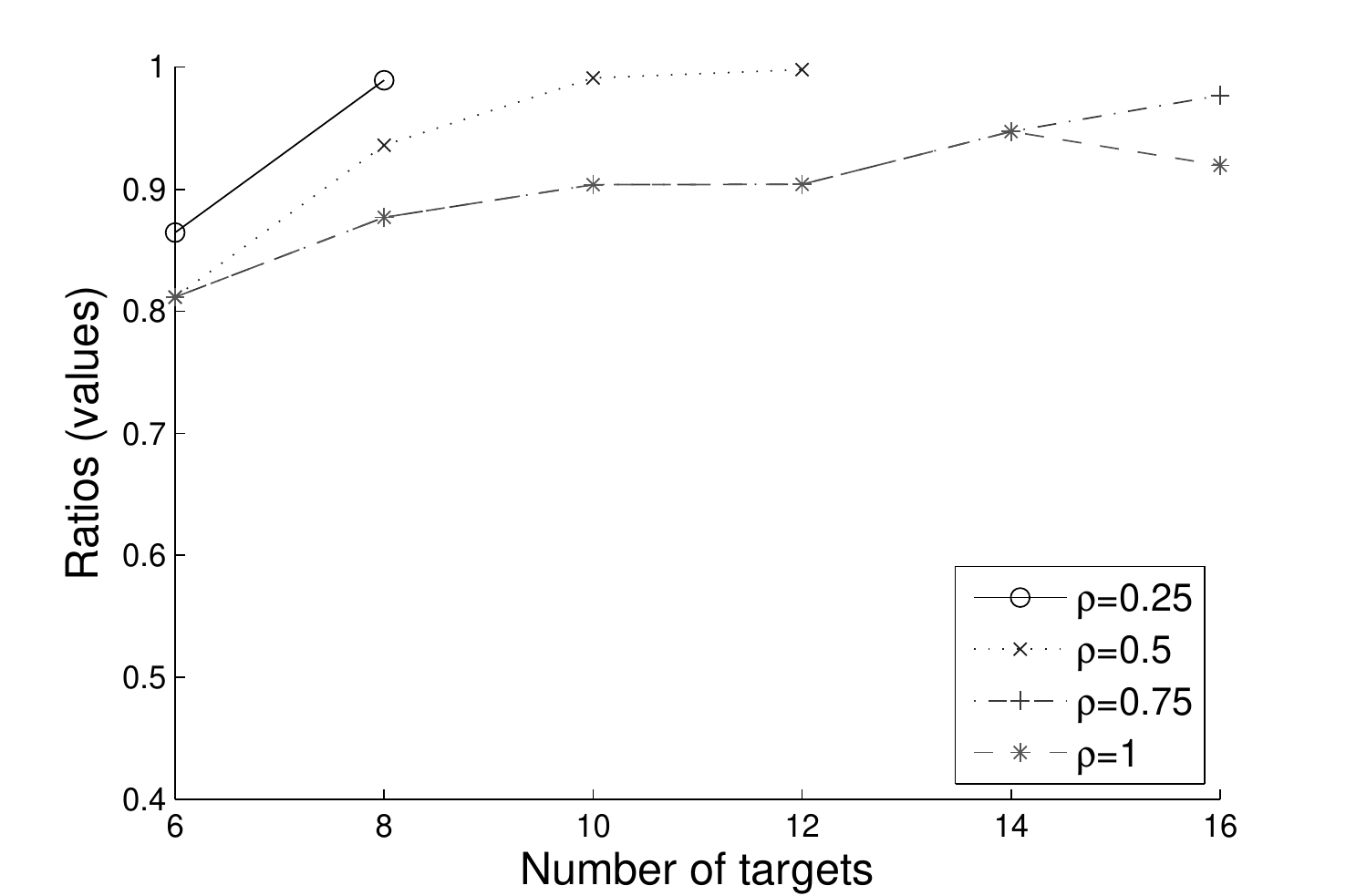}	\hspace{-0.8cm}	\\ \hline
													& 																							&									\\ 
													& Dynamic programming based approximation															&	Branch and bound based approximation	\\ 
													&																							&									\\\hline
\end{tabular}
\end{center}
\caption{Approximation ratios as $|T|$ varies.}
\label{fig:apx2}
\end{figure}

Figure~\ref{fig:apx1} and Figure~\ref{fig:apx2} report the actual approximation ratios (averaged over 100 instances) obtained with our approximation algorithms for different values of $|T|\in \{6,8,10,12,14,16\}$, i.e., the instances for which we know the optimal game value, and $\epsilon \in \{0.05, 0.10,0.25,0.50,0.75,1.00\}$. We remark that the ratios obtained with the approximation branch--and--bound algorithm for some values of $\rho$ are omitted. This is because the compute time needed by the algorithm is over $10^4$ seconds. The algorithm always terminates by the deadline for only $\rho \in \{0.75, 1.00\}$. We focus on the ratios obtained with the dynamic programming algorithm. Given a value of $\epsilon$, as $|T|$ increases, the ratio decreases up to a given threshold of $|T|$ and then it is a constant. The threshold increases as $\epsilon$ decreases, while the constant decreases as $\epsilon$ decreases. The value of the constant is high for every $\epsilon$, being larger than $0.8$. Although the ratios increase as RandRes increases, it is worth noting that the increase is not very significant, being of the order of $0.05$ between 10 RandRes and 50 RandRes. We focus on the ratios obtained with the branch--and--bound algorithm. Given a value of $\epsilon$, as $|T|$ increases, the ratio decreases up to a given threshold of $|T|$ and then it increases approaching a ratio of 1. The threshold increases as $\epsilon$ decreases, while the minimum ratio decreases as $\epsilon$ decreases. Interestingly, ratios with $\rho = 1.00$ are very close to ratios with $\rho \in 0.75$, showing that performing even significant backtracking around the solution found with $\rho = 1.00$ does not lead to a significant improvement of the solution. The solution can be effectively improved only with $\rho = 0.25$, but it is not affordable due to the excessive required compute time. This shows that the heuristic performs very well. Comparing the ratios of the two algorithms, it can be observed that the approximation dynamic programming algorithm performs better than the approximation branch--and--bound algorithm. While the dynamic programming one always provides a ratio larger than 0.8, the branch--and--bound one provides for combinations of $|T|$ and $\epsilon$ ratios lower than 0.4.

\begin{figure}[!htbp]
\begin{center}
\scriptsize
\begin{tabular}{cc}
$\epsilon = 0.05$		&		$\epsilon = 0.25$	\\
\hspace{-0.5cm}\includegraphics[scale=0.45]{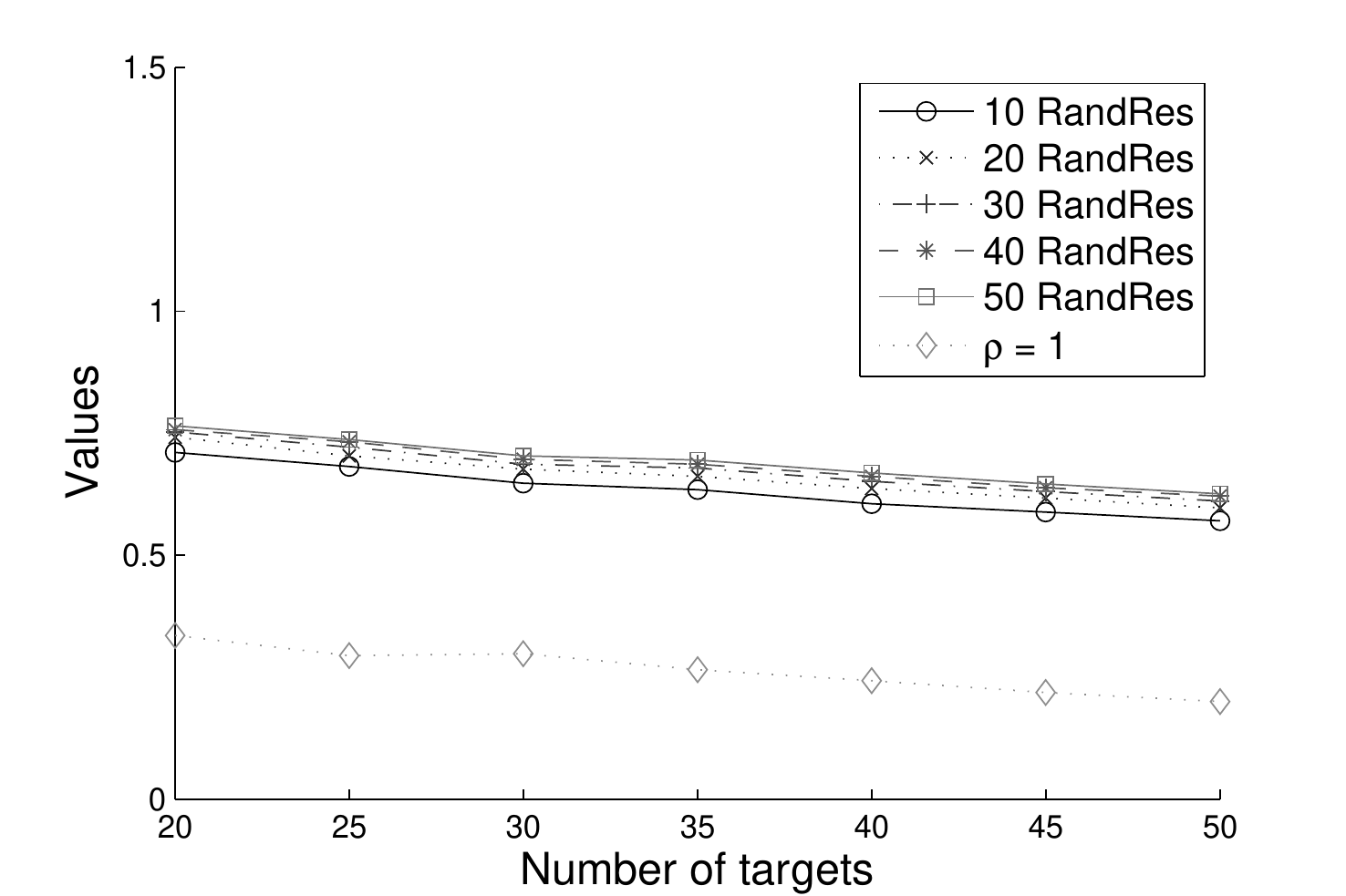}\hspace{-1.00cm}&\hspace{-1.00cm}\includegraphics[scale=0.45]{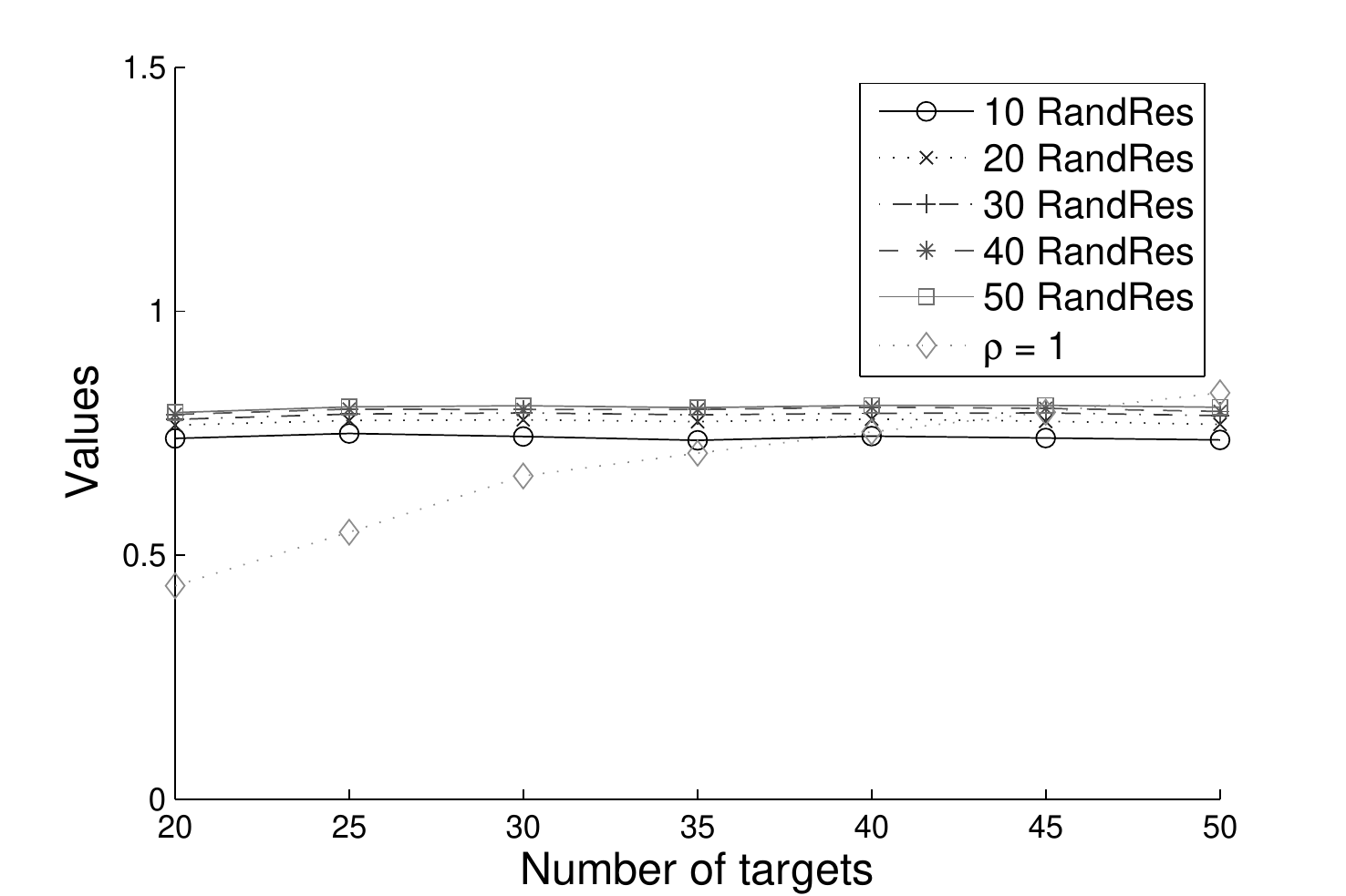}\hspace{-0.5cm}\\
\end{tabular}
\caption{Game values as $|T|$ varies.}
\label{fig:apxbeyondratio}
\end{center}
\end{figure}

Figure~\ref{fig:apxbeyondratio} reports the game values obtained with the approximation dynamic programming algorithm for every value of RandRes and with the approximation branch--and--bound algorithm when $|T|\in \{20,25,30,35,40,45,50\}$ only for $\rho = 1.00$. Indeed, with $\rho = 0.75$ the compute time is excessive and, as shown above, the purely heuristic solution cannot be significantly improved for $\epsilon\geq0.75$. We report experimental results only for $\epsilon \in \{0.05,0.25\}$. We notice that for these instances we do not have the optimal game value. However, since the optimal game value cannot be larger than 1 by construction of the instances, the game value obtained with our approximation algorithms represents a lower bound to the actual approximation ratio. It can be observed that, given a value of $\epsilon$, the ratios obtained with the dynamic programming algorithm are essentially constant as $|T|$ increases and this constant reduced as $\epsilon$ reduces. Surprisingly, after a certain value of $|T|$, the game values obtained with the branch and bound algorithm are higher than those obtained with the dynamic programming algorithm. This is because, fixed a value of $\epsilon$, as $|T|$ increases, the problem becomes easier and the heuristic used by the branch and bound algorithm performs well finding the best covering routes. This shows that there is not an algorithm outperforming the other for every combination of parameters $|T|$ and $\epsilon$. Furthermore, the above result shows that the worst cases for the approximation algorithms are those in which $\epsilon = O(\frac{1}{|T|})$, corresponding to instances in which the number of edges per vertex is a constant in $|T|$. It is not clear from our experimental analysis whether increasing $|T|$ with $\epsilon = \frac{\nu}{|T|}$ for some $\nu>1$ the game value approaches to 0 or to a strictly positive value. However, our approximation algorithms provide a very good approximation even with a large number of targets and a small value of $\epsilon$.

\begin{figure}[!htbp]
\begin{center}
\scriptsize
\begin{tabular}{cc}
$\epsilon = 0.05$		&		$\epsilon = 0.25$	\\
\hspace{-0.5cm}\includegraphics[scale=0.45]{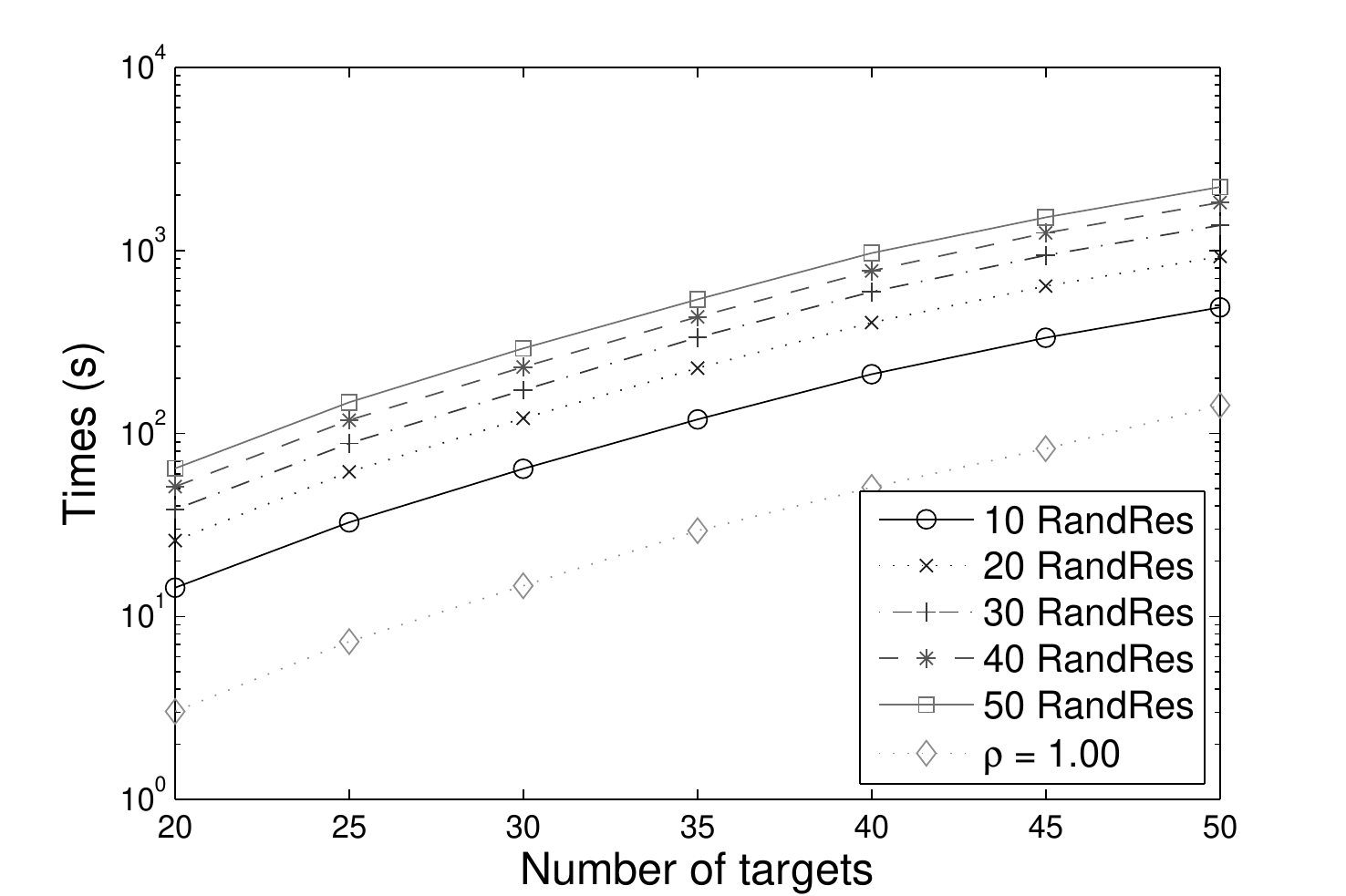}\hspace{-1.00cm}&\hspace{-1.00cm}\includegraphics[scale=0.45]{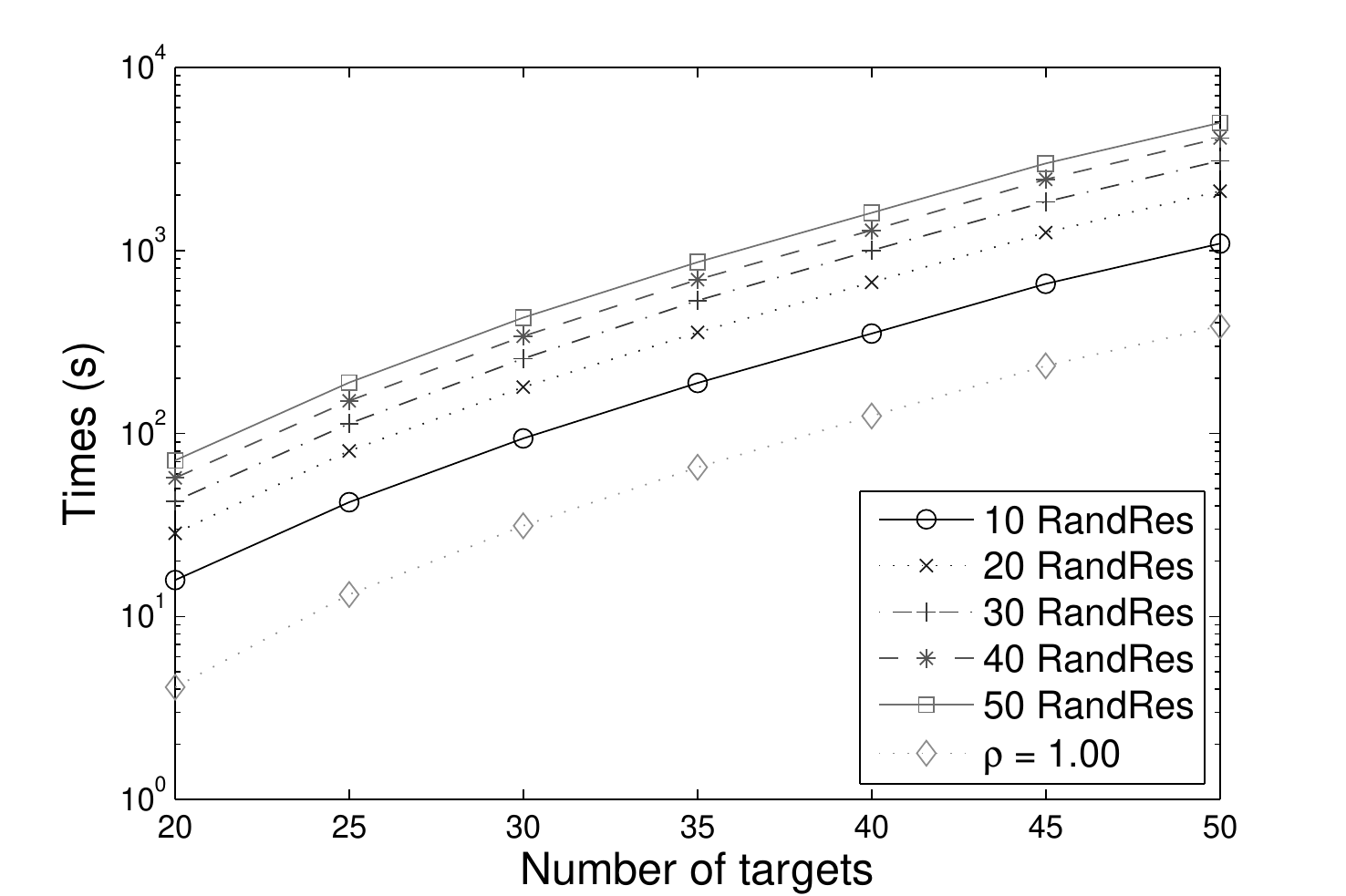}\hspace{-0.5cm}\\
\end{tabular}
\caption{Time ratios as $|T|$ varies.}
\label{fig:apxbeyondtimes}
\end{center}
\end{figure}

Figure~\ref{fig:apxbeyondtimes} reports the compute times required by the approximation dynamic programming algorithms. As it can be seen, the required time slightly increases when adopting a larger number of randomly generated orders with respect to the baseline with $\rho=1.00$.

\begin{figure}[!htbp]
\begin{center}
\scalebox{0.84}{ 
\begin{sideways}
\scriptsize
\begin{tabular}{cc}
Real map		&		Graph on real map	\\
\includegraphics[scale=0.063]{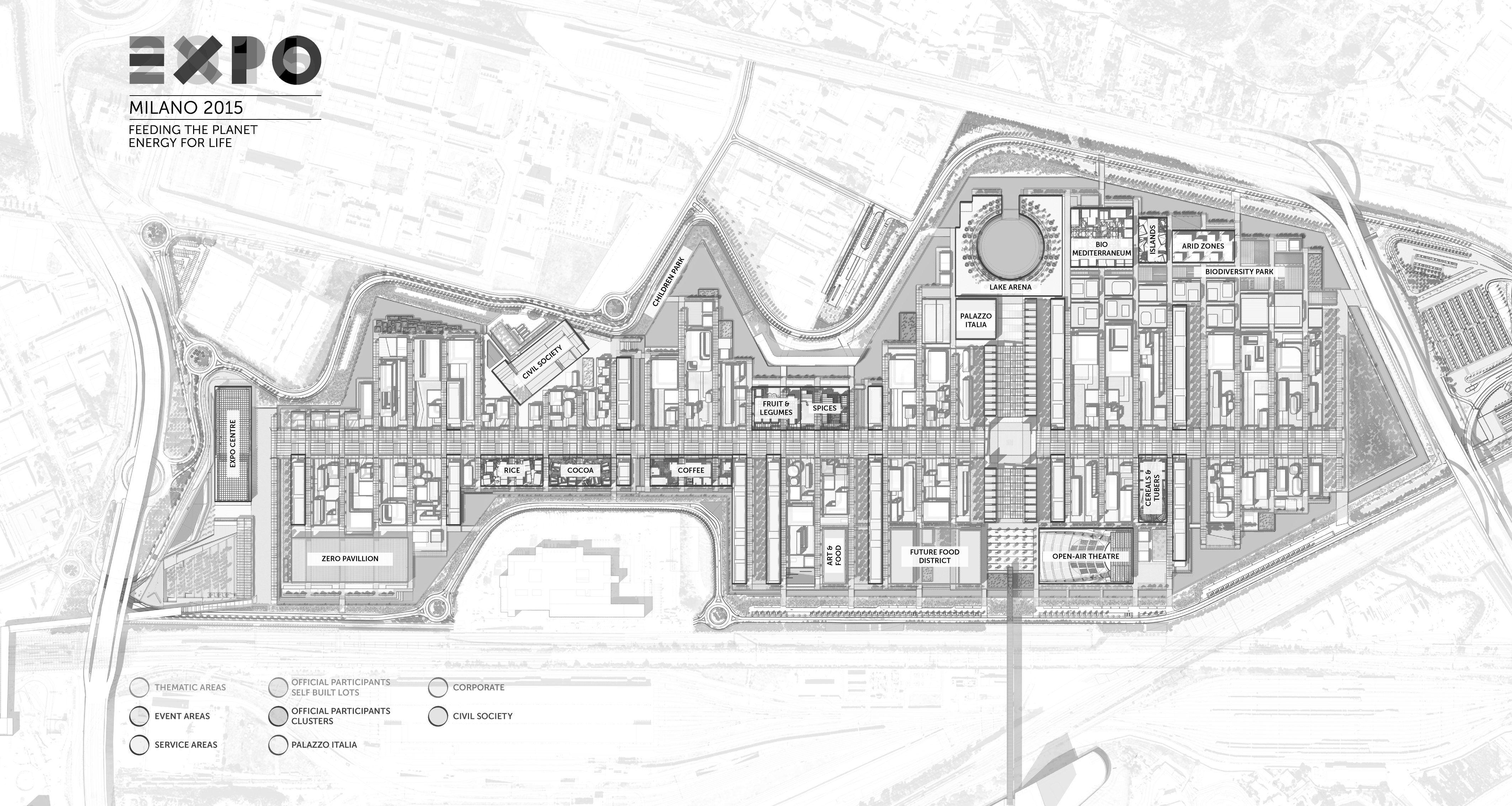}& \includegraphics[scale=0.3]{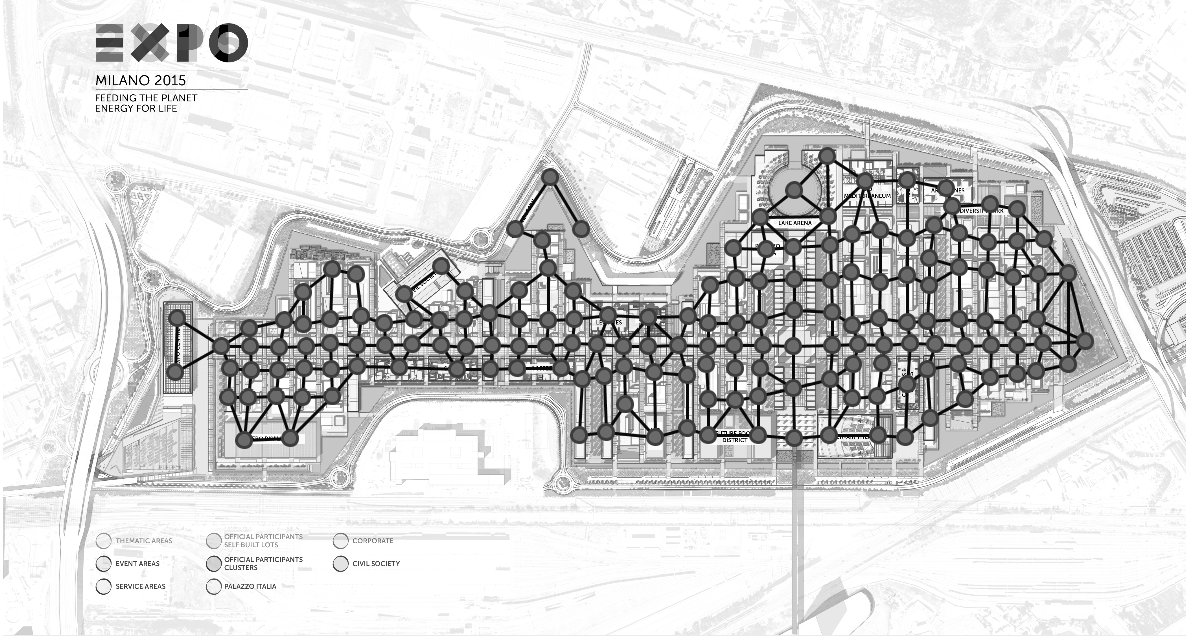}\\

Values		&		Deadlines	\\
\includegraphics[scale=0.37]{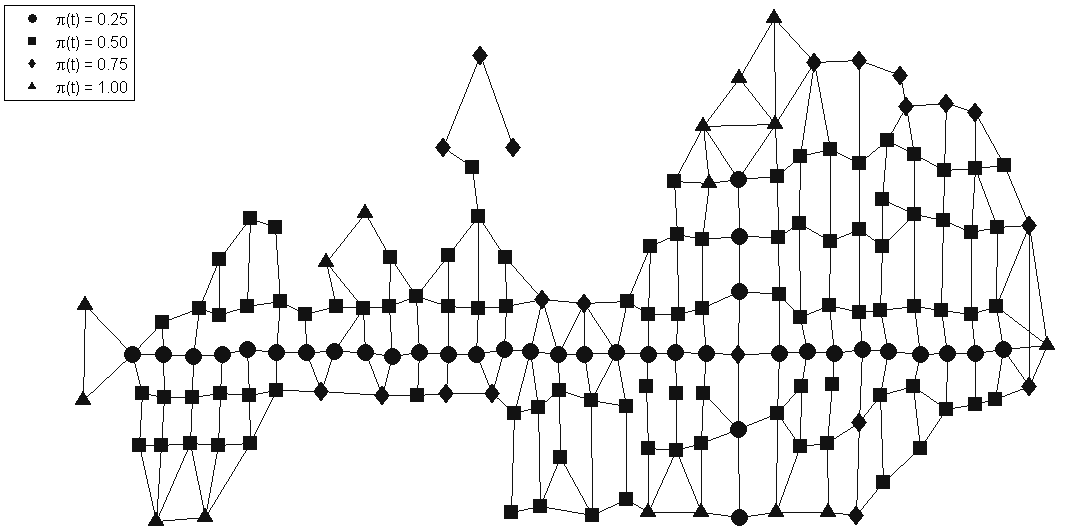} &\includegraphics[scale=0.37]{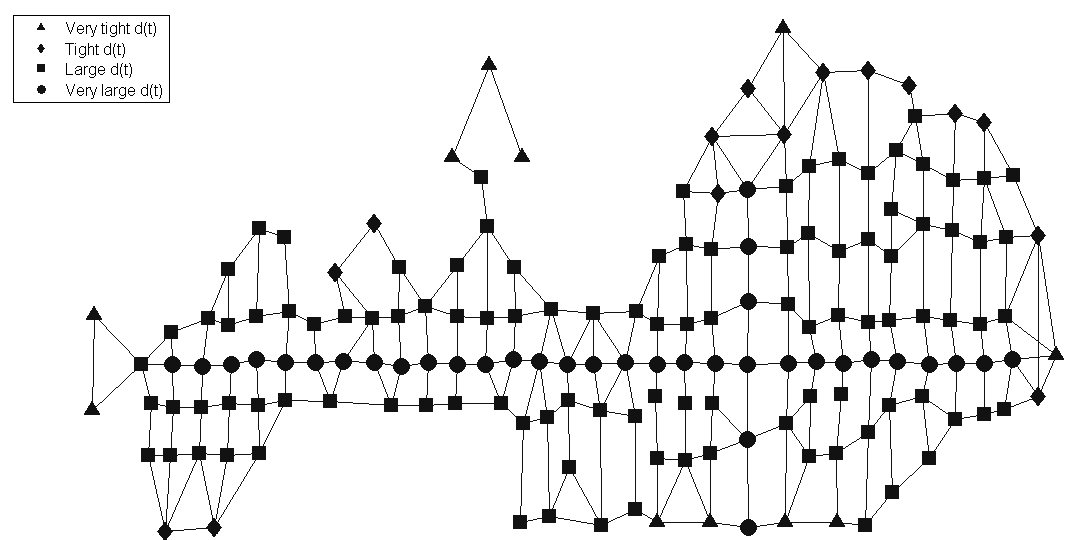}\\
\end{tabular}
\end{sideways}}
\end{center}
\caption{Expo 2015 instance.}
\label{fig:expo_instance}
\end{figure}

\subsection{Real case study}
\label{subsection:realisticexp}

In this section we present the results obtained by applying our approach to a real case study, in order to show an example of real application of our model. We imagine to face the task of protecting a fair site as we already discussed in Section~\ref{sec:expo} and we focus on the particular setting of Expo 2015.  Figure~\ref{fig:expo_instance} shows the map of the Expo 2015 site together with its graph representation. We manually build a discretized version of the site map by exploiting publicly available knowledge of the event\footnote{Detailed information can be found at \url{http://www.expo2015.org/}.}. We identify $\approx 170$ sensible locations which correspond to an equal number of targets in our graph. More specifically, we identify $\approx 130$ targets located at the entrances of each pavilion and in the surroundings of those areas which could be of interest for a high number of visitors. Some targets ($\approx 35$) are located over the main roads, being these critical mainly due to their high crowd. Such roads also define our set of edges which resulted in a density of $\approx 0.02$. Figure~\ref{fig:expo_instance} reports a graphical representation of chosen deadlines $d(\cdot)$ and values $\pi(\cdot)$, respectively. To determine such values in a reasonable way we apply a number of simple rules of thumb. First, to ease our task, we discretize the spaces of possible deadlines and values in four different levels. To assign a value to a target, we estimate the interest (in terms of popularity and expected crowd) of the corresponding area in the fair site. The higher the interest, the higher the value (actual values are reported in the figure). To assign deadlines we estimate the time an attacker should spend to escape from the attacked target after some malicious activity is started (for example, blending into the crowd without leaving any trace). In particular, we estimate a smaller escape time for those locations lying near the external border of the fair site while for locations that are more central we estimated a larger time. The smaller the escape time, the tighter the deadline for that target. Actual values are extracted from a normal distribution where $\sigma^2=1$ and $\mu$ is set according to the chosen level. The maximum distance between any two target locations is about $1.5$Km which we assume can be be covered in about $7.5$ minutes (we imagined a crowded scenario). Given such reference scenario, our means span from $5$ minutes (very tight) to $7.5$ minutes (very large). To derive our alarm system model we assume to have a number of panoramic cameras deployed in the environment at locations we manually choose in order to cover the whole environment and to guarantee a wide area of view for each camera (i.e., trying to keep, in general, more than one target under each camera's view). To map our set of cameras over the alarm system model, we adopt this convention: each group of cameras sharing an independent partial view of a target $t$ is associated to a signal $s \in S(t)$; if target $t$ is covered by $k$ signals then each signal is generated with probability $1/k$ once $t$ is attacked. Obviously, a deeper knowledge of the security systems deployed on the site can enable specific methods to set the parameters of our model. This is why we encourage involving agencies in charge of security when dealing with such task. 

\begin{figure}[!htbp]
\centering
\begin{tabular}{c}
\includegraphics[scale=0.45]{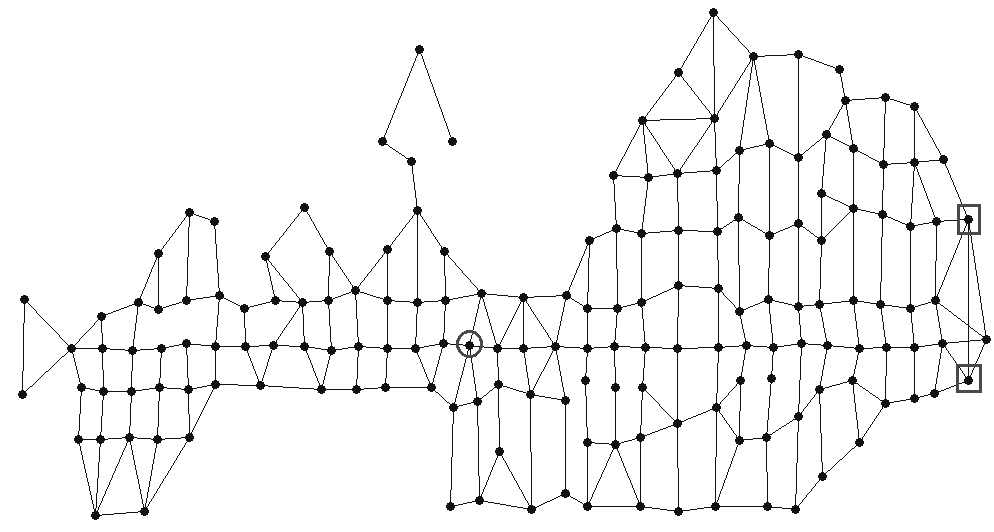}
\end{tabular}
\caption{Best placement and attack locations.}
\label{fig:expo_best_and_attack}
\end{figure}

\begin{figure}[!htbp]
\centering
\begin{tabular}{c}
\includegraphics[scale=0.50]{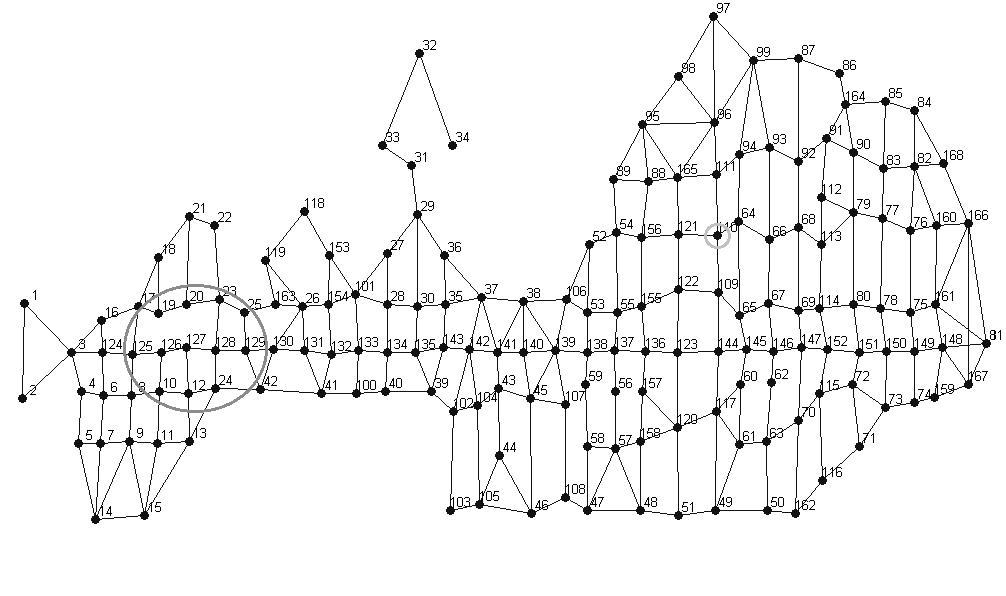}
\end{tabular}
\begin{scriptsize}
\begin{tabular}{| c | c |}
 \hline
    Covering set & Probability \\ \hline
    \{19, 20, 23, 25, 125, 126, 127, 128\} & $0.0194$   \\ \hline
    \{10, 12, 23, 24, 25, 126, 127, 128\} & 0.0231 \\ \hline
    \{10, 12, 24, 25, 126, 127, 128, 129\} & 0.0333 \\ \hline
    \{12, 23, 24, 25, 126, 127, 128, 129\} & 0.0494 \\ \hline
    \{10, 12, 23, 24, 25, 125, 126, 128\} & 0.0344 \\ \hline
    \{10, 12, 24, 25, 125, 126, 128, 129\} & 0.0488 \\ \hline
    \{10, 12, 25, 125, 126, 127, 128, 129\} & 0.0493 \\ \hline
    \{12, 23, 24, 25, 125, 126, 127, 128\} & 0.0502 \\ \hline
    \{12, 24, 25, 125, 126, 127, 128, 129\} & 0.0692 \\ \hline
    \{19, 20, 23, 25, 125, 126, 129\} & 0.0492 \\ \hline
    \{19, 20, 23, 125, 126, 128, 129\} & 0.0492 \\ \hline
    \{20, 23, 25, 126, 127, 128, 129\} & 0.0657 \\ \hline
    \{10, 23, 25, 125, 126, 127, 128\} & 0.0662 \\ \hline
    \{19, 20, 23, 25, 125, 128, 129\} & 0.0412 \\ \hline
    \{23, 25, 125, 126, 127, 128, 129\} & 0.1146 \\ \hline
    \{20, 23, 24, 25, 127, 128\} & 0.0645 \\ \hline
    \{10, 12, 24, 125, 126, 127\} & 0.0877 \\ \hline
    \{20, 23, 24, 25, 128, 129\} & 0.0846 \\ \hline
\end{tabular}
\end{scriptsize}
\caption{Example of response strategies to signal.}
\label{fig:expo_strategies}
\end{figure}

We first show a qualitative evaluation of our method. Figure~\ref{fig:expo_best_and_attack} depicts the best placement for the Defender (the circle in the figure) and the attacked targets (the squares in the figure, these are the actions played by the Attacker with non--null probability at the equilibrium). As intuition would suggest, the best location from where any signal response should start is a central one w.r.t. the whole fair site. Our simulations show that the optimal patrolling strategy coincides with such fixed placement even under false negatives rates of at least $\approx 0.3$. Notice that such false negatives value can be considered unrealistically pessimistic for alarm systems deployed in structured environment like the one we are dealing with. Attacked targets correspond to areas, which exhibit rather high interest and small escape time. Figure~\ref{fig:expo_strategies} reports an example of signal response strategy for a given starting vertex (the small circle in the figure) and a given signal (whose covered targets are depicted with the large circle in the figure). The table lists the computed covering sets and the probabilities with which the Defender plays the corresponding covering routes.

\begin{figure}[!htbp]
\centering
\subfigure[Time boxplots by signal]{
\includegraphics[scale=0.2]{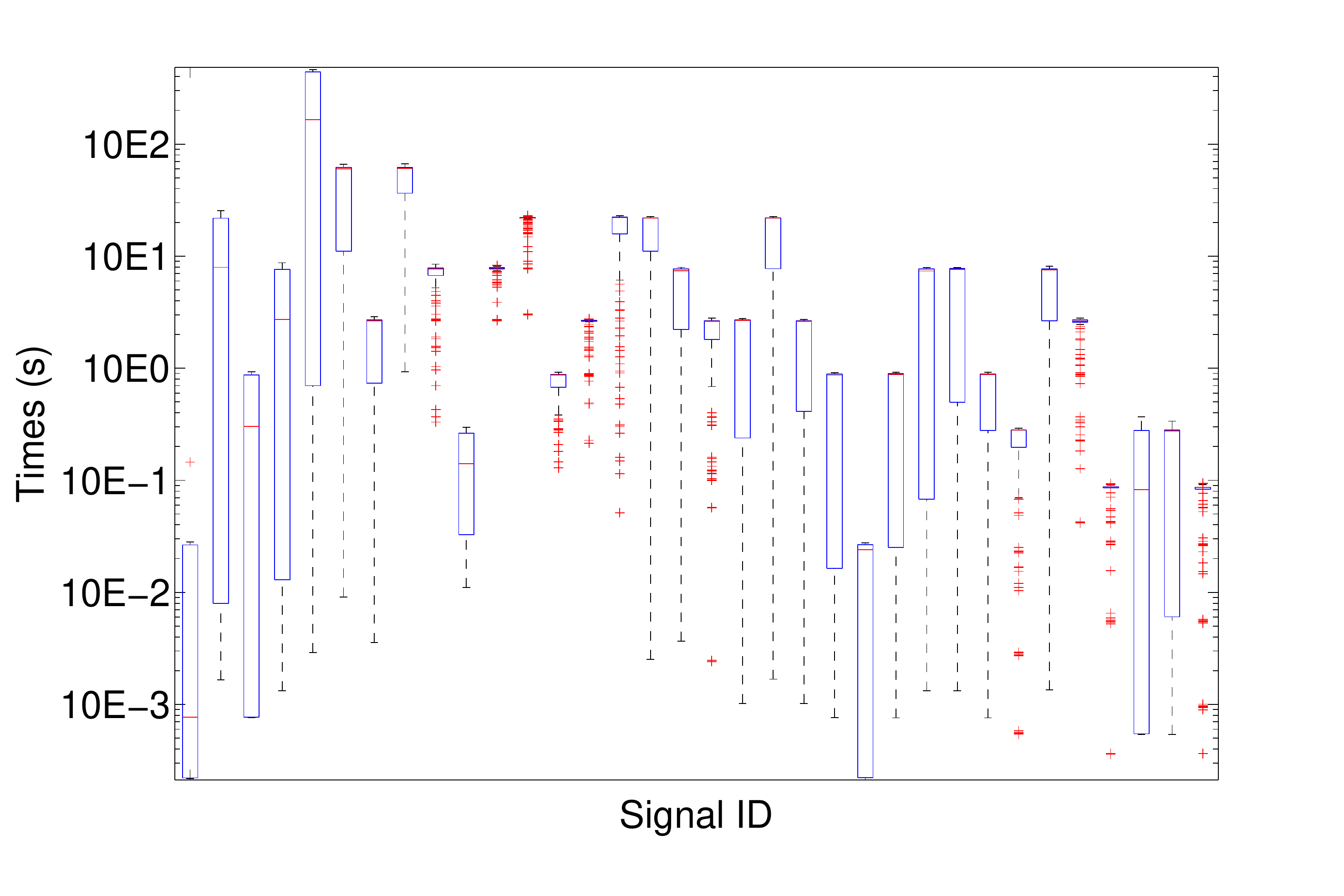}\label{fig:expo_boxplokts_by_signal}
}
\subfigure[Time boxplots by node]{
\includegraphics[scale=0.2]{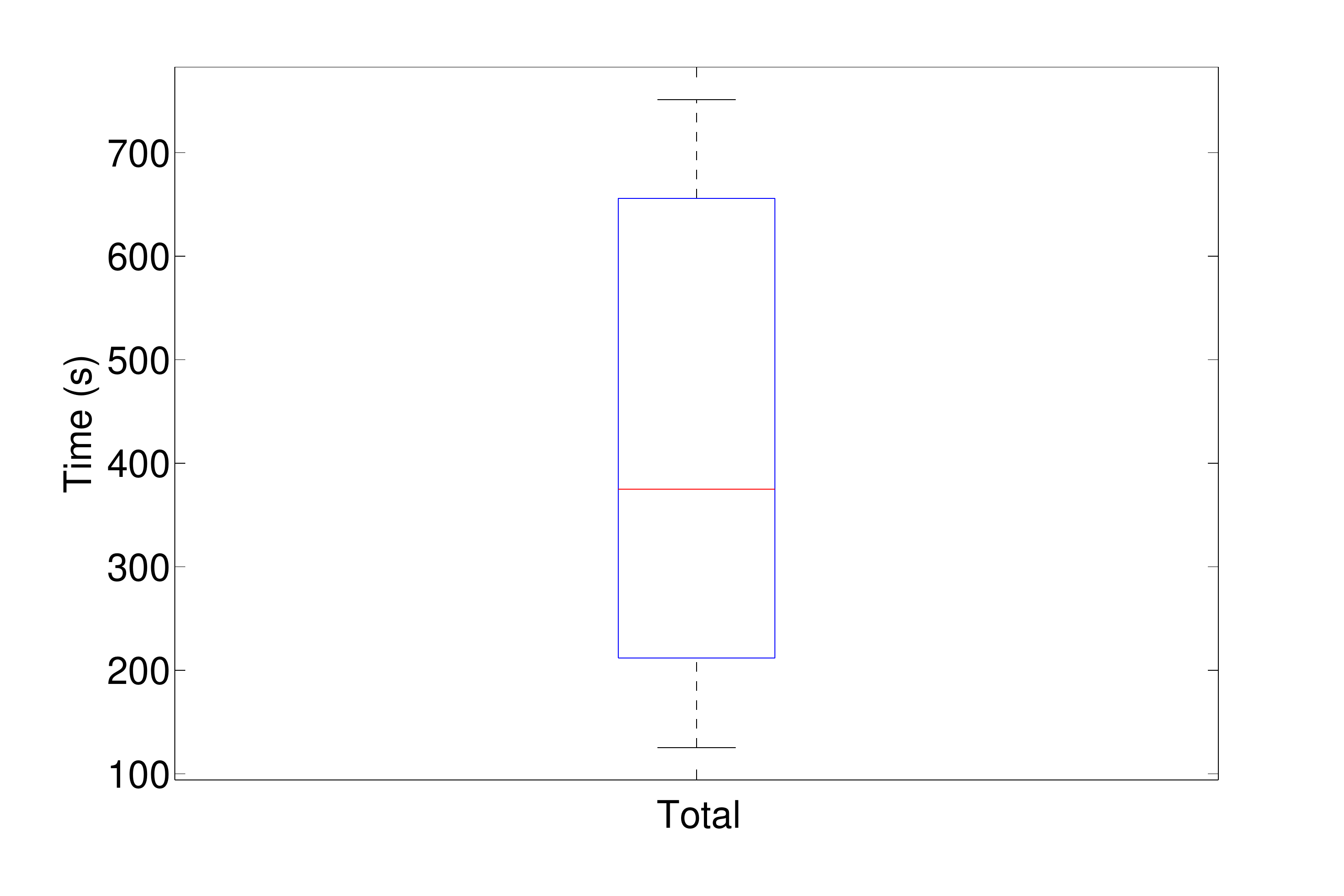}\label{fig:expo_boxplokts_by_node}
}
\caption{Time boxplots for our real case study.}
%
\end{figure}

Boxplots of Figure~\ref{fig:expo_boxplokts_by_signal} provide some quantitative insights on the computational effort we measured in solving such realistic instance. Given a signal, we report the statistical distribution of the time required by Algorithm~\ref{alg:dp} to compute covering routes from each possible start vertex. 
In general, we observe a high variance in each boxplot. Indeed, once fixed a signal $s$ in our realistic instance, it is easy to identify starting vertices from which computing covering routes is likely to be very easy or, instead, much harder. For the easy case, consider a starting vertex lying very much far away from the group of targets covered by $s$. In such case, Algorithm~\ref{alg:dp} will soon stop iterating through covering set cardinalities being not able to further generate feasible sets. Such feature is induced by the large distance of the starting vertex from the targets covered by $s$ together with the low edge density and the spatial locality shared among targets covered by the same signal (these last two are, indeed, features that frequently recur in realistic scenarios). For the harder case, just consider a situation in which the distance of the starting vertex from the targets covered by $s$ is such that a large number of covering routes is available. An example of this kind can be inspected in Figure~\ref{fig:expo_strategies}. Interestingly, a similar high variance trend cannot be observed when depicting the statistical distribution of the compute time per starting vertex. The boxplot of Figure~\ref{fig:expo_boxplokts_by_node} suggests that, by fixing the starting vertex and solving for different signals, hard instances counterbalance, on average, the easy ones.

\section{Related works}
\label{sec:related_works}
In the last few years, Security Games received an increasing interest from the Artificial Intelligence scientific community, leading to the exploration of a large number of research directions around this topic. In this section, we briefly discuss what we deem to be the most significant ones, starting from the game theoretical foundations on which these models are built.

Computing \emph{solution concepts} is the central problem upon which the real applicability of these game theoretical models is based. A lot of works concentrated on algorithmic studies of this topic, analysing the relationships holding among different kinds of solution concepts and their computational complexity.  In~\cite{relationships} the relationship between Stackelberg, Nash and min--max equilibria is studied, while in~\cite{refinements} some refinements of the Stackelberg equilibrium are proposed. Many efforts have been made to develop tractable algorithms for finding Stackelberg equilibria in Bayesian games~\cite{quality}. Furthermore, in~\cite{multiobjective} the authors analysed scenarios in which the Defender has multiple objectives, searching for the Pareto curve of the Stackelberg equilibria. 

Besides fundamental works like the ones cited above, a more recent research line devoted efforts towards the definition of game model refinements in the attempt to overcome some of their ideal assumptions. One remarkable issue belonging to this scope is how to model the behaviour of the Attacker.  In the attempt to have a more realistic behaviour, some works considered \emph{bounded rationality} and defined algorithms to deal with it. In~\cite{modelli} different models of the Attacker are analysed while in~\cite{an2013security,yang2014adaptive} the Attacker is allowed to have different observation and planning capabilities. Moreover, in~\cite{quantal} Quantal--Best Response is used to model the behaviour of the Attacker and in~\cite{scalingboundedrationality} algorithms that scale up with bounded rational adversaries are proposed. In our paper, we assume that the attacker is rational. 

Other model refinements focused on those cases in which games exhibit \emph{specific structures} that can be leveraged in the design of algorithms to compute the Stackelberg equilibrium. For instance, the study of the spread of contagion over a network is investigated in~\cite{contagion}. When no scheduling constraints are present and payoffs exhibit a special form, the computation of a Stackelberg equilibrium can be done very efficiently enabling the resolution of remarkably big scenarios~\cite{massive}. In~\cite{BlumAAAI14} realistic aspects of infrastructures to be protected are taken into account.
\section{Conclusions and future research}
\label{sec:conclusions_and_future_research}
In this paper we provide the first Security Game for large environments surveillance, e.g. for fair sites protection, that can exploit an alarm system with spatially uncertain signals.
To monitor and protect large infrastructure such as stations, airports, and cities, a two--level paradigm is commonly adopted: a broad area surveillance phase, where an attack is detected but only approximately localized due to the spatially uncertainty of the alarm system, triggers a local investigation phase, where guards have to find and clear the attack. Abstracting away from technological details, we propose a simple model of alarm systems that can be widely adopted with every specific technology and we include it in the state--of--art patrolling models, obtaining a new security game model. We show that the problem of finding the best patrolling strategy to respond to a given alarm signal is $\mathcal{APX}$--hard with arbitrary graphs even when the game is zero--sum. Then, we provide two exponential--time exact algorithms to find the best patrolling strategy to respond to a given alarm signal. The first algorithm performs a breath--first search by exploiting a dynamic programming approach, while the second algorithm performs a depth--first approach by exploiting a branch--and--bound approach. We provide also a variation of these two algorithms to find an approximate solution. We experimentally evaluate our exact and approximation algorithms both in worst--case instances, to evaluate empirically the gap between our hardness results and the theoretical guarantees of our approximation algorithms, and in one realistic instance, Expo 2015. The limit of our exact algorithms is about 16 targets with worst--case instances while we were able to compute an optimal solution for a realistic instance with $\approx 170$ targets. On the other side, our approximation algorithms provide a very effective approximation even with worst--case instances. We provide also results for special topologies, showing that our dynamic programming algorithm requires polynomial time with linear and cycle graphs, while the problem is $\mathcal{NP}$--hard with tree graphs. Finally, we focus on the problem of patrolling the environment, showing that if every target is alarmed and no false positives and missed detections are present, then the best patrolling strategy prescribes that the patroller stays in a given place waiting for an alarm signal. Furthermore, we show that such a strategy may be optimal even for missed detection rates up to 50\%.

Of course, our research does not end here since some problems related to our model remain open. The main theoretical issue is the closure of the approximation gap of SRG--$v$. We believe that investigating the relationship between our model and the DEADLINE--TSP could help in closing the gap. Another interesting problem is the study of approximation algorithms for tree graphs. Our $\mathcal{NP}$--hardness result does not exclude the existence of a PTAS (i.e., polynomial time approximation scheme), even if we conjecture that the existence is unlikely. In addition, a number of extensions of our model are worth being explored. The most important extension is to include false positives and missed detections, allowing the patroller to patrol even in absence of alarm signals. Other interesting extensions regard cases in which the number of patrollers is larger than one or there are multiple attackers, which coordinate to perform their malicious attack. Finally, a different research direction stemming from the problem concerns the alarm system dimension. Indeed, trying to deploy sensors and devices in the environment in such a way to maximize the utility in responding to alarms is a non--trivial and interesting problem, mainly due to the inherent budget constraint and trade--offs that would exhibit.

\bibliographystyle{elsarticle-num} 
\bibliography{patrolling_citations}

\appendix
\section{Notation}\label{appendix:notation}
We report in Tab.~\ref{tab:activityTracking} the symbols used along the paper.

\begin{table}[!htbp]
\small
    \centering
\begin{tabular}{|l|l|l|}
	\cline{1-3}
    & Symbol & Meaning \\
	\hline \hline
    \multirow{14}{*}{\rotatebox{90}{Basic model}}
    & $\mathcal{A}$ & Attacker \\ \cline{2-3}
    & $\mathcal{D}$ & Defender \\ \cline{2-3}
    & $G$ & Graph \\ \cline{2-3}
    & $V$ & Set of vertices \\ \cline{2-3}
    & $v$ & Vertex \\ \cline{2-3}
    & $v_i$ & $i$--th vertex \\ \cline{2-3}
    & $E$ & Set of edges \\ \cline{2-3}
    & $(v, v')$ & Edge \\ \cline{2-3}
    & $\omega^*_{v, v'}$ & Temporal cost (in turns) of the shortest path between $v$ and $v'$ \\ \cline{2-3}
    & $T$ & Set of targets \\ \cline{2-3}
	& $t$ & Target \\ \cline{2-3}
    & $t_i$ & $i$--th target \\ \cline{2-3}    
    & $\pi(t)$ & Value of target $t$ \\ \cline{2-3}
    & $d(t)$ & Penetration time of target $t$ \\
    
	\hline \hline
	
	\multirow{8}{*}{\rotatebox{90}{Signals}}
    & $S$ & Set of signals \\ \cline{2-3}
    & $s$ & Signal \\ \cline{2-3}
    & $p$ & Function specifying the probability of having the system generating \\
    & & signal $s$ given that target t has been attacked \\ \cline{2-3}
    & $T(s)$ & Targets having a positive probability of raising $s$ if attacked \\ \cline{2-3}
    & $S(t)$ & Signals having a positive probability of being raised if $t$ is attacked \\ \cline{2-3}
    & $\bot$ & No signals have been generated \\ \cline{2-3}
    & $\bigtriangleup$ & No targets are under attack \\
    
	\hline \hline    
    
    \multirow{13}{*}{\rotatebox{90}{Routes}}
    & $R_{v, s}$ & Set of routes starting from vertex $v$ when signal $s$ is generated \\ \cline{2-3}
    & $r$ & Route \\ \cline{2-3}
    & $r_i$ & $i$--th route \\ \cline{2-3}
    & $r(i)$ & $i$--th element visited along route $r$ \\ \cline{2-3}
    & $U_{\mathcal{A}}(r_i, t_i)$ & Attacker's utility given a route $r$ and a target $t$ \\ \cline{2-3}
    & $\sigma^\mathcal{D}$ & Defender's strategy \\ \cline{2-3}
    & $\sigma^\mathcal{D}_{v}$ & Defender's strategy starting from vertex $v$ \\ \cline{2-3}
    & $\sigma^\mathcal{D}_{v, s}$ & Defender's strategy starting from vertex $v$ when signal $s$ is generated \\ \cline{2-3}
    & $\sigma^\mathcal{A}$ & Attacker's strategy \\ \cline{2-3}
    & $\sigma^\mathcal{A}_v$ & Attacker's strategy when $\mathcal{D}$ is in $v$\\ \cline{2-3}
    & $g_v$ & Value of the game (utility of $\mathcal{A}$) \\ \cline{2-3}
    & $A(r(i))$ & Time needed by $\mathcal{D}$ to visit $r(i)$ starting from $r(0)$ \\ \cline{2-3}
    & $T(r)$ & Set of targets covered by route $r$ \\ \cline{2-3}
    & $c(r)$ & Temporal cost (in turns) associated to $r$ \\ 
	\hline
\end{tabular}
\caption[Symbols table]{Symbols table.}
 \label{tab:activityTracking}
\end{table}
\section{Additional experimental results}
\label{appendix:boxplots}
\begin{figure}[!htbp]
\scriptsize
\begin{center}
\begin{tabular}{|cc|cc|}\hline
\begin{sideways}\hspace{2.25cm}$\epsilon = 0.05$\end{sideways}		& \hspace{-0.50cm}	\includegraphics[scale=0.5]{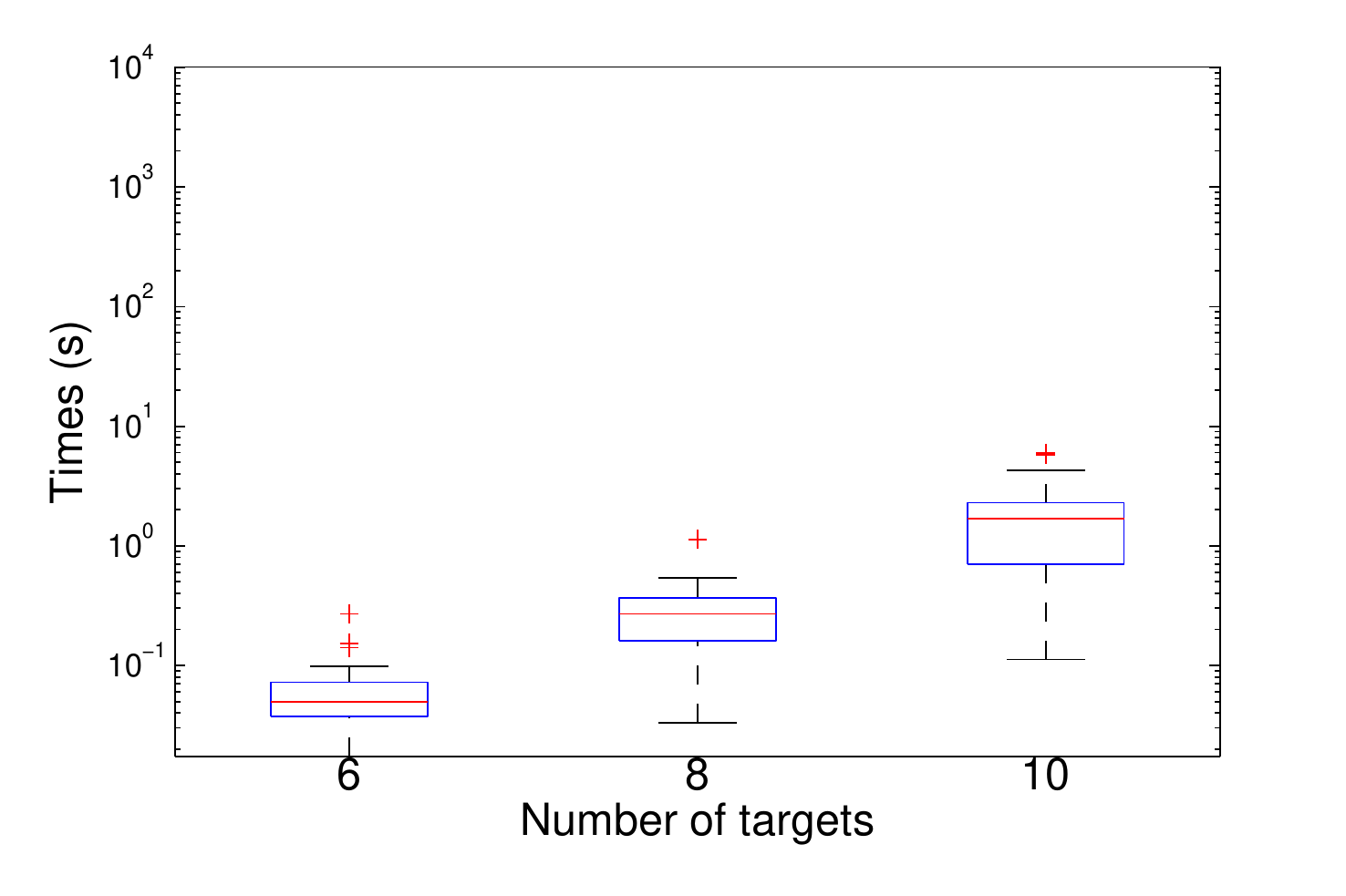}			\hspace{-0.55cm}	&
\begin{sideways}\hspace{2.25cm}$\epsilon = 0.10$\end{sideways}		& \hspace{-0.55cm}	\includegraphics[scale=0.5]{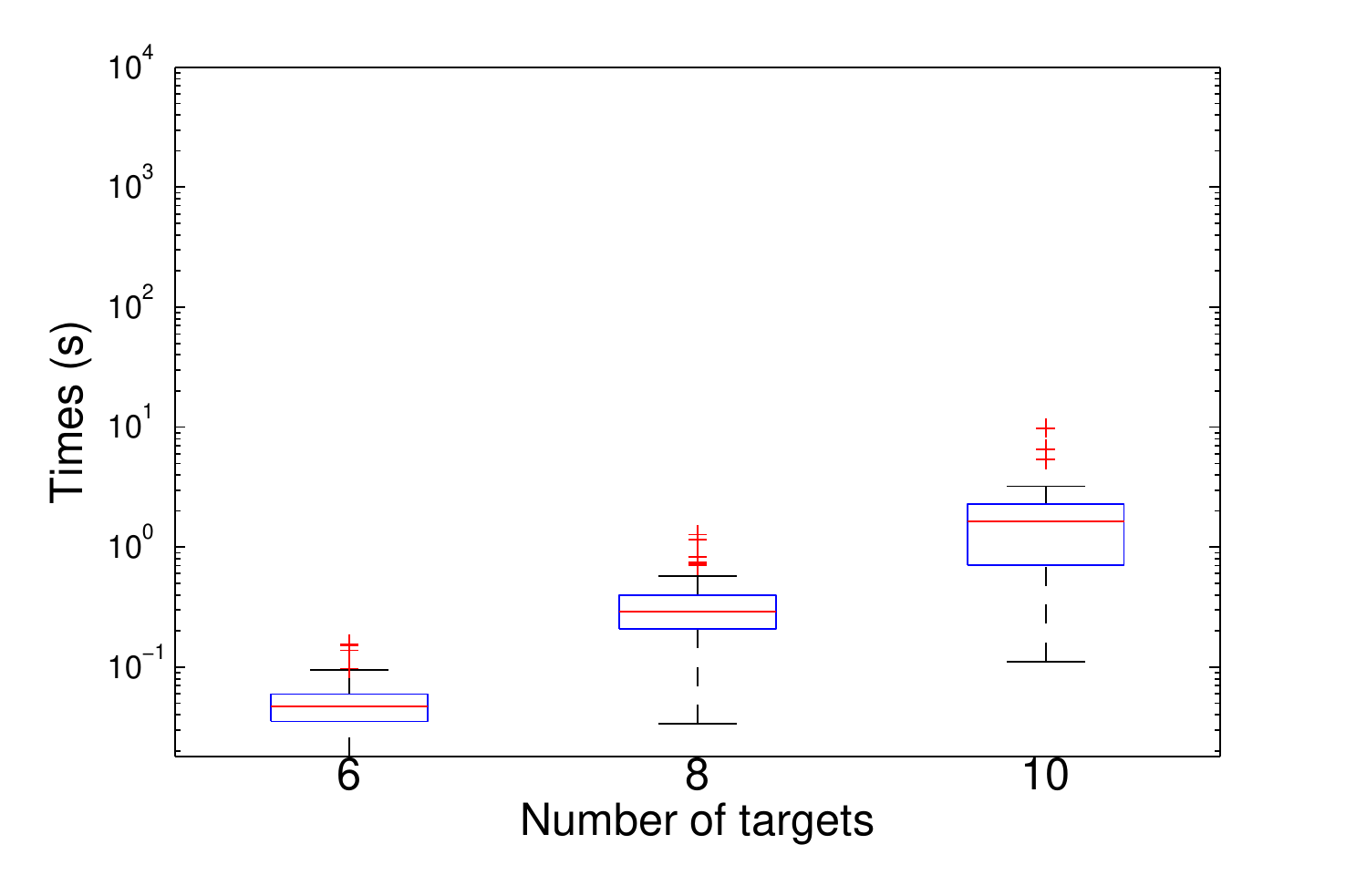}			\hspace{-0.8cm}		\\	\hline
\begin{sideways}\hspace{2.25cm}$\epsilon = 0.25$\end{sideways}		& \hspace{-0.50cm}	\includegraphics[scale=0.5]{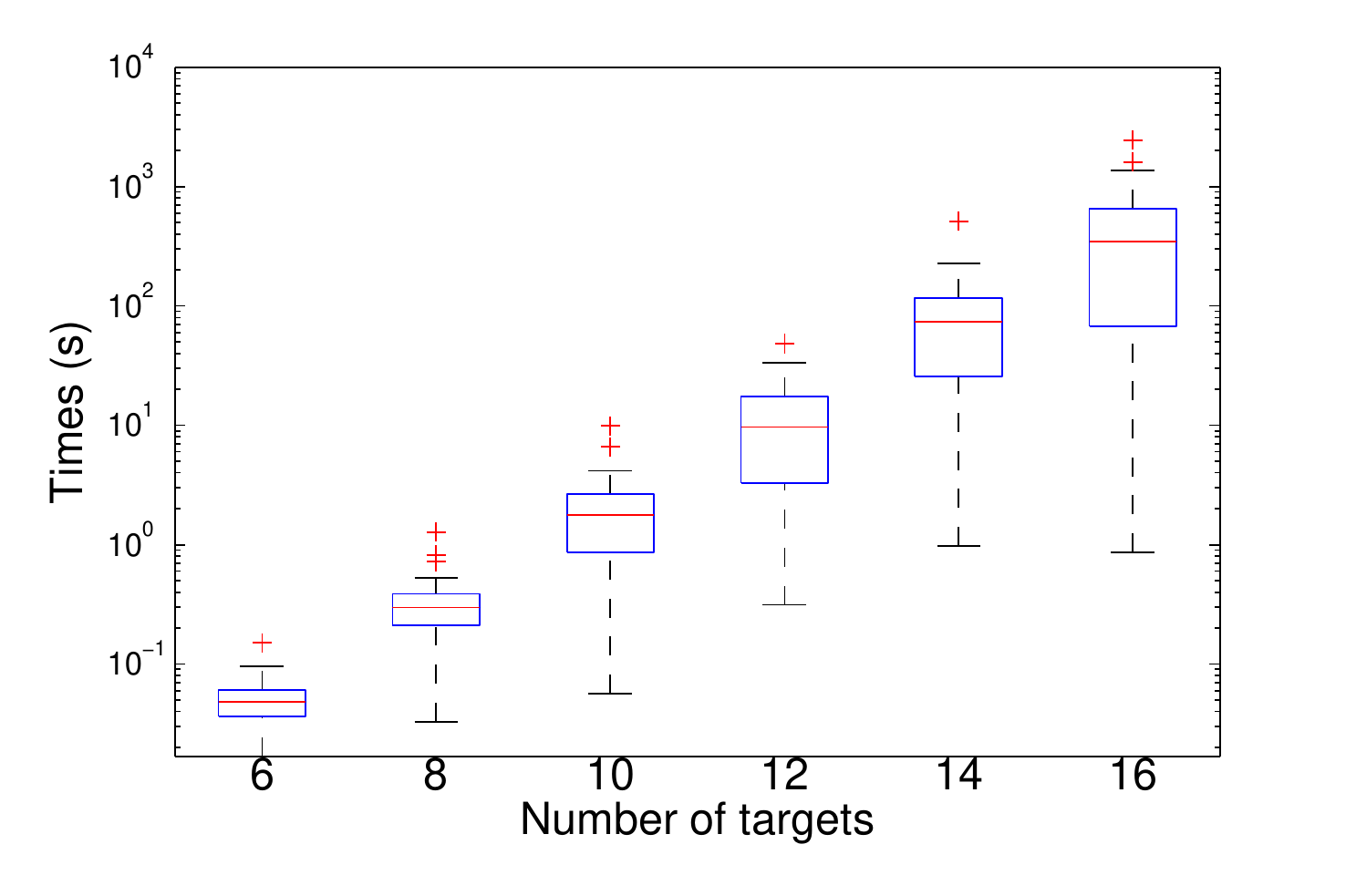}			\hspace{-0.55cm}	&	
\begin{sideways}\hspace{2.25cm}$\epsilon = 0.50$\end{sideways}		&\hspace{-0.55cm}	\includegraphics[scale=0.5]{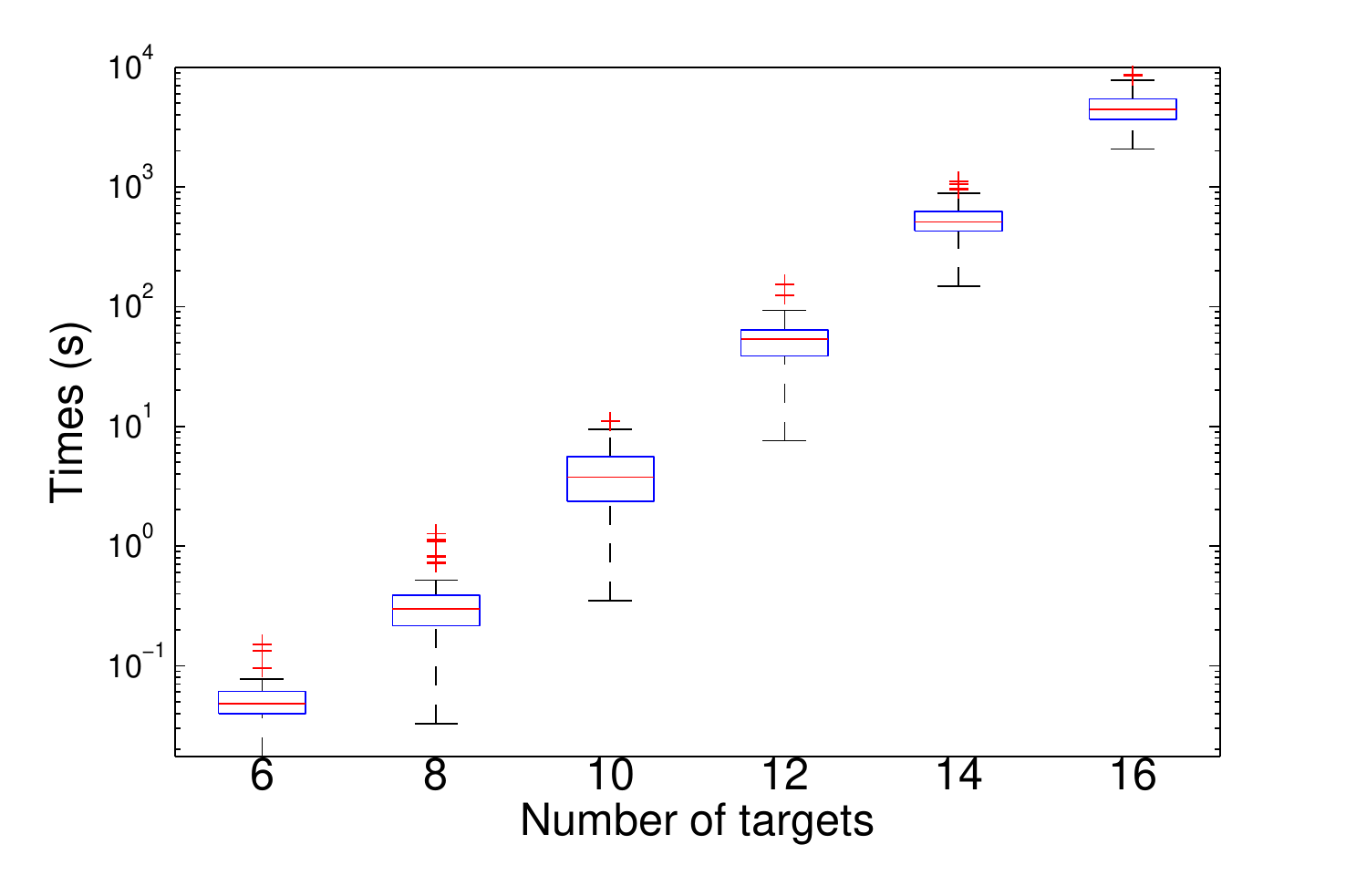}			\hspace{-0.8cm}		\\	\hline
\begin{sideways}\hspace{2.25cm}$\epsilon = 0.75$\end{sideways}		& \hspace{-0.50cm}	\includegraphics[scale=0.5]{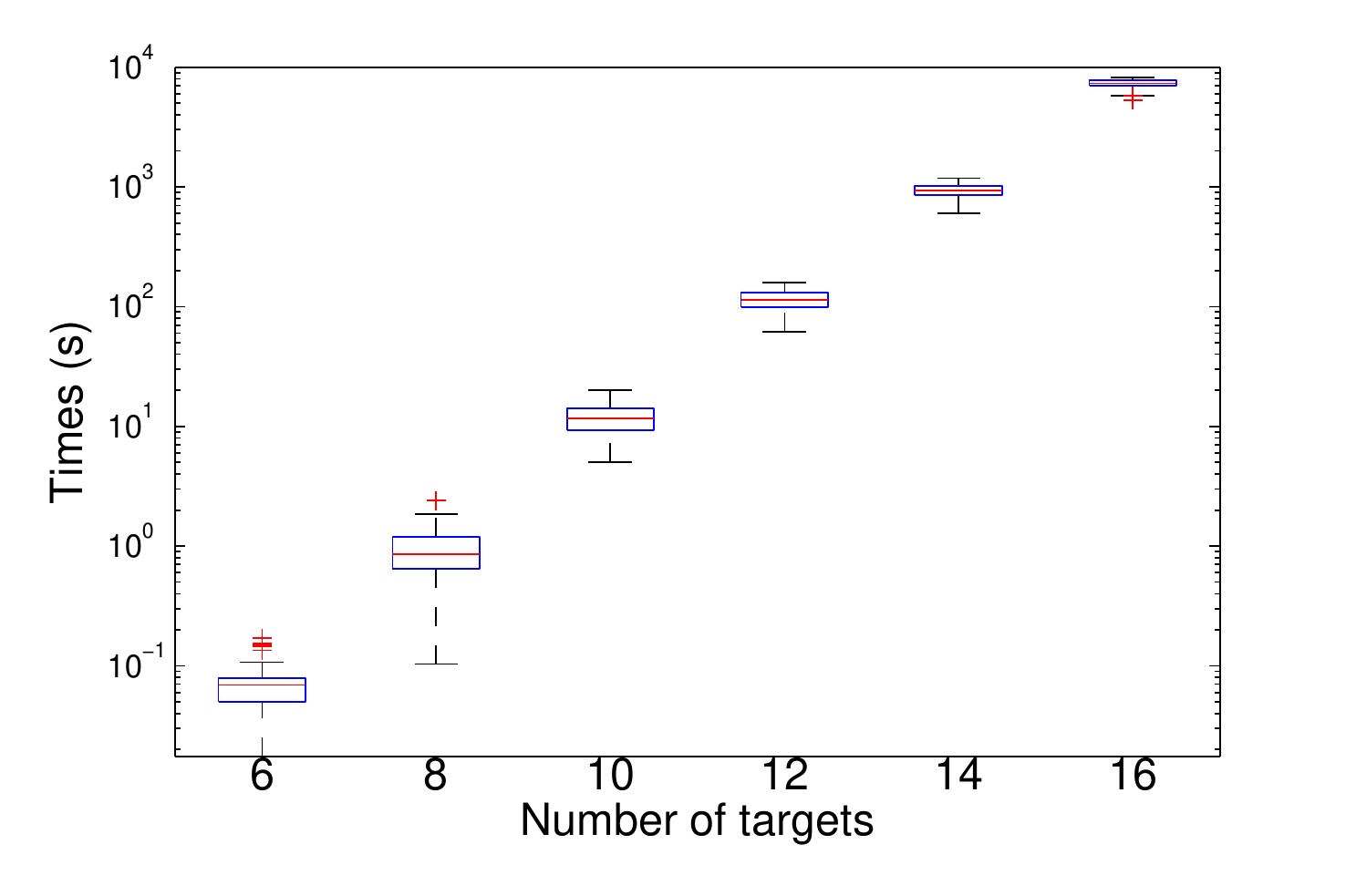}			\hspace{-0.55cm}	&	
\begin{sideways}\hspace{2.25cm}$\epsilon = 1.00$\end{sideways}		&\hspace{-0.55cm}	\includegraphics[scale=0.5]{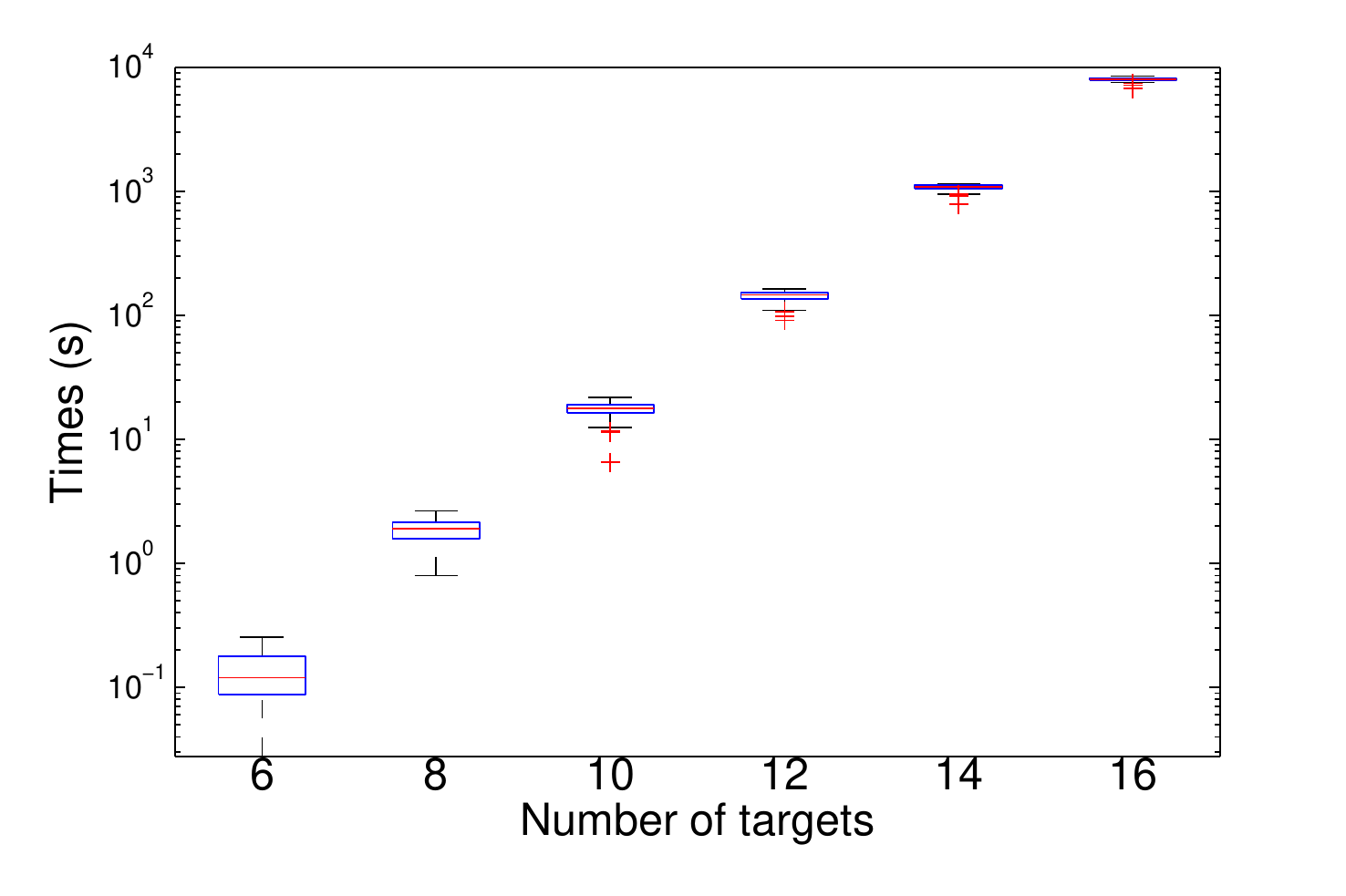}			\hspace{-0.8cm}		\\ 	\hline
\end{tabular}
\end{center}
\caption{Boxplots of compute times required by our exact dynamic programming algorithm.}
\label{fig:boxplotcomputetime}
\end{figure}

We report in Fig.~\ref{fig:boxplotcomputetime} the boxplots of the results depicted in Fig.~\ref{fig:exacttimes}. They show that the variance of the compute times drastically reduces as $\epsilon$ increases. This is because the number of edges increases as $\epsilon$ increases and so the number of proper covering sets increases approaching $2^{|T|}$. On the other hand, with small values of $\epsilon$, the number of proper covering sets of different instances can be extremely different.

\end{document}